\documentclass[sn-mathphys-num]{sn-jnl}%

\usepackage{graphicx}%
\usepackage{multirow}%
\usepackage{amsmath,amssymb,amsfonts}%
\usepackage{amsthm}%
\usepackage{mathrsfs}%
\usepackage[title]{appendix}%
\usepackage{xcolor}%
\usepackage{textcomp}%
\usepackage{manyfoot}%
\usepackage{booktabs}%
\usepackage{algorithm}%
\usepackage{algorithmicx}%
\usepackage{algpseudocode}%
\usepackage{listings}%
\usepackage{array}

\usepackage{hyperref}
\usepackage{xcolor}

\hypersetup{
    colorlinks=true,
    linkcolor=magenta, 
    filecolor=magenta, 
    urlcolor=pink 
}

\usepackage{bm} 
\usepackage{subcaption}

\newcommand{\rebuttal}[1]{\textcolor{black}{#1}}

\theoremstyle{thmstyleone}%

\theoremstyle{thmstyletwo}%

\theoremstyle{thmstylethree}%

\begin{document}

\title[Article Title]{Periodic Vibration Gaussian: Dynamic Urban Scene Reconstruction and Real-time Rendering}

\author[1]{\fnm{Yurui} \sur{Chen}}
\equalcont{These authors contributed equally to this work.}

\author[1]{\fnm{Chun} \sur{Gu}}
\equalcont{These authors contributed equally to this work.}

\author[1]{\fnm{Junzhe} \sur{Jiang}}

\author[2]{\fnm{Xiatian} \sur{Zhu}}

\author*[1]{\fnm{Li} \sur{Zhang}}\email{lizhangfd@fudan.edu.cn}

\affil*[1]{\orgname{School of Data Science, Fudan University}}

\affil[2]{\orgname{University of Surrey}}

\abstract{
Modeling dynamic, large-scale urban scenes is challenging due to their highly intricate geometric structures and unconstrained dynamics in both space and time. Prior methods often employ high-level architectural priors, separating static and dynamic elements, resulting in suboptimal capture of their synergistic interactions. To address this challenge, we present a unified representation model, called {\bf\em Periodic Vibration Gaussian} ({\bf PVG}). PVG builds upon the efficient 3D Gaussian splatting technique, originally designed for static scene representation, by introducing periodic vibration-based temporal dynamics. This innovation enables PVG to elegantly and uniformly represent the characteristics of various objects and elements in dynamic urban scenes. To enhance temporally coherent and large scene representation learning with sparse training data, we introduce a novel temporal smoothing mechanism and a position-aware adaptive control strategy respectively. Extensive experiments on Waymo Open Dataset~\cite{sun2020scalability} and KITTI benchmarks~\cite{geiger2012we} demonstrate that PVG surpasses state-of-the-art alternatives in both reconstruction and novel view synthesis for both dynamic and static scenes. Notably, PVG achieves this without relying on manually labeled object bounding boxes or expensive optical flow estimation. Moreover, PVG exhibits {\bf 900}-fold acceleration in rendering over the best alternative. The code is available at \url{https://github.com/fudan-zvg/PVG}.}


\maketitle

\begin{figure}[h]
  \centering
  \begin{subfigure}[b]{0.95\linewidth}
      \includegraphics[width=\linewidth]{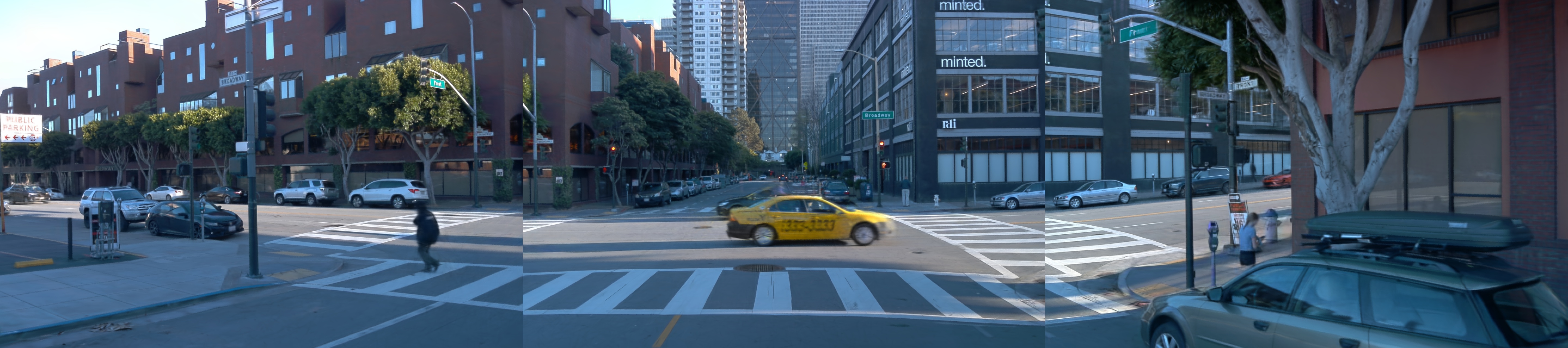}
      \caption{Dynamic scene}
  \end{subfigure} \\
  \begin{subfigure}[b]{0.95\linewidth}
      \includegraphics[width=\linewidth]{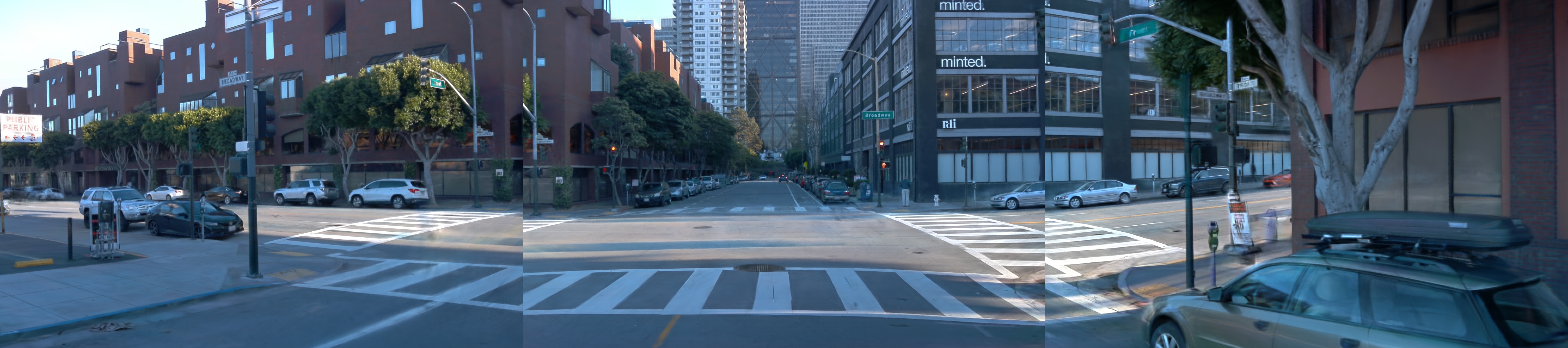}
      \caption{Remove the dynamic scene elements}
  \end{subfigure}
  \caption{Our proposed \textbf{Periodic Vibration Gaussian} is crafted to effectively and uniformly capture both static and dynamic elements of a large, dynamic urban scene.
  (\textbf{a}) It not only reconstructs a \textbf{\em dynamic} urban scene but also enables \textbf{\em real-time rendering}, while efficiently isolating dynamic components from the intricacies of the highly unconstrained and complex scene.
  (\textbf{b}) This capability facilitates flexible manipulation, such as the removal of dynamic scene elements.}
\label{fig:teaser}
\end{figure}

\section{Introduction}

The geometric reconstruction of extensive urban spaces, such as streets and cities, has played a pivotal role in applications like digital maps, auto-navigation, and autonomous driving \cite{sun2020scalability,geiger2012we,caesar2020nuscenes}. Our world is inherently dynamic and complex in both spatial and temporal dimensions. Despite advancements in scene representation techniques like Neural Radiance Fields (NeRFs) \cite{rematas2022urban,xie2023s,guo2023streetsurf}, which primarily focus on static scenes, they overlook more challenging dynamic elements.

Recent approaches to model dynamic scenes include NSG~\cite{ost2021neural}, which decomposes dynamic scenes into scene graphs and learns a structured representation. PNF~\cite{kundu2022panoptic} further decomposes scenes into objects and backgrounds, incorporating a panoptic segmentation auxiliary task. However, scalability issues arise in real-world scenarios, where obtaining accurate object-level supervisions (e.g., 3D object boxes, segmentation masks) is challenging, and explicitly representing each object linearly increases model complexity with the number of objects.

SUDS~\cite{turki2023suds} later proposes using optical flow to relax the stringent requirement of object labeling in a three-branch architecture for separately modeling static and dynamic elements and the environmental factors of a scene. EmerNeRF~\cite{yang2023emernerf} uses a self-supervised method to reduce dependence on optical flow. Despite adopting implicit NeRF representation, these methods suffer from low efficiency in both training and rendering, posing a significant bottleneck for large-scale scene rendering and reconstruction. Additionally, manually separating constituent parts introduces design complexity and limits the ability to capture intrinsic correlations and interactions.

To overcome the identified limitations, this paper introduces a novel dynamic scene representation method termed {\bf Periodic Vibration Gaussian} (PVG). This approach provides a unified representation of both static and dynamic elements within a scene through a single formulation. Building upon the efficient 3D Gaussian Splatting (3DGS))~\cite{kerbl20233d}, originally devised for static scene representation, we incorporate periodic vibration-based temporal dynamics. This modification allows for a cohesive representation of static and dynamic scene elements with explicit motion properties such as velocity and staticness. 
We also propose a position-aware point adaptive control strategy to fit the distant view better.
To improve the temporal continuity in representation learning with typically limited training data, we introduce a novel temporal smoothing mechanism.

Our {\bf contributions} are summarized as follows:
\textbf{(i)}
Introduction of the very first unified representation model, PVG, for large-scale dynamic urban scene reconstruction. In contrast to previous NeRF-based solutions, PVG employs the 3D Gaussian Splatting paradigm, uniquely extending it to elegantly represent dynamic scenes. This is accomplished by seamlessly integrating periodic vibration-based temporal dynamics into the conventional 3DGS formulation.
\textbf{(ii)}
Development of a novel temporal smoothing mechanism to enhance the temporal continuity of representation and a position-aware point adaptive control strategy for unbounded urban scenes.
\textbf{(iii)}
Extensive experiments on two large benchmarks (KITTI and Waymo) demonstrate that PVG outperforms all previous state-of-the-art alternatives in novel view synthesis. Moreover, it provides significant efficiency benefits in both training and rendering processes, achieving a remarkable {\bf 900}-fold acceleration in rendering compared to the leading competitor, EmerNeRF~\cite{yang2023emernerf}.
\rebuttal{We also show that PVG is superior in both visual quality and rendering efficiency over concurrent 3DGS based models.}

\section{Related work}
\label{gen_inst}

\noindent{\bf Neural rendering}
In the domain of novel view synthesis, Neural Radiance Fields (NeRF)~\cite{mildenhall2020nerf} have emerged as a noteworthy approach. NeRF employs a coordinate-based multi-layer perception representation of 3D scenes, leveraging volumetric rendering and the spatial smoothness of multi-layer perception to generate high-quality novel views. However, its implicit nature comes with significant drawbacks, including slow training and rendering speeds, as well as high memory usage.

To tackle these challenges, several studies have proposed solutions to enhance training speed. Techniques such as voxel grids~\cite{sun2022direct}, hash encoding~\cite{muller2022instant}, and tensor factorization~\cite{chen2022tensorf,Chen2023factor} have been explored. Others have focused on improving rendering speed by transforming implicit volumes into explicit textured meshes, as demonstrated in works like~\cite{chen2023mobilenerf,reiser2023merf,yariv2023bakedsdf}. Additionally, endeavors such as~\cite{barron2021mip,barron2022mip360,verbin2022ref,barron2023zipnerf} aim to enhance rendering quality by addressing issues like antialiasing and reflection modeling.
Recently, 3D Gaussian Splatting (3DGS) \cite{kerbl20233d} introduces an innovative point-based 3D scene representation, seamlessly integrating the high-quality volume rendering principles of NeRF with the swift rendering speed characteristic of rasterization.

\noindent{\bf Dynamic scene models}
Reconstructing dynamic scenes poses distinctive challenges, particularly in effectively handling temporal correlations across various time steps. Expanding on the accomplishments of NeRF~\cite{mildenhall2020nerf}, several extensions have been proposed to tailor NeRF to dynamic scenarios.
In one research direction, certain studies~\cite{li2022neural,fridovich2023k,cao2023hexplane,wang2023mixed,attal2023hyperreel} introduce time as an additional input to the radiance field, treating the scene as a 6D plenoptic function. However, this approach couples positional variations induced by temporal dynamics with the radiance field, lacking geometric priors about how time influences the scene.
An alternative approach~\cite{pumarola2021d,park2021nerfies,park2021hypernerf,tretschk2021non,abou2022particlenerf,luiten2023dynamic,wu20244d} focuses on modeling the movement or deformation of specific static structures, assuming that the dynamics arise from these static elements within the scene. Point-based methods~\cite{abou2022particlenerf,luiten2023dynamic,wu20244d} have shown promise in addressing the challenges of reconstructing dynamic scenes due to their flexibility. Building upon the progress in 3DGS, recent works~\cite{yang2023deformable,wu20244d} propose the use of a set of deformable 3D Gaussians for optimization across different timestamps. While this kind of methods need to learn the deform function in the dense space which is difficult to migrate to large scenarios. \rebuttal{\cite{yang2024real} extends 3DGS to a 4D formulation, enabling modeling of the full time-space manifold. While the formulation induces a dynamic opacity model through spatio-temporal 4D Gaussians, it does not explicitly model the separation between static and dynamic elements. In contrast, we introduce a staticity coefficient to represent the per-point degree of motion, enabling effective disentanglement of static and dynamic components. To address the challenge of temporally sparse observations, we propose a temporal smoothing training strategy that enhances temporal consistency and reconstruction robustness. Furthermore, our framework incorporates several domain-specific designs for autonomous driving scenarios, including LiDAR-projected depth supervision, cube map-based sky modeling, and a positional-aware control mechanism for handling large-scale distant structures. These innovations collectively contribute to higher-quality reconstruction and synthesis in complex urban environments.
}

\noindent{\bf Urban scene reconstruction}
NeRF-based techniques have shown their efficacy in autonomous driving scenarios~\cite{geiger2012we, sun2020scalability}. One research avenue has focused on enhancing the modeling of static street scenes by utilizing scalable representations~\cite{tancik2022block, turki2022mega, li2023read, rematas2022urban}, achieving high-fidelity surface reconstruction~\cite{rematas2022urban, wang2023neural, guo2023streetsurf}, and incorporating multi-object composition~\cite{xie2023s}. However, these methods face difficulties in handling dynamic elements commonly encountered in autonomous driving contexts.
Another research direction seeks to address these challenges. Notably, these techniques require additional input, such as leveraging panoptic segmentation to refine the dynamics of reconstruction~\cite{kundu2022panoptic}. Moreover, in~\cite{ost2021neural}, scene graphs are employed to decompose dynamic multi-object scenes, while in~\cite{yang2023unisim}, neural shape priors are learned for completing dynamic object reconstructions. In \cite{wu2023mars} foreground instances and background environments are decomposed.
\rebuttal{3DGS-based methods~\cite{yan2024street, chen2024omnire, zhou2023drivinggaussian, zhou2024hugs} have been concurrently proposed for dynamic urban scene reconstruction. However, these approaches generally depend on object-level supervision via 3D bounding boxes. While such supervision can improve the reconstruction quality of dynamic objects, it increases model complexity and reduces flexibility. Moreover, reliance on automatically generated bounding boxes introduces noisy supervision and often fails in challenging cases such as distant or occluded objects. In addition, explicitly separating dynamic objects from the background may compromise the overall consistency of the reconstructed scene.}
In \cite{turki2023suds}, a scalable hash table is proposed for large-scale dynamic scenes, relying on an off-the-shelf 2D optical flow estimator to track dynamic actors.~\cite{yang2023emernerf} reduces dependence on optical flow by self-supervision, however, it still suffers low image quality and rendering speed.

In this paper, we present an elegant extension of 3D Gaussian Splatting~\cite{kerbl20233d} with the additional time dimension to handle the complexities of dynamic scenes. Our model provides a uniform, efficient representation, excelling in reconstructing dynamic, large-scale urban scenes without the dependence on manual annotations or pre-trained models.

\section{Method}

Utilizing the sequentially acquired and calibrated multi-sensor data, encompassing a set of images $\mathcal{I}$, each captured by cameras equipped with corresponding intrinsic matrices $\mathbf{I}$ and extrinsic matrices $\mathbf{E}$, alongside their respective capture timestamps $t$, collectively represented as $\{\mathcal{I}_i, t_i, \mathbf{E}_i, \mathbf{I}_i| i=1,2,\dots N_c\}$, and the spatial coordinates of LiDAR point clouds annotated with timestamps $\{(x_i, y_i, z_i, t_i) | i=1,2,\dots N_l\}$, where $N_c$ and $N_l$ are the number of image frames and Lidar points, our research aims to achieve precise 3D reconstruction and synthesize novel viewpoints at any desired timestamp $t$ and camera pose $[\mathbf{E}_o, \mathbf{I}_o]$. To this end, our framework is meticulously engineered to approximate a rendering function $\hat{\mathcal{I}} = \mathcal{F}_{\theta}(\mathbf{E}_o, \mathbf{I}_o, t)$.

\subsection{Preliminary}
\label{section:3D Gaussian}

3DGS~\cite{kerbl20233d} utilizes a collection of 3D Gaussians to represent a scene. Through a tile-based rasterization process, 3DGS facilitates real-time alpha blending of numerous Gaussians. The scene is modeled by a set of points $\{P_i\}$, where each point $P$ is linked to a mean $\bm{\mu}\in \mathbb{R}^3$, a covariance matrix $\Sigma \in \mathbb{R}^{3 \times 3}$, an opacity $o$, and a color $\mathbf{c}$. These attributes collectively define the point's influence within the 3D space as:
\begin{equation}
\label{eq:gaussian}
	G(\bm{x})~= e^{-\frac{1}{2}(\bm{x}-\bm{\mu})^{T}\Sigma^{-1}(\bm{x}-\bm{\mu})}.
\end{equation}

To create an image from a particular viewpoint, 3DGS maps each Gaussian point onto the image plane, yielding a collection of 2D Gaussians. Calculating the means of these projected Gaussians is straightforward. However, determining the formulation for the projected variance involves
\begin{equation}
	\label{eq:project}
	\Sigma' = J W ~\Sigma ~W ^{T}J^{T},
\end{equation}
where $W$ and $J$ are the view transform matrix and Jacobian of the nonlinear projective transform matrix, respectively. 
Sorting the Gaussians according to their depth in camera space, we can query the attributes of each 2D Gaussian and facilitate the subsequent volume rendering process to determine the color of each pixel:
\begin{equation}
\label{alpha-blending}
    C = \sum^{N}_{i=1} T_i\alpha_i \bm{c}_i \hspace{0.5em} \\ \text{ with } \hspace{0.5em} T_i=\prod^{i-1}_{j=1}(1-\alpha_j),
\end{equation}
where $\alpha$ is derived through the product of the opacity ${o}$ and the contribution from the 2D covariance calculated using $\Sigma'$ and the corresponding pixel coordinates in the image space. 
The covariance matrix holds a meaningful interpretation when it is positive semi-definite. In the context of 3DGS, it is decomposed into a scaling matrix, denoted as a diagonal matrix represented by $\bm{s}\in \mathbb{R}^{3}$, and a rotation matrix represented by a unit quaternion $\bm{q}$.

The differentiable rendering function for a new view of a scene containing $N$ points is expressed as
\begin{equation}
\label{eq: 3d gs render}
    \hat{\mathcal{I}} = \mathrm{Render}(\{\mathcal{C}_i | i=1, 2, \dots, N \}; \mathbf{E}, \mathbf{I}),
\end{equation}
where $\hat{\mathcal{I}}$ is the rendered image, and $\mathbf{E}$ and $\mathbf{I}$ denote the camera extrinsic and intrinsic matrices, respectively.
The training of the model entails optimizing the parameter set for each point, represented as $\mathcal{C} = \{\bm{\mu}, \bm{q}, \bm{s}, o, \bm{c}\}$.

\noindent{\bf Flexible rendering}
This rendering method can tackle different targets, like depth and opacity, by replacing the color $\bm{c}$ in Eq.~\eqref{alpha-blending}. For example, the normalized depth map can be computed as: $\sum^{N}_{i=1} T_i\alpha_i z_i / \sum^{N}_{i=1} T_i\alpha_i$, where $z_i$ represents the distance of the center of a Gaussian point from the image plane.

\noindent{\bf Limitation}
The 3DGS model represents static points in a scene, lacking the ability to capture dynamic changes over time, essential for modeling dynamic urban scenes. To address this limitation, we propose the Periodic Vibration Gaussian (PVG) model.

\subsection{Periodic Vibration Gaussian (PVG)}
\label{section:harmonic gaussian}

\begin{figure*}[t] 
  \centering
  \centering
  \includegraphics[width=\textwidth]{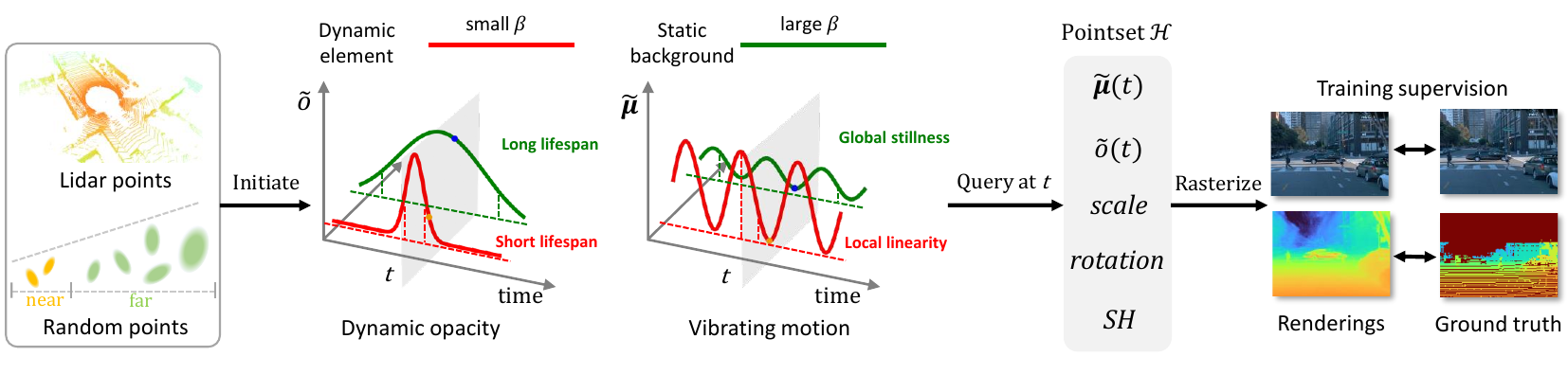}
  \caption{\rebuttal{PVG learns an adaptive opacity decay rate %
  to distinguish between dynamic and static scene elements. The model represents dynamic objects with short-lifespan points that quickly fade, while static regions are modeled by points with a longer lifespan, allowing them to exhibit globally consistent behavior over time. This learning process is guided by supervision from RGB and LiDAR depth signals at each time step.}}
  \label{fig:pipeline}
\end{figure*}

\noindent{Our PVG model exhibits several distinctive features:}

\noindent{\bf Dynamics introduction}: We introduce the concept of \textbf{\em life peak}, denoted as $\tau$, which represents the point's moment of maximum prominence over time. The motivation behind this concept is to assign a distinct lifespan to each Gaussian point, defining when it actively contributes and to what degree. This fundamentally infuses a dynamic nature into the model, enabling variations in the collection of Gaussian points that influence the rendering of the scene over time.

\noindent{\bf Periodic vibration}: 
We modify the traditional 3D Gaussian's mean $\bm{\mu}$ and opacity $o$ to be time-dependent functions centered around the life peak $\tau$, denoted as $\widetilde{\bm{\mu}}(t)$ and $\widetilde{o}(t)$. Both functions peak at $\tau$. This adaptation empowers the model to effectively capture dynamic motions, enabling each point to adjust based on temporal changes. 

Formally, our  model, denoted as $\mathcal{H}$, is expressed as:
\begin{align}
\label{eq: transto3d}
        \mathcal{H}(t) &= \{ \widetilde{\bm{\mu}}(t),\bm{q},\bm{s},\widetilde{o}(t),\bm{c} \}, \\
        \label{eq: mu}
        \quad \widetilde{\bm{\mu}} (t) &= \bm{\mu}+ \frac{l}{2\pi}\cdot \sin(\frac{2\pi (t-\tau)}{l})\cdot \bm{v}, \\
        \widetilde{o} (t) &= o \cdot  e^{-\frac{1}{2}(t-\tau)^2 \beta^{-2}},
\end{align}
where $\widetilde{\bm{\mu}}(t)$ represents the vibrating motion centered at $\bm{\mu}$ occurring at the life peak $\tau$, and $\widetilde{o}(t)$ denotes the vibrating opacity, which decays away from the peak $\tau$ with a decay rate inversely proportional to $\beta$. Notably, the parameter $\beta$ governs the lifespan around $\tau$, with larger values indicating bigger lifespans. The hyper-parameter $l$ represents the cycle length, serving as the scene prior.
The learnable parameter $\bm{v}=\frac{\mathrm{d} \widetilde{\bm{\mu}}(t)}{\mathrm{d} t} |_{t=\tau} \in \mathbb{R}^{3}$ signifies the vibrating direction and denotes the instant velocity at time $\tau$. Therefore, the per-point learnable parameters of our model $\mathcal{H}$ include $\{\bm{\mu},\bm{q},\bm{s},o, \bm{c}, \tau, \beta, \bm{v}\}$.

In particular, we express the mean vector (position) $\widetilde{\bm{\mu}}(t)$ through periodic vibrations, 
providing a cohesive framework for both static and dynamic patterns. To enhance clarity, we introduce the \textbf{\em staticness coefficient} $\rho=\frac{\beta}{l}$, which quantifies the degree of staticness exhibited by a PVG point and is also associated with the point's lifespan.
Periodic vibration facilitates convergence around $\bm{\mu}$ when $\rho$ is large. This is due to the bounded nature of $\widetilde{\bm{\mu}}(t)$ by $\bm{v}$ and the fact that $\mathbb{E} [\widetilde{\bm{\mu}}(t)]= \bm{\mu}$ holds for any time interval with a length that is a multiple of $l$, independent of $\bm{v}$. 

We note that the 3DGS represents a particular case of PVG in which $\bm{v} = \mathbf{0}$ and $\rho=+\infty$. PVGs with large $\rho$ effectively capture the static aspects in a scene, provided that $\|\bm{v}\|$ remains within a reasonable range.

The ability of our PVG to represent the dynamic aspects of a scene is particularly evident in points with small $\rho$. Points approaching $\rho \to 0$ manifest by appearing and disappearing almost instantaneously, executing linear movements around the time $\tau$. As time progresses, these points undergo oscillations, with some vanishing and others emerging. At a specific timestamp $t$, dynamic objects are more likely to be predominantly represented by points with $\tau$ close to $t$. In essence, different points take charge of representing dynamic objects at distinct timestamps. 

\rebuttal{
Conversely, the static components of a scene can be effectively represented by points exhibiting large $\rho$.
Introducing a threshold on $\rho$ enables us to discern whether a point represents dynamic elements (see Fig.~\ref{fig:teaser}(b)). Conceptually, for the periodic-function-based design, this does not imply that the underlying motion in the scene is inherently periodic; Rather, our periodic formulation can be regarded as a primitive for fitting complex motions, while still preserving the global static property.}

It is crucial to emphasize that at any given time $t$, our model takes the form of a specific 3D Gaussian model, represented by $\mathcal{H}(t)$. We train a collection of PVG points, denoted as $\{\mathcal{H}_i\}$, to effectively portray a dynamic scene. The rendering process is then executed as:
\begin{equation}
\label{eq: render shooting gaussian}
\hat{\mathcal{I}} = \mathrm{Render}(\{\mathcal{H}_i(t) | i=1, \dots, N_H \}; \mathbf{E}, \mathbf{I}),
\end{equation}
where $N_H$ represents the number of PVG points in a scene. Our training pipeline can be seen in Fig.~\ref{fig:pipeline}(b).

\subsection{Position-aware point adaptive control}
\label{section:adaptive control}

The conventional adaptive control method, as introduced in~\cite{kerbl20233d}, treating each Gaussian point uniformly, proves inadequate for urban scenes. This is mainly attributed to the substantial distance of the mean vector (position) $\bm{\mu}$ for most points from the center of the unbounded scene. To faithfully represent the scene with fewer points without sacrificing accuracy, we advocate utilizing larger points for distant locations and smaller points for nearby areas. 

Assuming camera poses are centered, the inclusion of the scale factor $\gamma(\bm{\mu})$ as defined below is essential for effective control over each PVG point:
\begin{align}
\begin{split}
\gamma(\bm{\mu}) = \left \{
\begin{array}{ll}
    1                  &\mathrm{if} \quad \|\bm{\mu}\|_2 < 2r\\
    \|\bm{\mu}\|_2/r - 1     & \mathrm{if}  \quad \|\bm{\mu}\|_2 \geq 2r,
\end{array}
\right.
\end{split}
\end{align}
where $r$ denotes the scene radius (i.e., the scene scope). 
Specifically, we employ a densification strategy for a PVG $\mathcal{H}(t)$ when its backward gradient on view space surpasses a specified threshold. We opt to clone the PVG if $\max(\bm{s}) \leq g \cdot \gamma(\bm{\mu})$, with $g$ serving as the threshold for scale. Conversely, if this condition is not met, we initiate a split operation. Additionally, we undertake pruning of points with $\max(\bm{s}) > b \cdot \gamma(\bm{\mu})$, employing $b$ as the scale threshold to discern whether a given PVG is excessively large.

\subsection{Model training}
\begin{figure}[t] 
  \centering
  \centering
  \includegraphics[width=\textwidth]{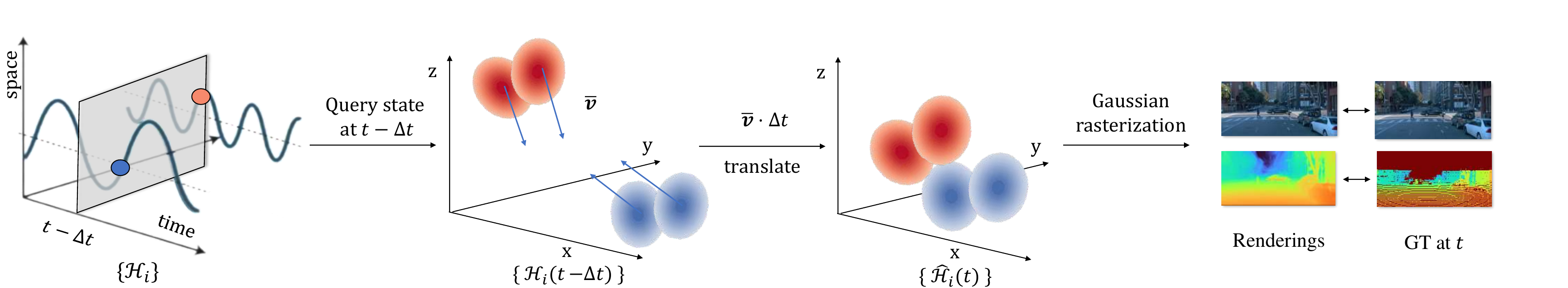}
  \caption{
 \rebuttal{Our temporal smoothing mechanism. We query the status of PVG point set at $t-\Delta t$ and translate each point with its 3D flow translation $\bm{\Bar{v}}\cdot \Delta t $, we further render the translated set of points to do the supervision at timestamp $t$.
  }}  \label{fig:flow based training}
\end{figure}

\noindent{\bf Temporal smoothing by intrinsic motion}
Reconstructing dynamic scenes in autonomous driving poses a significant challenge, primarily attributed to the sparse data in terms of both views and timestamps, as well as the unconstrained variations across frames. Specifically, in PVG, individual points encompass only a narrow time window, resulting in constrained training data and an increased susceptibility to overfitting. To address this, we capitalize on the inherent dynamic properties of PVG, which establish connections between the states of consecutive observations.

Instead of flow estimation, we introduce the \textbf{\textit{average velocity}} metric as:
\begin{equation}
        \bm{\Bar{v}} = \frac{\mathrm{d} \widetilde{\bm{\mu}}(t)}{\mathrm{d} t} \bigg|_{t=\tau}\cdot \exp(-\frac{\rho}{2})= \bm{v}\cdot \exp(-\frac{\rho}{2}).
\end{equation}
The intuition comes from the average velocity weighted by opacity decay
\begin{equation}
        \bm{\Bar{v}} \propto \frac{1}{\sqrt{2\pi}\beta}\int^{+\infty}_{-\infty} \frac{\mathrm{d} \widetilde{\bm{\mu}}(t)}{\mathrm{d} t} \cdot e^{-\frac{1}{2}(t-\tau)^2 \beta^{-2}} \mathrm{d} t.
\end{equation}
This metric is bounded as it satisfies that $
    \lim_{\rho \to \infty} \bm{\Bar{v}} = \mathbf{0}$, and $\lim_{\rho \to 0} \bm{\Bar{v}} = \bm{v}$.

In practical scenarios, dynamic objects often maintain a constant speed within a short time interval. This observation leads to the emergence of a linear relationship between consecutive states of PVG. 

Formally, consider two adjacent timestamps, $t_1$ and $t_2$ ($t_1 < t_2$) with their respective states represented as $\{\mathcal{H}_i(t_1)\}$ and $\{\mathcal{H}_i(t_2)\}$.
These states are linearly connected by a \textbf{\textit{flow translation}} for each point, denoted as $\Delta \bm{\mu} = \bm{\Bar{v}} \cdot (t_2 - t_1) = \bm{\Bar{v}} \Delta t$. Specifically, we estimate the underlying state of ${\mathcal{H}}(t_2)$ as:
\begin{equation}
\widehat{\mathcal{H}}(t_2) = \{\widetilde{\bm{\mu}}(t_1) + \bm{\Bar{v}} \cdot \Delta t, \bm{q}, \bm{s}, \widetilde{o}(t_1), \bm{c}\}.
\end{equation}
We note that this estimation process is applied to each individual PVG point. A visual representation of this estimation is illustrated in Fig.~\ref{fig:flow based training}.

We utilize estimated states to improve model training. Specifically, we assign a probability of $\eta$ to set $\Delta t$ as 0 (indicating no estimation), and for the remaining probability, we randomly sample $\Delta t$ from a uniform distribution $\mathrm{U}(-\delta, +\delta)$. In the latter case, we replace $\mathcal{H}$ (Eq. \eqref{eq: render shooting gaussian}) with $\widehat{\mathcal{H}}$ during training.
This strategy helps each point to learn its correct motion trend from the adjacent training frames which acts like a \textbf{\textit{self-supervision}} mechanism, fostering a more consistent representation without imposing a significant increase in computational demands as well as allowing eliminate the dependence on optical flow estimation.
By adopting this approach, we improve temporal coherence and consistency, thereby alleviating the challenges posed by sparse data and the risk of overfitting.

\noindent{\bf Sky refinement}
\rebuttal{Representing the static sky using Gaussian points is theoretically possible, but in practice, this would require placing them at extremely large distances with extremely large scales, which causes optimization challenge. To address this, we adopt a cube map representation, where the sky color depends solely on the viewing direction. This approach is physically reasonable, lightweight, and does not affect rendering speed.}
Specifically, we utilize a high-resolution learnable environment cube map $f_{sky}(d)=c_{sky}$ as the background. The final color is articulated as $C_f=C + (1-O)f_{sky}(d)$, where $O=\sum^{N_H}_{i=1} T_i\alpha_i$ represents the rendered opacity. During the training phase, we incorporate random perturbations to the ray direction $d$ within its unit pixel length to enhance anti-aliasing.

\noindent{\bf Objective}
Our objective loss function is formulated as:
\begin{equation}
\begin{aligned}
       \mathcal{L} = (1-\lambda_{r})\mathcal{L}_1 + \lambda_r \mathcal{L}_{\mathrm{ssim}} + \lambda_d \mathcal{L}_d +
       \lambda_{o} \mathcal{L}_{o} + \lambda_{\bm{\Bar{v}}}\mathcal{L}_{\bm{\Bar{v}}},
\end{aligned}
\end{equation}
where $\mathcal{L}_1$ and $\mathcal{L}_{\mathrm{ssim}}$ are L1 and SSIM loss \cite{kerbl20233d} for supervision of RGB rendering.

The term $\mathcal{L}_d = \frac{1}{hw}\sum \|\mathcal{D}^{s}-\mathcal{D}\|_1$ is a depth loss for geometry awareness, where $\mathcal{D}^{s}$ is a sparse inverse depth map generated by projecting the LiDAR points to the camera plane, $\mathcal{D}$ denotes the inverse of the rendered depth map, and $h$ and $w$ denote the rendering spatial size.

The term $\mathcal{L}_{o}=-\frac{1}{hw}\sum O\cdot\log O-\frac{1}{hw}\sum M_{sky}\cdot\log(1-O)$ is the opacity loss
where $M_{sky}$ is the sky mask estimated by a pretrained segmentation model~\cite{xie2021segformer}.
This loss aims to drive the opacity values towards either 0 (representing a transparent sky) or 1. Specifically, it regularizes opacity to 0 for predicted sky pixels.

The last term $\mathcal{L}_{\bm{\Bar{v}}} = \frac{1}{hw}\sum \|\mathcal{\Bar{V}}\|_1$ is the sparse velocity loss where $\mathcal{\Bar{V}}$ is the rendered average velocity $\bm{\Bar{v}}$ map. This loss not only leads to a sparse $\|\bm{v}\|$ but also encourages larger $\beta$ (corresponding to static scene components). 
The rationale behind this is that most elements of a scene are static.

\section{Experiments}

\noindent{\bf Competitors}
 For dynamic scenes, we evaluate our method alongside S-NeRF~\cite{xie2023s}, StreetSurf~\cite{guo2023streetsurf}, Mars~\cite{wu2023mars}, 3DGS~\cite{kerbl20233d}, NSG~\cite{ost2021neural}, SUDS~\cite{turki2023suds} and EmerNeRF~\cite{yang2023emernerf}. 
In line with previous methods, we conduct evaluations on both image reconstruction and novel view synthesis tasks, selecting every fourth timestamp from each camera as the test set for novel view synthesis. Although our primary focus is on dynamic scenes, to be fair, we also provide a quantitative comparison with methods~\cite{xie2023s,guo2023streetsurf,kerbl20233d} tailored for static scenes. This is to demonstrate our model’s superiority at uniformly managing both static and dynamic environments.
 \rebuttal{Furthermore, we also compare PVG with concurrent approaches~\cite{yan2024street,chen2024omnire, yang2024deformable, zhou2024hugs, yang2024real} based on bounding boxes or 3D Gaussian Splatting. The comparison includes both the reconstruction quality of the static background and the dynamic foreground.} 
 
\noindent{\bf Implementation details}
For points initialization, we sample $6\times 10^5$ LiDAR points, $2\times 10^5$ near points whose distance to the origin is uniformly sampled from $(0,r)$, $2\times 10^5$ far points whose inverse distance is uniformly sampled from $(0,1/r)$, where $r$ is the foreground radius different across scenes, around 30 meters. $\beta$ is set to $0.3$, $\bm{v}$ is set to $\mathbf{0}$. 
We employ the Adam optimizer~\cite{kingma2014adam} and maintain a similar learning rate for most parameters as the original 3DGS implementation while we adjust the learning rate of the velocity $\bm{v}$, opacity decaying $\beta$ and opacity $o$ to $1\times10^{-3}$, $0.02$ and $0.005$ respectively. Regarding the densification schedule, we set the image space densification threshold to $1.7\times10^{-4}$ and reset the opacity of Gaussians to $0.01$ every $3,000$ iterations to remove superfluous points. For regularization, we use coefficients $\lambda_r=0.2$, $\lambda_d=0.1$,  $\lambda_{o}=0.05$ and $\lambda_{\bm{\Bar{v}}}=0.01$. For Temporal smoothing training mechanism, we uniformly sample time intervals $\Delta t$ from a 1.5 frame span in the camera sequence with a probability $\eta=0.5$. We set the cube map resolution to $1024$ to capture the high-frequency details in the sky. The training process commences from a downsampled scale of $16$, which is then gradually increased every $5,000$ iterations. We conduct all experiments on a single NVIDIA RTX A6000 GPU for a total of $30,000$ iterations which takes about an hour to yield the final results, and the rendering speed can achieve 50 FPS. We rescale the time interval between two consecutive frames to 0.02 and fix $l=0.2$. As shown in Fig. \ref{fig:teaser}, \textbf{\em we remove the PVG points whose $\rho < 1$ to preserve the static part of the scene}.

\noindent \textbf{Metrics }
Consistent with SUDS~\cite{turki2023suds}, we use the PSNR, SSIM, and LPIPS metrics to measure image reconstruction and novel view synthesis.

\begin{figure}[t]
\centering
\begin{subfigure}[b]{0.19\linewidth}
\begin{tabular}{c}
   \includegraphics[width=\linewidth]{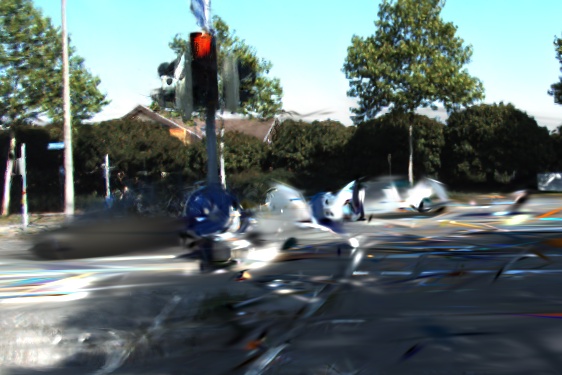}\\\includegraphics[width=\linewidth]{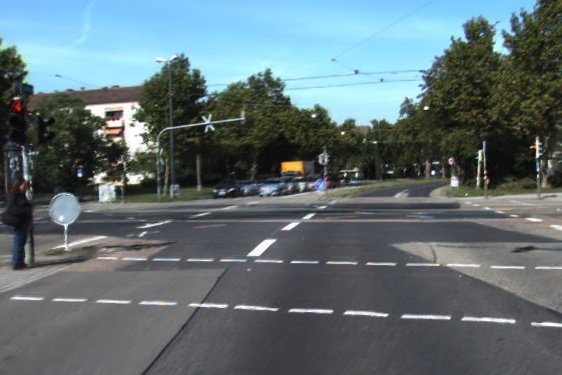}
\end{tabular}
\caption{3DGS~\cite{kerbl20233d}}
\end{subfigure}
\begin{subfigure}[b]{0.19\linewidth}
\begin{tabular}{c}
   \includegraphics[width=\linewidth]{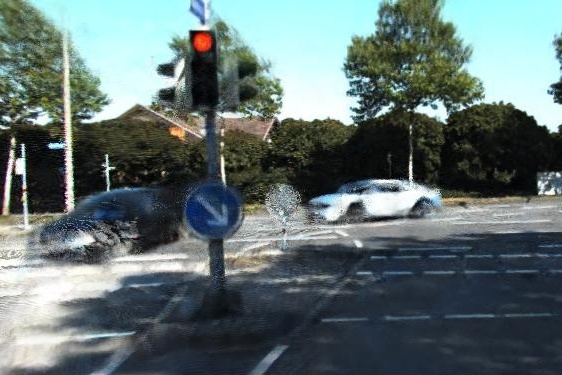}\\\includegraphics[width=\linewidth]{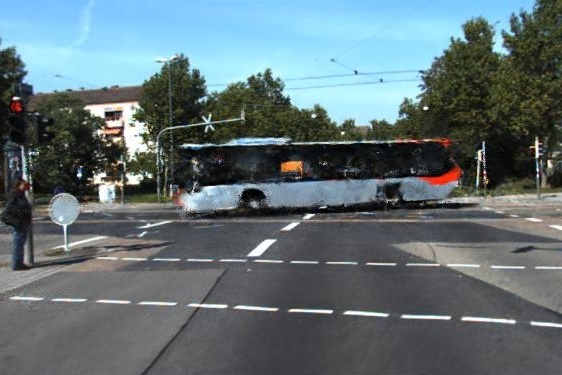}
\end{tabular}
\caption{SUDS~\cite{turki2023suds}}
\end{subfigure}
\begin{subfigure}[b]{0.19\linewidth}
\begin{tabular}{c}
   \includegraphics[width=\linewidth]{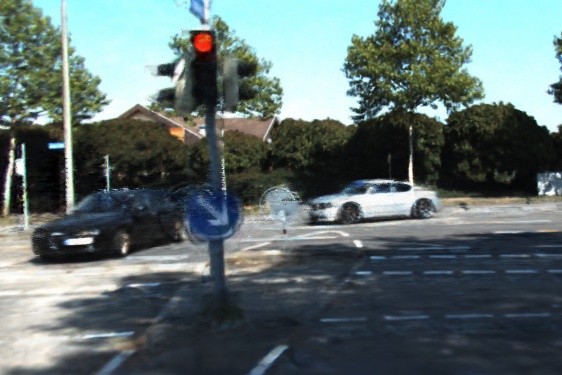}\\\includegraphics[width=\linewidth]{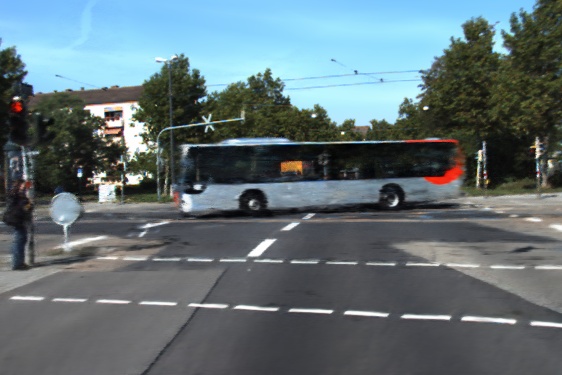}
\end{tabular}\hspace{-1mm}
\caption{\tiny{EmerNeRF}~\cite{yang2023emernerf}}
\end{subfigure}
\begin{subfigure}[b]{0.19\linewidth}
\begin{tabular}{c}
   \includegraphics[width=\linewidth]{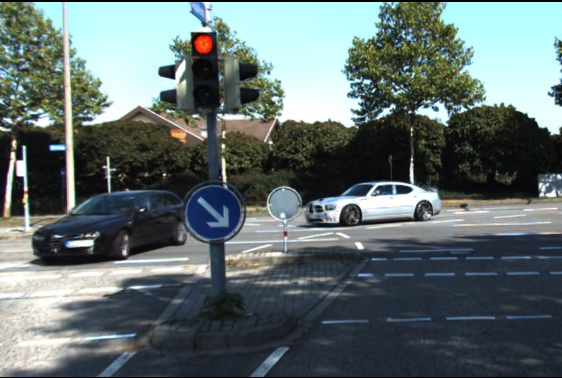}\\\includegraphics[width=\linewidth]{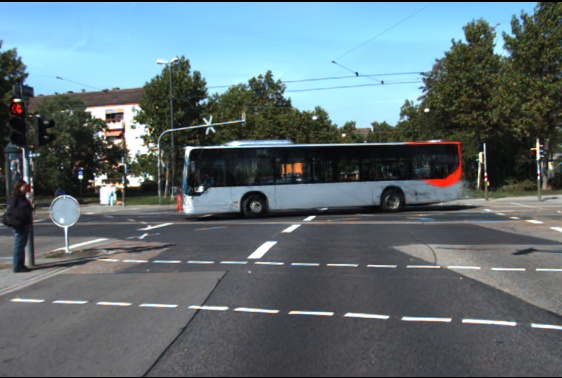}
\end{tabular}
\caption{PVG (Ours)}
\end{subfigure}
\begin{subfigure}[b]{0.19\linewidth}
\begin{tabular}{c}
   \includegraphics[width=\linewidth]{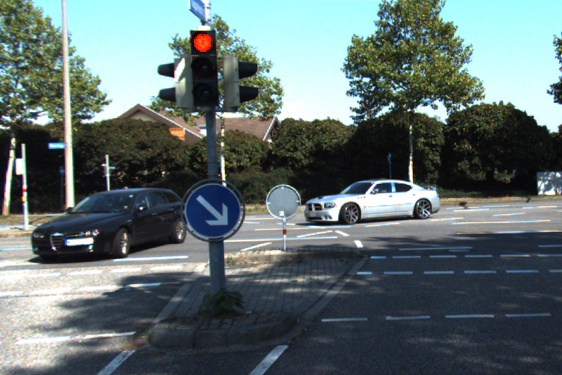}\\\includegraphics[width=\linewidth]{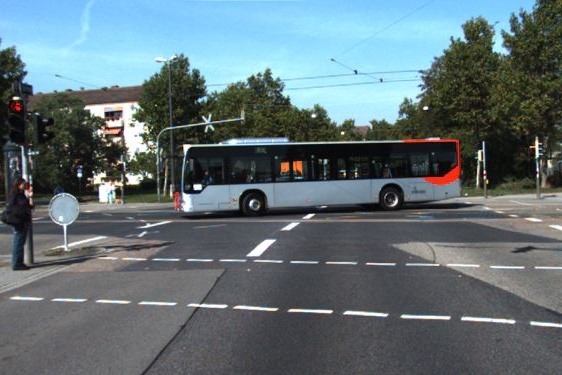}
\end{tabular}
\caption{GT}
\end{subfigure}
\caption{Novel view synthesis of dynamic scenes on KITTI.}
\label{fig:kitti_dynamic}
\end{figure}

\subsection{Comparison with state of the art}
\label{sec:experiment waymo kitti}

\makeatletter
\newcommand{\thickhline}{%
    \noalign{\ifnum0=`}\fi\hrule height 1.2pt
    \futurelet\reserved@a\@xhline}
\makeatother

\begin{table*}[ht]

\caption{Quantitative results on dynamic scenes on Waymo and KITTI. The bold text denotes the best result.
}
\scalebox{0.93}{
\begin{tabular}{r|c|ccc|ccc|}
&\multicolumn{7}{c|}{Waymo Open Dataset }
\\
& \multicolumn{1}{c}{}& \multicolumn{3}{c}{Image reconstruction } &  \multicolumn{3}{c|}{Novel view synthesis}\\
& FPS & PSNR $\uparrow$ & SSIM $\uparrow$ & LPIPS$\downarrow$ & PSNR $\uparrow$ & SSIM $\uparrow$ & LPIPS$\downarrow$
\\ \hline
S-NeRF~\cite{xie2023s}& 0.0014 & 19.67 & 0.528 & 0.387 & 19.22 & 0.515 & 0.400 \\
StreetSurf~\cite{guo2023streetsurf}& 0.097 & 26.70&0.846 &0.3717 &23.78 &0.822 &0.401\\

3DGS~\cite{kerbl20233d}& \textbf{63} & 27.99&0.866& 0.293&25.08 &0.822 &0.319\\

NSG~\cite{ost2021neural}& 0.032 & 24.08 & 0.656 & 0.441 & 21.01 & 0.571 & 0.487
\\
 Mars~\cite{wu2023mars}& 0.030 & 21.81 & 0.681 & 0.430 & 20.69 & 0.636 &	0.453 \\ 
 SUDS~\cite{turki2023suds}& 0.008 & 28.83 & 0.805 & 0.289 & 21.83 & 0.656  & 0.405
 \\ 
 EmerNeRF~\cite{yang2023emernerf}& 0.053 & 28.11 & 0.786 & 0.373 & 25.92 & 0.763 &	0.384\\ 
 \hline
\bf PVG (Ours) & 50 & \textbf{32.46} & \textbf{0.910} & \textbf{0.229} &\textbf{28.11} & \textbf{0.849} & \textbf{0.279} 
\\
\bf PVG (5cam)& 41 & \textbf{32.05} & \textbf{0.895} & \textbf{0.252} &\textbf{27.50} & \textbf{0.828} & \textbf{0.312}
\\
\multicolumn{8}{c}{} \\

 &  \multicolumn{7}{c|}{KITTI} \\
 &  \multicolumn{1}{c}{} &\multicolumn{3}{c}{Image reconstruction}  &  \multicolumn{3}{c|}{Novel view synthesis} \\
 & FPS & PSNR $\uparrow$ & SSIM $\uparrow$ & LPIPS$\downarrow$ & PSNR $\uparrow$ & SSIM $\uparrow$ & LPIPS$\downarrow$
\\ \hline
S-NeRF~\cite{xie2023s}& 0.0075 &19.23 & 0.664 & 0.193 & 18.71 & 0.606 & 0.352 \\
StreetSurf~\cite{guo2023streetsurf}&0.37& 24.14&0.819 &0.257 &22.48 &0.763 &0.304 \\

3DGS~\cite{kerbl20233d}&\textbf{125}&21.02 &0.811 &0.202 &19.54 &0.776 &0.224 \\

NSG~\cite{ost2021neural}& 0.19 & 26.66 & 0.806 & 0.186 & 21.53 & 0.673 & 0.254
\\
 Mars~\cite{wu2023mars}& 0.31 & 27.96 & 0.900 & 0.185 & 24.23 & 0.845 & 0.160 \\ 
 SUDS~\cite{turki2023suds}&0.04& 28.31 & 0.876 & 0.185 & 22.77 & 0.797 & 0.171
 \\ 
 EmerNeRF~\cite{yang2023emernerf}& 0.28 & 26.95 & 0.828 & 0.218 & 25.24 & 0.801 & 0.237 \\ 
 \hline
\bf PVG (Ours) &59& \textbf{32.83}& \textbf{0.937} & \textbf{0.070} & \textbf{27.43} & \textbf{0.896} & \textbf{0.114}
 
\end{tabular}
}
\label{tab:experiment}
\end{table*}

\noindent{\bf Results on Waymo}
The Waymo Open Dataset encompasses over 1,000 driving segments, each with a duration of 20 seconds, recorded using five high-resolution LiDARs and five cameras facing the front and sides.

In our experiments, we utilize the three frontal cameras to assess performance on four challenging dynamical scenes (each contains around 50 frames), chosen due to their substantial movement. Table~\ref{tab:experiment} displays the average error metrics across these selected dynamic scenes for both image reconstruction and novel view synthesis tasks. Our model markedly outperforms baselines~\cite{xie2023s,guo2023streetsurf,kerbl20233d,ost2021neural,turki2023suds,wu2023mars} across all metrics in both tasks. Specifically, for the image reconstruction task, we note a $12.6\%$ increase in PSNR, $13.0\%$ in SSIM, and a $20.8\%$ decrease in LPIPS compared to the leading SUDS~\cite{turki2023suds} baseline. For novel view synthesis, our technique synthesizes high-quality views of unseen timestamps and also significantly surpasses the best EmerNeRF~\cite{yang2023emernerf} performance by $8.4\%$ in PSNR, $11.3\%$ in SSIM, and $27.3\%$ in LPIPS. We show more visualization in Fig.~\ref{fig:supp_waymo_reconstruction} and~\ref{fig:supp_waymo_novel}. Noteworthy is our method's efficiency, completing the entire training process around an hour, and stands out as the method with the fastest training speed compared to the baselines. We also report training results of our method on fully 5 cameras, for we use every forth frame of 5 cameras as our testing set, our performance dropped off slightly. Our rendering speed, measured in Frames Per Second (FPS), significantly outperforms competing methods and stands slightly below that of 3DGS~\cite{kerbl20233d}. 

\begin{figure}[htbp]
    \vspace{-3mm}
    \centering
    \begin{minipage}[t]{0.96\textwidth}
        \begin{tabular}{cc}
            \raisebox{0.1\textwidth}[0pt][0pt]{\rotatebox[origin=c]{90}{PVG}}
            &\hspace{-3mm}\includegraphics[width=0.9\textwidth]{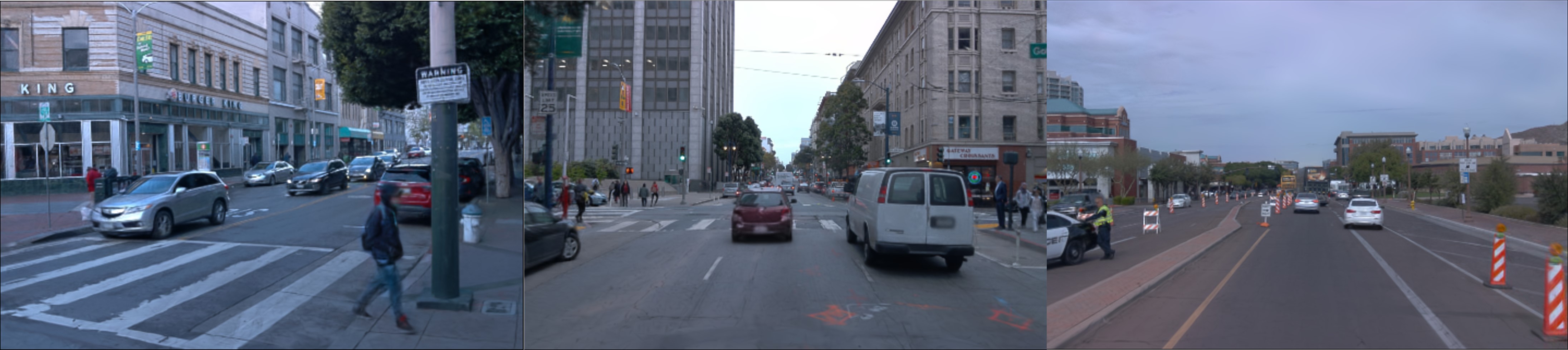}\\
            \raisebox{0.1\textwidth}[0pt][0pt]{\rotatebox[origin=c]{90}{EmerNeRF}}
            &\hspace{-3mm}\includegraphics[width=0.9\textwidth]{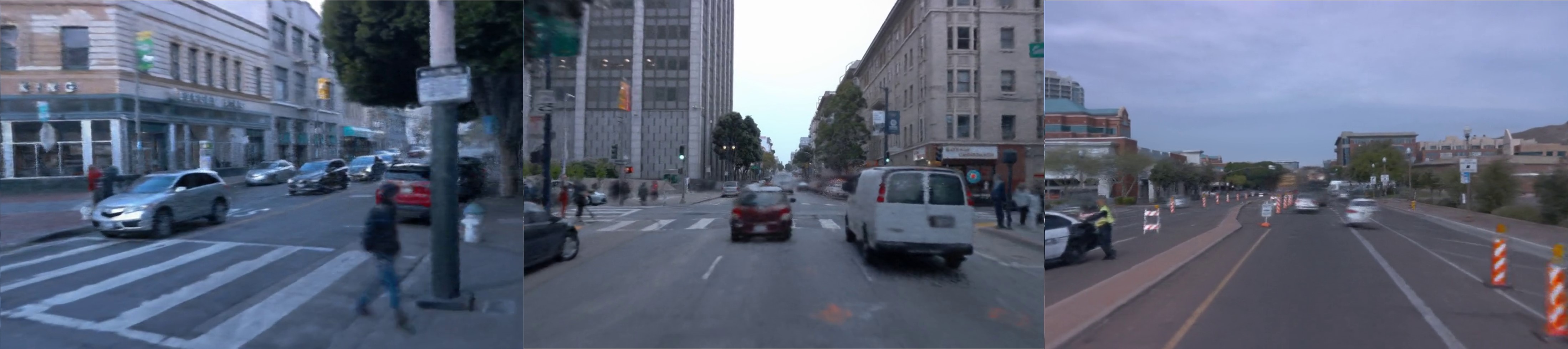}\\
            \raisebox{0.1\textwidth}[0pt][0pt]{\rotatebox[origin=c]{90}{GT}}
            &\hspace{-4mm} \includegraphics[width=0.9\textwidth]{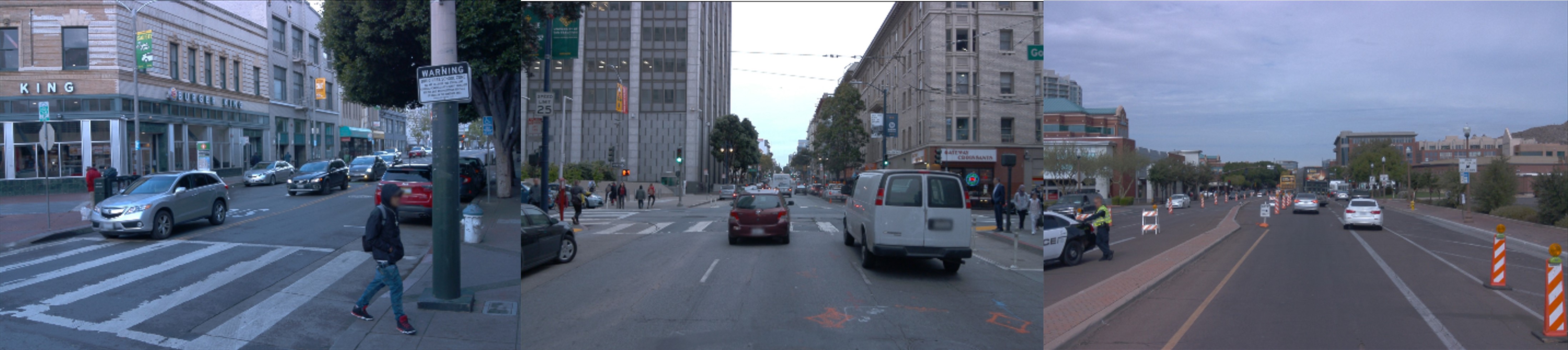}\\
        \end{tabular}
        \captionsetup{font=scriptsize}
        \caption{PVG vs. EmerNeRF~\cite{yang2023emernerf}.}
        \label{fig:emernerf}
    \end{minipage}
    \\
    \begin{minipage}[t]{0.96\textwidth}
        \begin{tabular}{cc}
            \raisebox{0.1\textwidth}[0pt][0pt]{\rotatebox[origin=c]{90}{RGB}}
            & \hspace{-3mm}\includegraphics[width=0.9\textwidth]{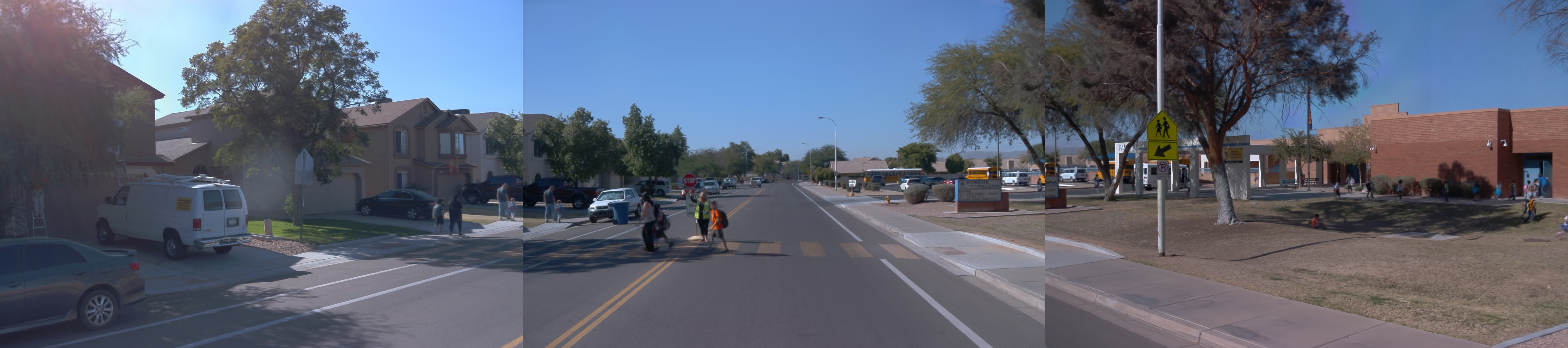}\\
            \raisebox{0.1\textwidth}[0pt][0pt]{\rotatebox[origin=c]{90}{Depth}}
            & \hspace{-3mm}\includegraphics[width=0.9\textwidth]{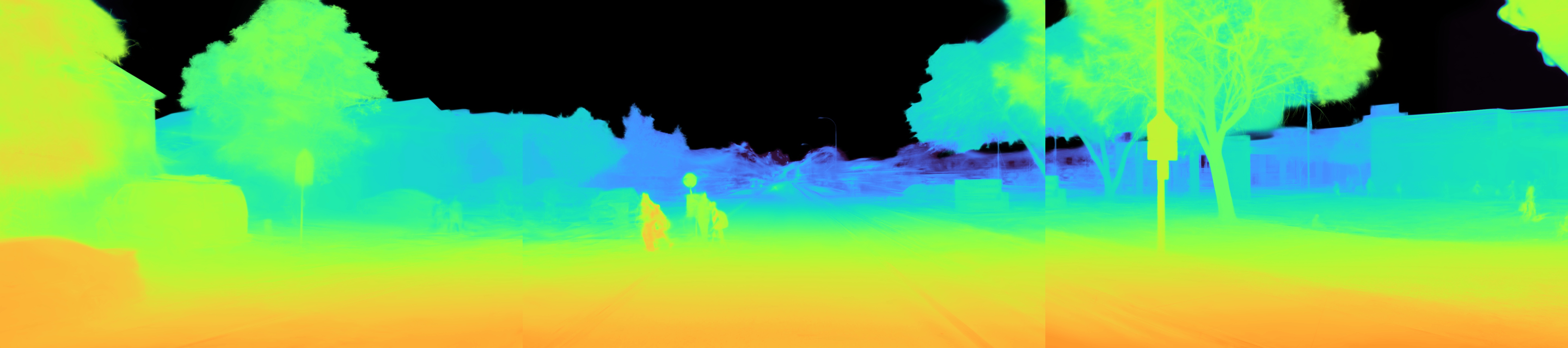}\\
            \raisebox{0.1\textwidth}[0pt][0pt]{\rotatebox[origin=c]{90}{Semantic}}
            & \hspace{-3mm}\includegraphics[width=0.9\textwidth]{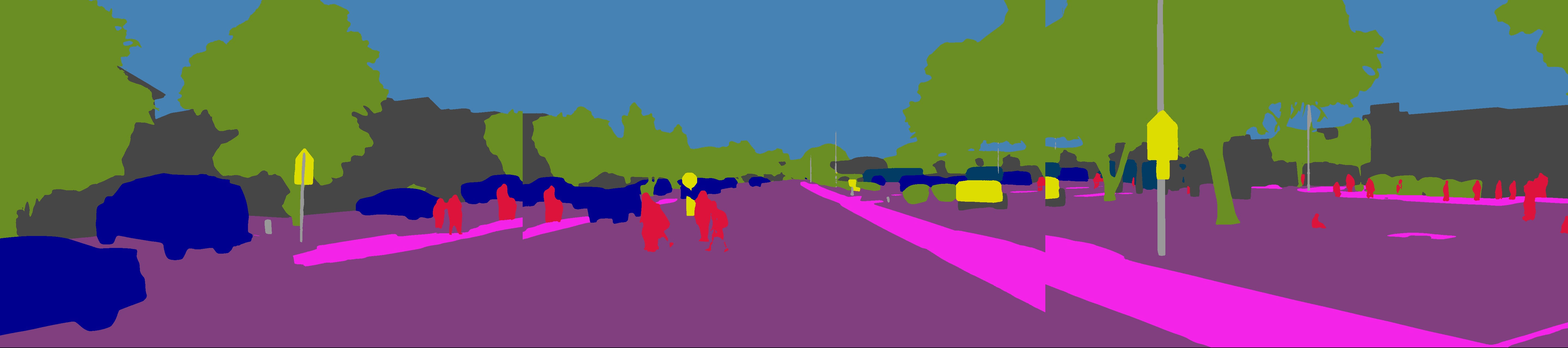}\\
        \end{tabular}
        \captionsetup{font=scriptsize}
        \caption{Rendered RGB image, depth and semantic label.}
        \label{fig:segmentation}
    \end{minipage}
\vspace{-5mm}
\end{figure}

For fair comparison with EmerNeRF \cite{yang2023emernerf}, we adopt its training setup, i.e., 4$\times$ downsample rate for image resolution and using the whole sequences for training.  We randomly selected four Waymo scenes for test in Table~\ref{tab:dynamic waymo emernerf}. It is shown that PVG is clearly superior over EmerNeRF. More visualization is given in Fig. \ref{fig:emernerf}.

\begin{figure}[t]
\centering
\begin{subfigure}[b]{0.19\linewidth}
    \includegraphics[width=\linewidth]{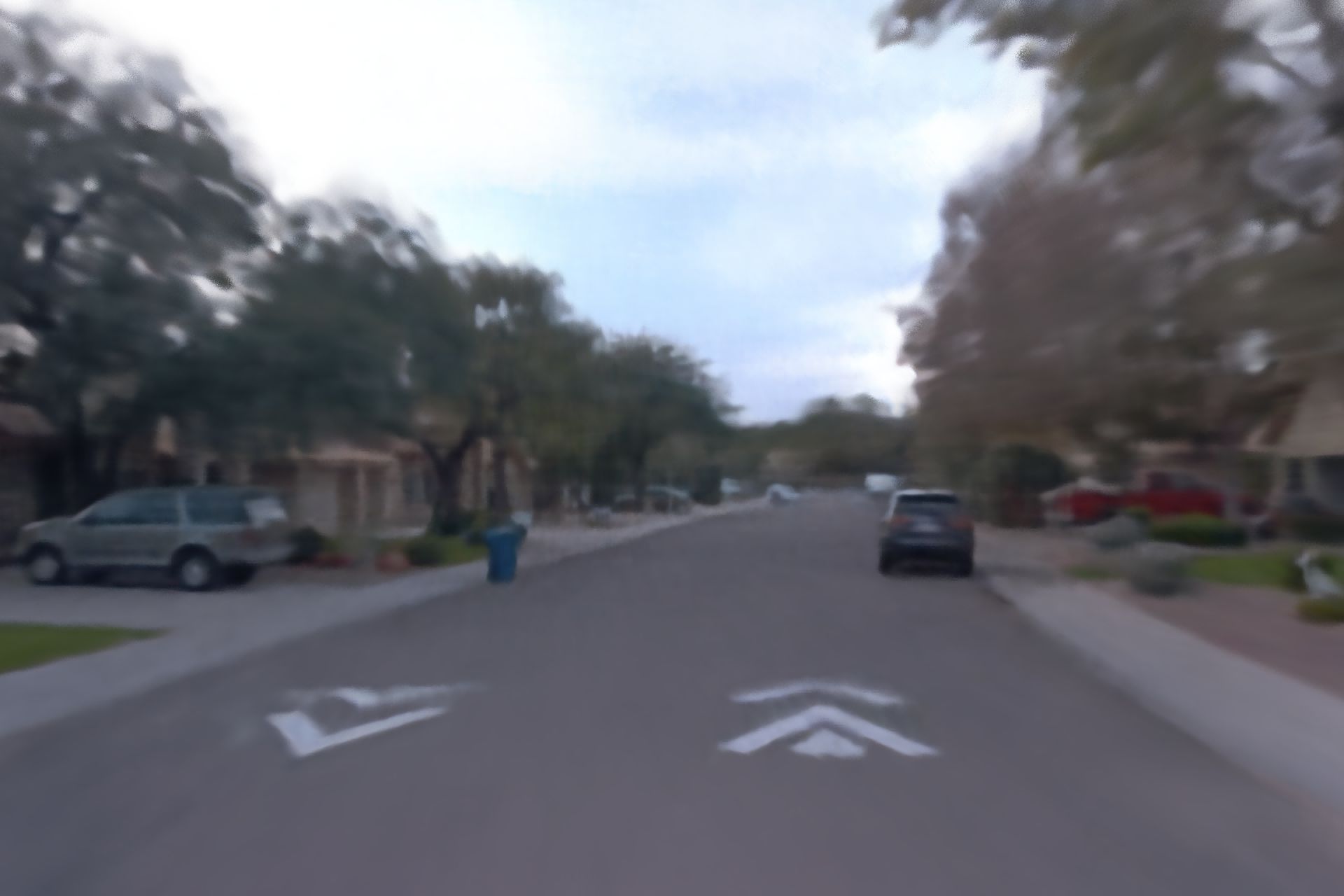}
    \caption{S-NeRF~\cite{xie2023s}}
\end{subfigure}
\begin{subfigure}[b]{0.19\linewidth}
    \includegraphics[width=\linewidth]{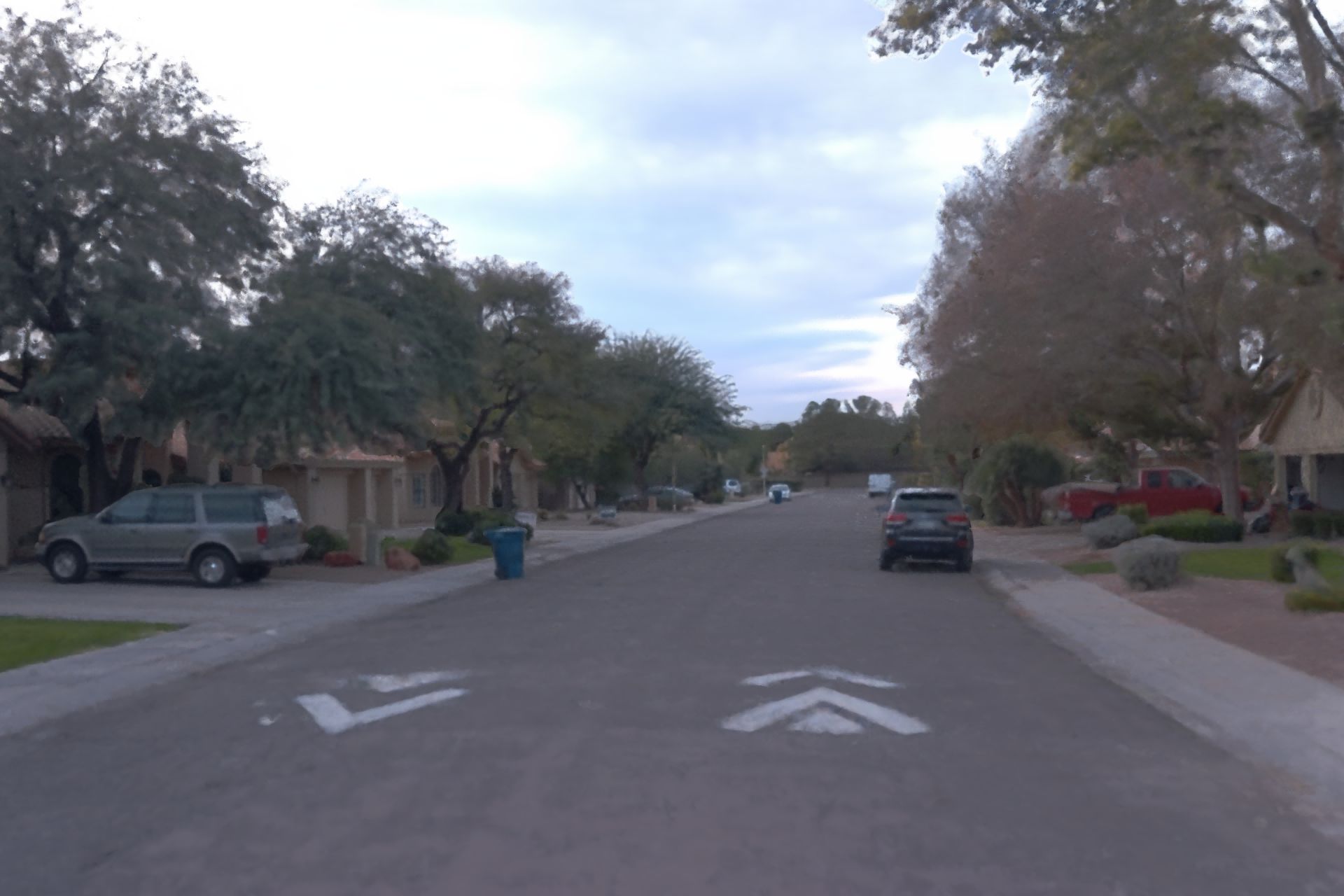}
    \caption{StreetSurf~\cite{guo2023streetsurf}}
\end{subfigure}
\begin{subfigure}[b]{0.19\linewidth}
    \includegraphics[width=\linewidth]{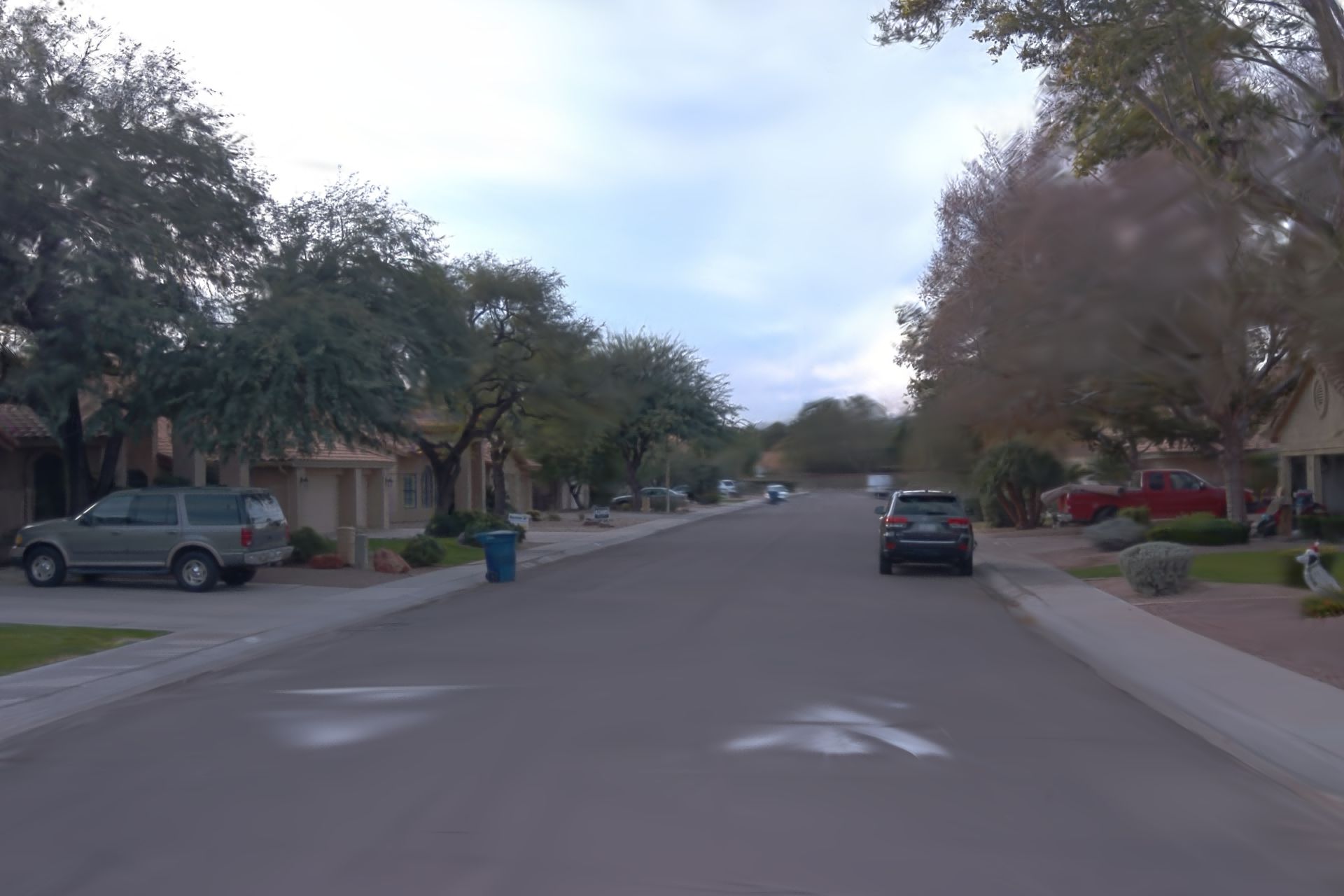}
    \caption{3DGS~\cite{kerbl20233d}}
\end{subfigure}
\begin{subfigure}[b]{0.19\linewidth}
    \includegraphics[width=\linewidth]{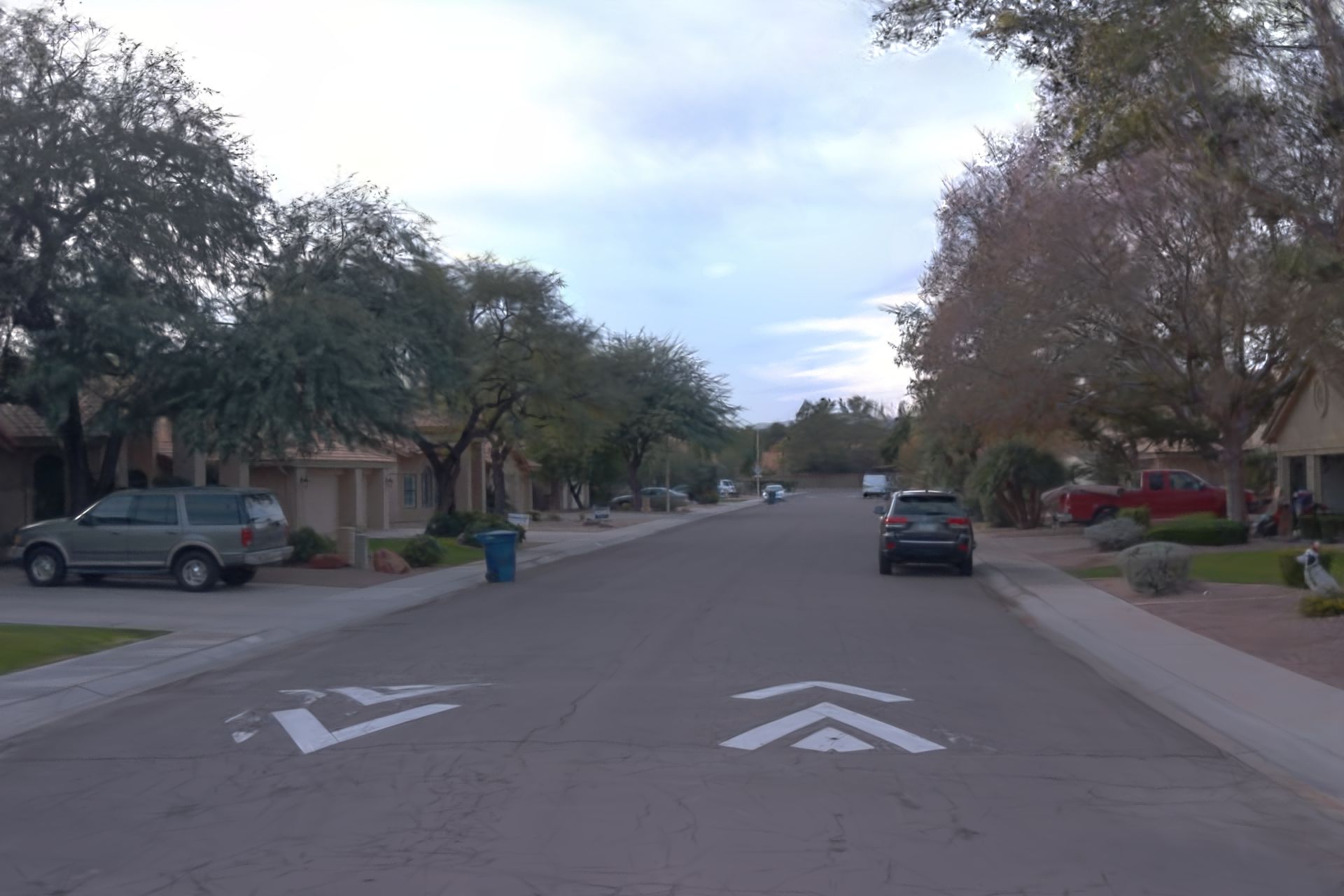}
    \caption{PVG (Ours)}
\end{subfigure}
\begin{subfigure}[b]{0.19\linewidth}
    \includegraphics[width=\linewidth]{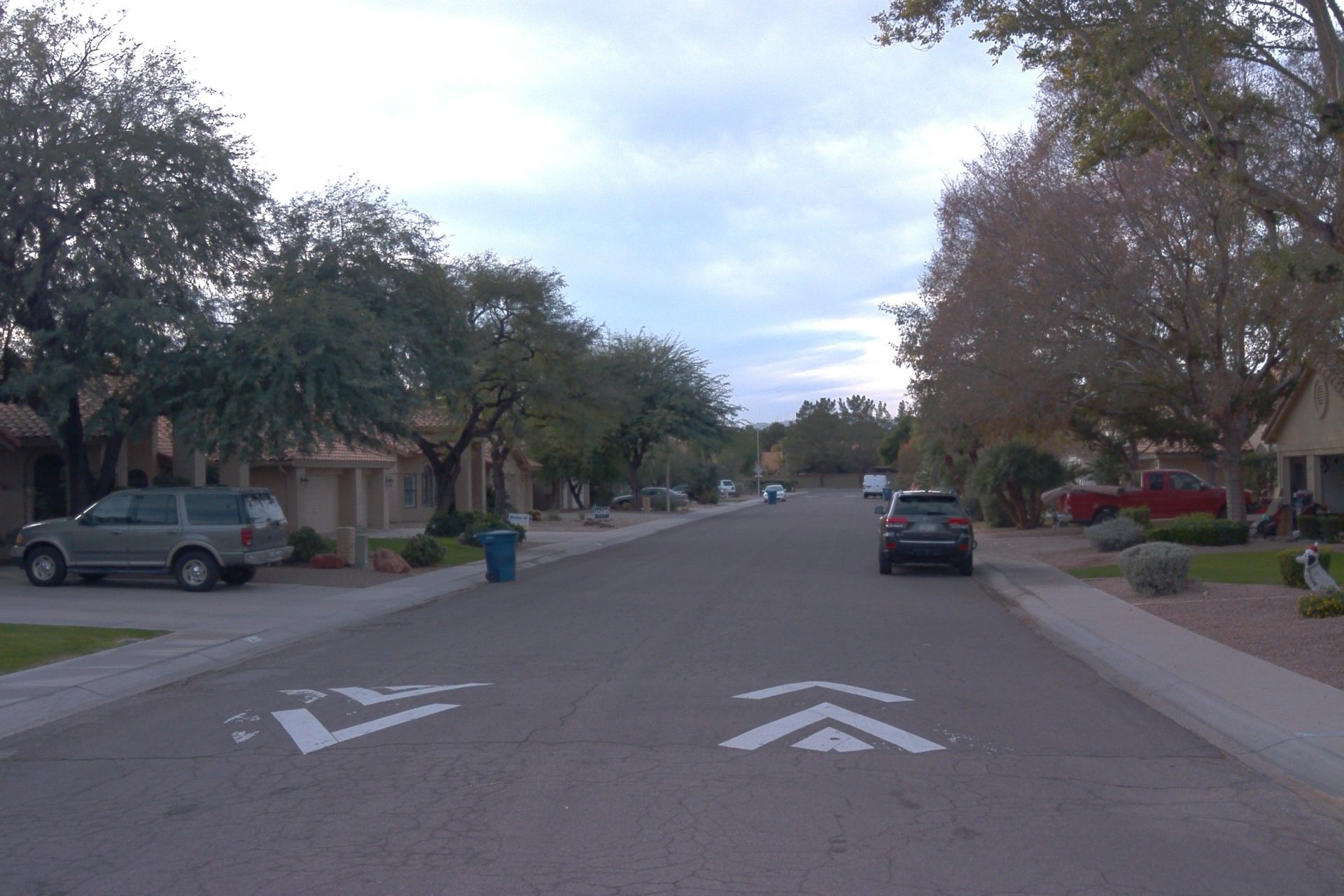}
    \caption{GT}
\end{subfigure}
\caption{Novel view synthesis of a static scene on Waymo.}
\label{fig:waymo_static}
\end{figure}

\begin{table*}
    \centering
    \begin{minipage}[ht]{0.59\textwidth}
        \caption{Dynamic scenes' PSNR, same settings as EmerNeRF~\cite{yang2023emernerf}.}
        \scalebox{0.74}{
        
        \begin{tabular}{lcccc}
            \toprule
            Sequence & \multicolumn{2}{c}{Image Reconstruction} & \multicolumn{2}{c}{Novel View Synthesis} \\
            \cmidrule(lr){2-5}
            & EmerNeRF & PVG (Ours)& EmerNeRF & PVG (Ours) \\
            \midrule
            seg1023...  & 28.93&\textbf{34.15}&28.64&\textbf{29.52} \\
            seg1171...  & 28.42&\textbf{32.34}&26.29&\textbf{27.16} \\
            seg1039... & 29.10&\textbf{32.01}&27.73&\textbf{28.47}\\
            seg1096...  & 28.39&\textbf{31.46} & 27.35&\textbf{28.28}\\
            \midrule
            Average  & 28.69 & \textbf{32.49}& 27.50 & \textbf{28.35} \\
            \bottomrule
        \end{tabular}
        }
        \label{tab:dynamic waymo emernerf}
    \end{minipage}
    \hfill
    \begin{minipage}[ht]{0.37\textwidth}
        \caption{Static scenes' PSNR, same settings as StreetSurf~\cite{guo2023streetsurf}.}
        \scalebox{0.74}{
        \begin{tabular}{lcc}
            \toprule
            Sequence & \multicolumn{2}{c}{Image Reconstruction} \\
            \cmidrule(lr){2-3}
            & StreetSurf & PVG (Ours) \\
            \midrule
            seg1534... &27.26 & \textbf{31.22} \\
            seg4058... &28.08 & \textbf{34.91} \\
            seg3425... & 29.42 & \textbf{32.37} \\
            seg1347... & 28.20 & \textbf{31.11} \\
            \midrule
            Average & 28.24 & \textbf{32.40} \\
            \bottomrule
        \end{tabular}
        }
        \label{tab:static waymo streetsurf}
    \end{minipage}
\end{table*}

For static scenes, as shown in Table~\ref{tab:static waymo snerf}, we align with the settings in S-NeRF~\cite{xie2023s}, employing all five cameras for training and designating every fourth timestamp's frame as the test set. The training sequences utilized are the same as those reported in S-NeRF \cite{xie2023s}. 
Our model not only outperforms the baselines~\cite{xie2023s,guo2023streetsurf,kerbl20233d} which are focused on static scenes across three metrics but also demonstrates a discernible enhancement in image quality, as evidenced in Fig.~\ref{fig:waymo_static}. To be more convinced, we conduct additional experiments using the same setup as in StreetSurf~\cite{guo2023streetsurf}, using the randomly selected four sequences from the results reported in StreetSurf \cite{guo2023streetsurf}. Our evaluation, focusing on PSNR, as StreetSurf only reports PSNR in their papers, reveals in Table~\ref{tab:static waymo streetsurf} that our model achieves significant advancements over previous approaches dedicated to static scenes.

We show that by making use of the 2D semantic labels provided in a video sequence, our method can derive semantic categories in novel viewpoints or timestamps and render as shown in Fig.~\ref{fig:segmentation}. This is achieved by assigning a 19-dimension vector for each Gaussian, representing the probability of which category the Gaussian belongs to and the rendered probability map is supervised by 2D semantic labels with cross-entropy.

\noindent{\bf Results on KITTI}
Our approach is also quantitatively evaluated on the KITTI benchmark, following SUDS~\cite{turki2023suds}. We select sequences characterized by extensive movement for analysis. The proposed method surpasses all competitors across every evaluated metric. Notably, while these sequences present a substantial number of dynamic objects, our temporal smoothing mechanism secures a concise scene representation, thereby mitigating over-fitting and ensuring superior image quality in novel viewpoints. Table~\ref{tab:experiment} and Fig.~\ref{fig:kitti_dynamic} demonstrate our significant improvement over the leading SUDS~\cite{turki2023suds} in image reconstruction, with improvements of $16.0\%$ in PSNR, $7.0\%$ in SSIM, and a $62.2\%$ decrease in LPIPS and surpass the leading EmerNeRF~\cite{yang2023emernerf} in novel view synthesis with improvement of $8.7\%$ in PSNR, $11.9\%$ in SSIM, and a $51.9\%$ decrease in LPIPS.  More visualization results can be seen in Fig.~\ref{fig:supp_kitti_reconstruction} and~\ref{fig:supp_kitti_novel}.
3DGS achieves much inferior results on KITTI as there were  much more dynamic objects in the scenes than Waymo.

\begin{figure}[t]
    \centering
    \includegraphics[width=\linewidth]{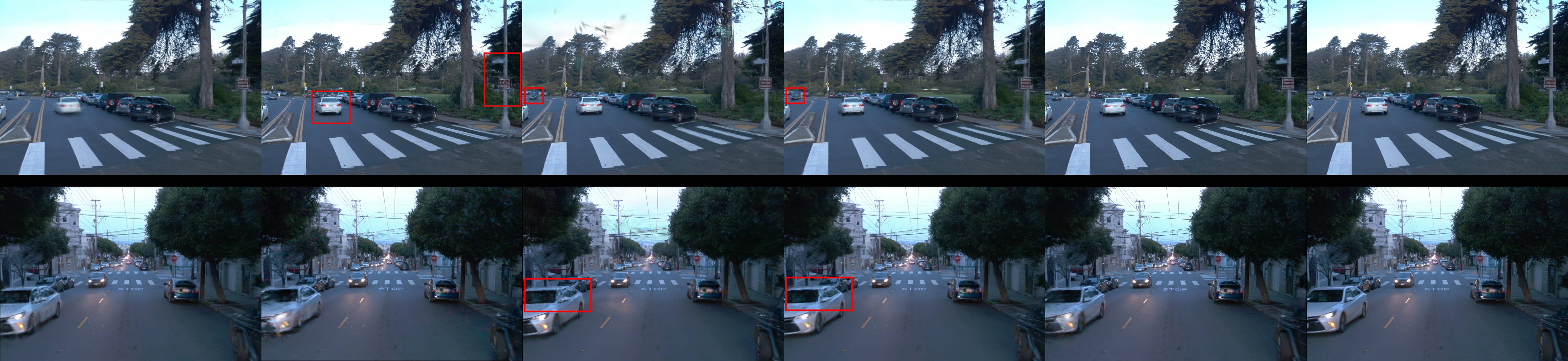}
    \par\vspace{1.5pt}
\resizebox{\textwidth}{!}{%
\begin{tabular}{*{6}{>{\centering\arraybackslash}p{3.5cm}}}
(a) DeformableGS~\cite{yang2024deformable} & (b) 4DGS~\cite{yang2024real} & (c) StreetGS~\cite{yan2024street} & (d) OmniRe~\cite{chen2024omnire} & (e) PVG (Ours) & (f) GT \\
\end{tabular}
}
    \caption{\rebuttal{Novel view synthesis of two scenes on the Waymo dataset.}}
    \label{fig:waymo_dynamic}
\end{figure}

\begin{table*}[t]
\centering
\caption{\rebuttal{
Comparison with Gaussian Splatting models on the Waymo dataset. 
We report average PSNR, SSIM, and LPIPS for both full images and dynamic vehicle regions. 
For the pixel-level metric PSNR, results are also separated for static (S) and dynamic (D) parts of the scene. 
Following OmniRe~\cite{chen2024omnire}, all evaluations use a resolution of 960×640. The running speed is tested using a NVIDIA RTX A6000 GPU. 
{\texttt{Box}: The need for object bounding boxes for model training.}
}}
\resizebox{\textwidth}{!}{%
\begin{tabular}{r|>{\centering\arraybackslash}p{0.45cm}|>{\centering\arraybackslash}p{1.1cm}|>{\centering\arraybackslash}p{0.7cm}|>{\centering\arraybackslash}p{0.9cm}>{\centering\arraybackslash}p{0.9cm}>{\centering\arraybackslash}p{0.9cm}>{\centering\arraybackslash}p{1.45cm}>{\centering\arraybackslash}p{1.45cm}|>{\centering\arraybackslash}p{0.9cm}>{\centering\arraybackslash}p{0.9cm}>{\centering\arraybackslash}p{0.9cm}>{\centering\arraybackslash}p{1.45cm}>{\centering\arraybackslash}p{1.45cm}}
 &  \multicolumn{3}{c}{}& \multicolumn{5}{c|}{Image reconstruction}& \multicolumn{5}{c}{Novel view synthesis}\\
  &Box& Time &FPS$\uparrow$& PSNR$\uparrow$ & SSIM$\uparrow$ & LPIPS$\downarrow$ & PSNR(S)$\uparrow$ & PSNR(D)$\uparrow$ & PSNR$\uparrow$ & SSIM$\uparrow$ & LPIPS$\downarrow$ & PSNR(S)$\uparrow$ & PSNR(D)$\uparrow$\\

\hline

StreetGaussian~\cite{yan2024street} & \checkmark & Jan 24&67.8&32.18 &0.918 &0.090 & 32.68 &28.64 &28.92 &0.877& 0.110 & 29.38& 25.54 \\
HUGS~\cite{zhou2024hugs} & \checkmark& Mar 24 & 46.1 &31.58 &0.904 &0.094 & 32.79 & 25.78& 29.34& 0.865 &0.110 &  30.42& 23.84 \\
Omnire~\cite{chen2024omnire} & \checkmark& Aug 24 & 58.3&32.32 &0.923 &0.084 & 32.92 &28.36 &29.41 &0.884& 0.101 & 29.91&\textbf{25.85}\\
\hline
4DGS~\cite{yang2024real} & $\times$& Oct 23& 67.9 &34.94 &0.935 &0.068& 35.36 &31.80 &26.56& 0.793 & 0.132& 26.87 &23.97\\
DeformableGS~\cite{yang2024deformable} &$ \times $& Sep 23 & 4.9&32.34 &0.923 &0.086 &  32.94 &27.39 &29.52& 0.889 &\textbf{0.100} & 30.37 &24.66 \\
\bf PVG (Ours) & $\times$& Nov 23& \textbf{70.4}& 
\textbf{36.02} & \textbf{0.952} & \textbf{0.059} & \textbf{36.79} & \textbf{32.17} & \textbf{30.95} & \textbf{0.897} & 0.105 & \textbf{31.72}& 25.82\\
\end{tabular}
}
\label{tab:dyn psnr}
\end{table*}

\subsection{Comparison with concurrent Gaussian Splatting models}
\rebuttal{We next conduct an extensive comparison against concurrent 3D Gaussian Splatting methods. For this experiment, we evaluated 8 scenes from StreetGaussian~\cite{yan2024street} and 4 original scenes from our paper, with each scene captured using 3 cameras at a resolution of 960×640.
The results are reported in Table~\ref{tab:dyn psnr} and Figure~\ref{fig:waymo_dynamic}.}

\rebuttal{Existing methods can be broadly categorized into two groups. 
The first are box-based methods (e.g., StreetGaussian, HUGS, Omnire), which rely on object-level supervision from 3D bounding boxes. While this approach can achieve high PSNR scores for dynamic objects, it makes the model more complex and less flexible. This reliance on automatically generated bounding boxes can introduce noisy supervision and can struggle with distant or occluded objects, where detection performance is poor. Furthermore, treating dynamic objects separately from the background can reduce scene consistency.}

\rebuttal{
The second group are box-free methods (e.g., 4DGS, DeformableGS), which are more principled as they learn scene dynamics directly without external supervision.
However, intuitive strategies like extending 3DGS to spatiotemporal 4D Gaussian primitives (4DGS) or learning a deformation field (DeformableGS) are often inferior when dealing with the fast motion and sparse views typical of urban scenes.
As our evaluation confirms, PVG achieves the best overall performance among all competitors, without reliance on object bounding boxes. The results show our superior ability to model dynamic scenes, particularly with challenging distant and occluded objects. This is because, our PVG model quantifies dynamics at the primitive level with meaningful concepts like life peak and cycle, allowing it to model both static and dynamic elements in a unified manner. This approach avoids the limitations of both box-based methods and other box-free methods.}

\begin{figure}[t]
\centering
\begin{subfigure}{0.32\textwidth}
  \includegraphics[width=\linewidth]{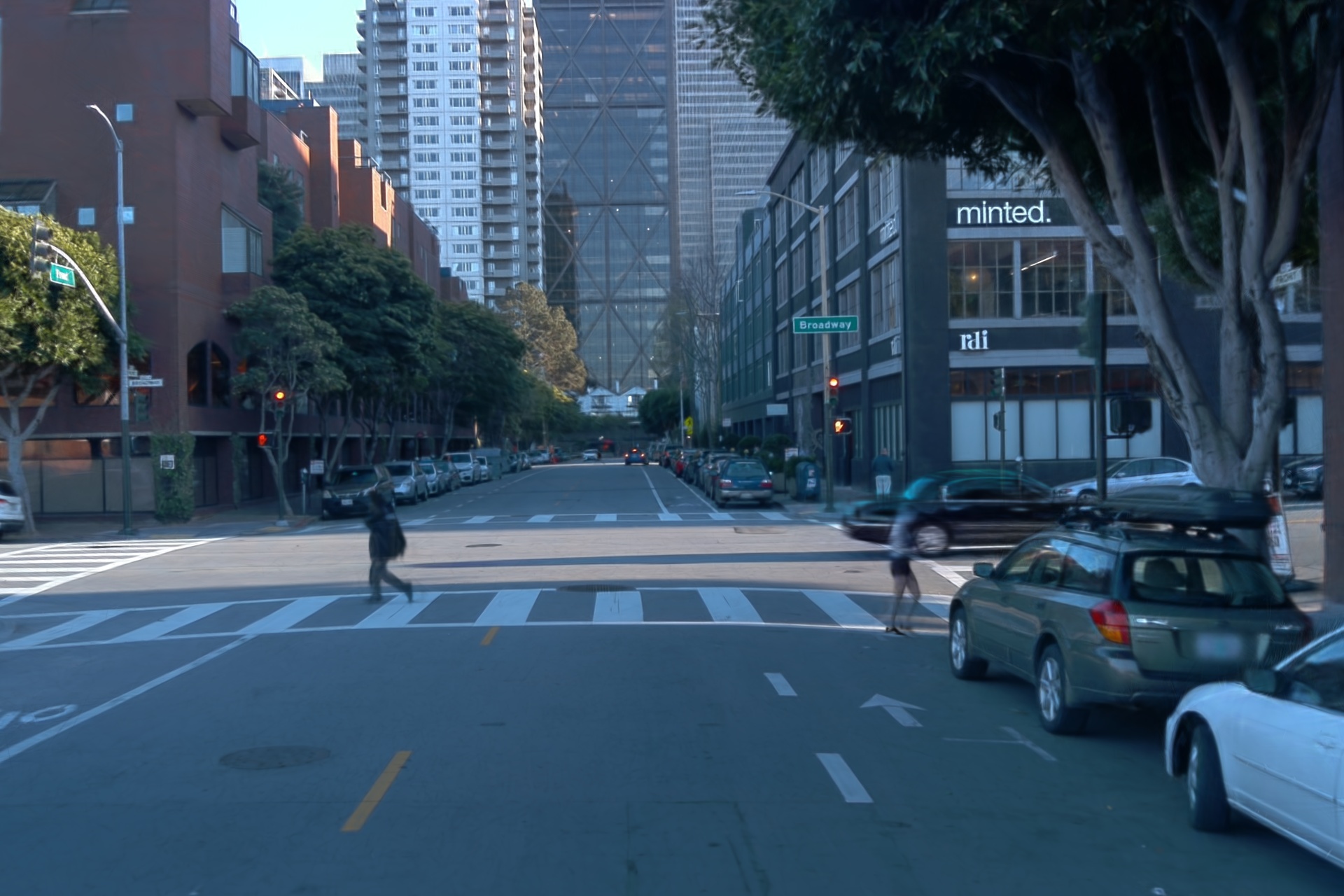}
  \caption{render RGB}
  \label{fig:supp_mvs_rgb}
\end{subfigure}
\begin{subfigure}{0.32\textwidth}
  \includegraphics[width=\linewidth]{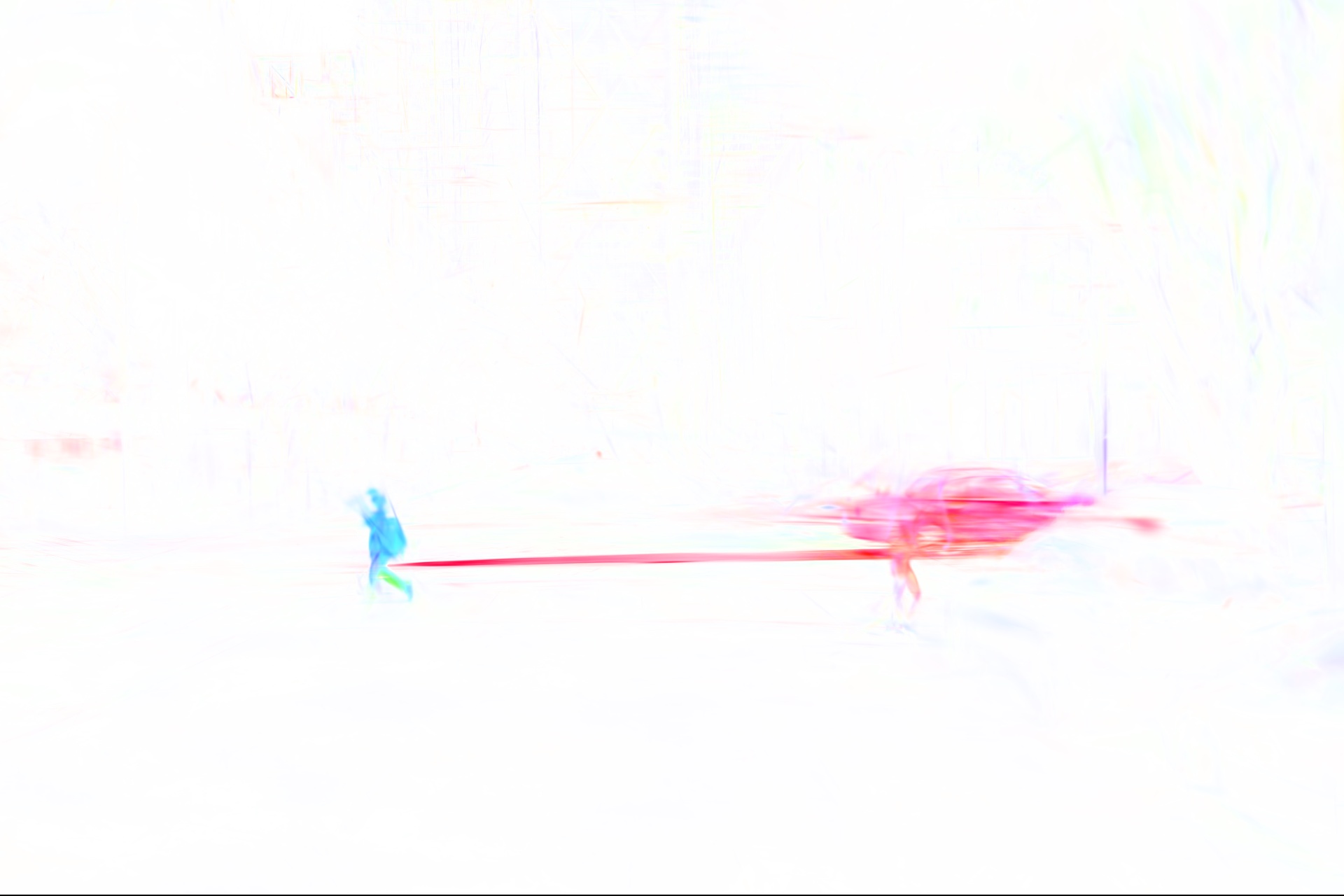}
  \caption{Velocity map}
  \label{fig:supp_velocity}
\end{subfigure}
\begin{subfigure}{0.32\textwidth}
  \includegraphics[width=\linewidth]{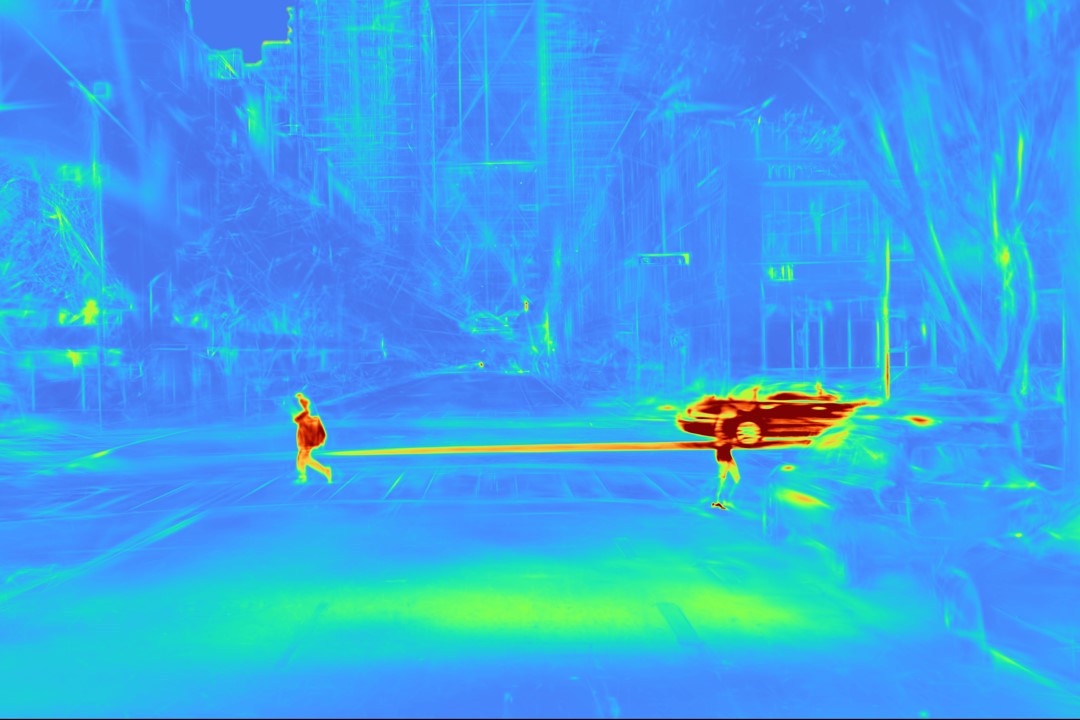}
  \caption{Staticness map}
  \label{fig:supp_mvs_rho}
\end{subfigure}

\caption{Visualization of (b) the velocity map and (c) $\rho$ map of (a) a scene with a left-to-right moving car and two walking pedestrians. It is evident that our model captures the motion, dynamic (including even the car's shadow) and static parts of the scene.
In $\rho$ map, blue/red: large/small $\rho$ pointing to static/dynamic areas.
}
\label{fig:supp_mvs}
\end{figure}

\begin{figure*}[h]
\centering

\includegraphics[width=0.3\textwidth]{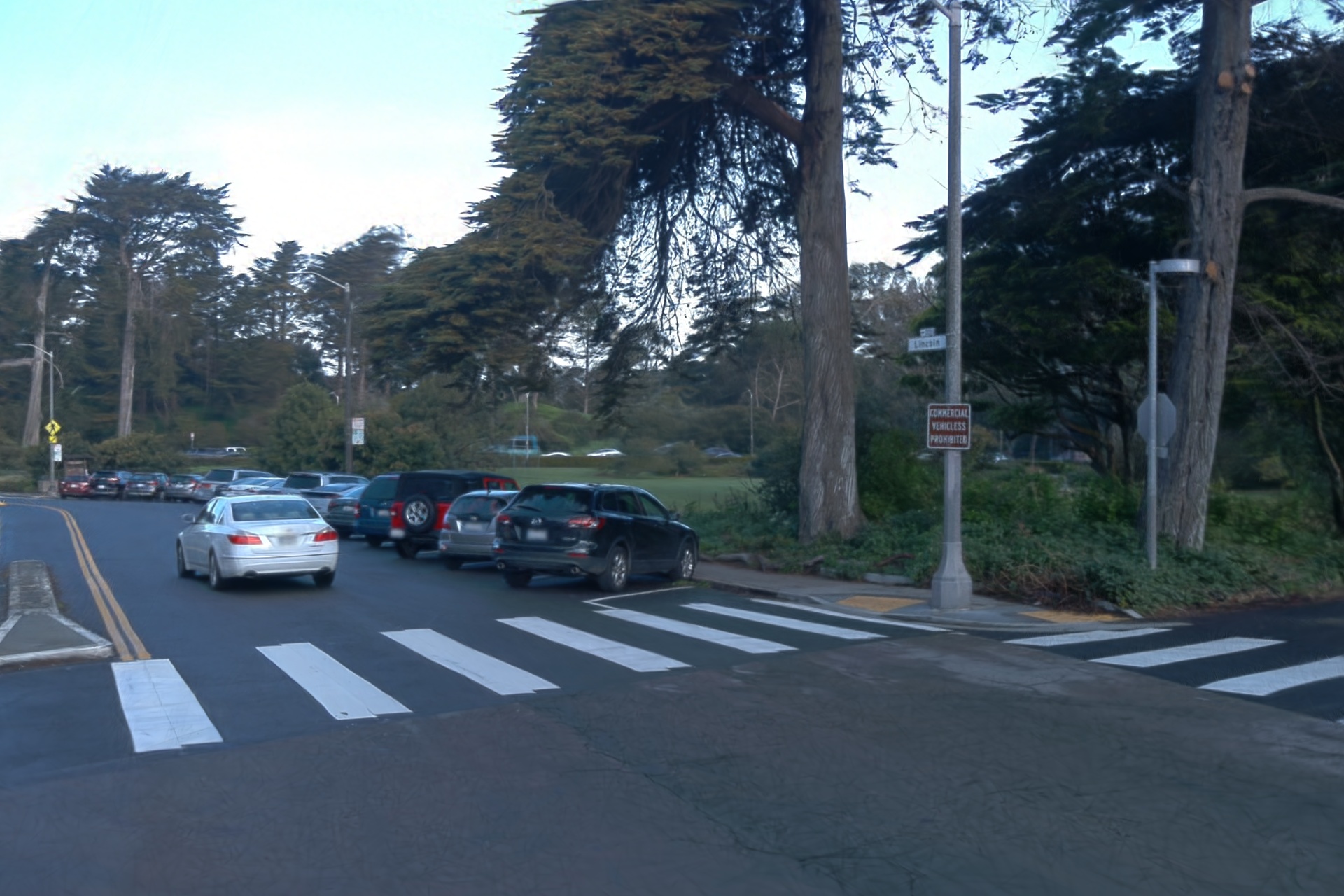}
\hspace{0.01\textwidth}
\includegraphics[width=0.3\textwidth]{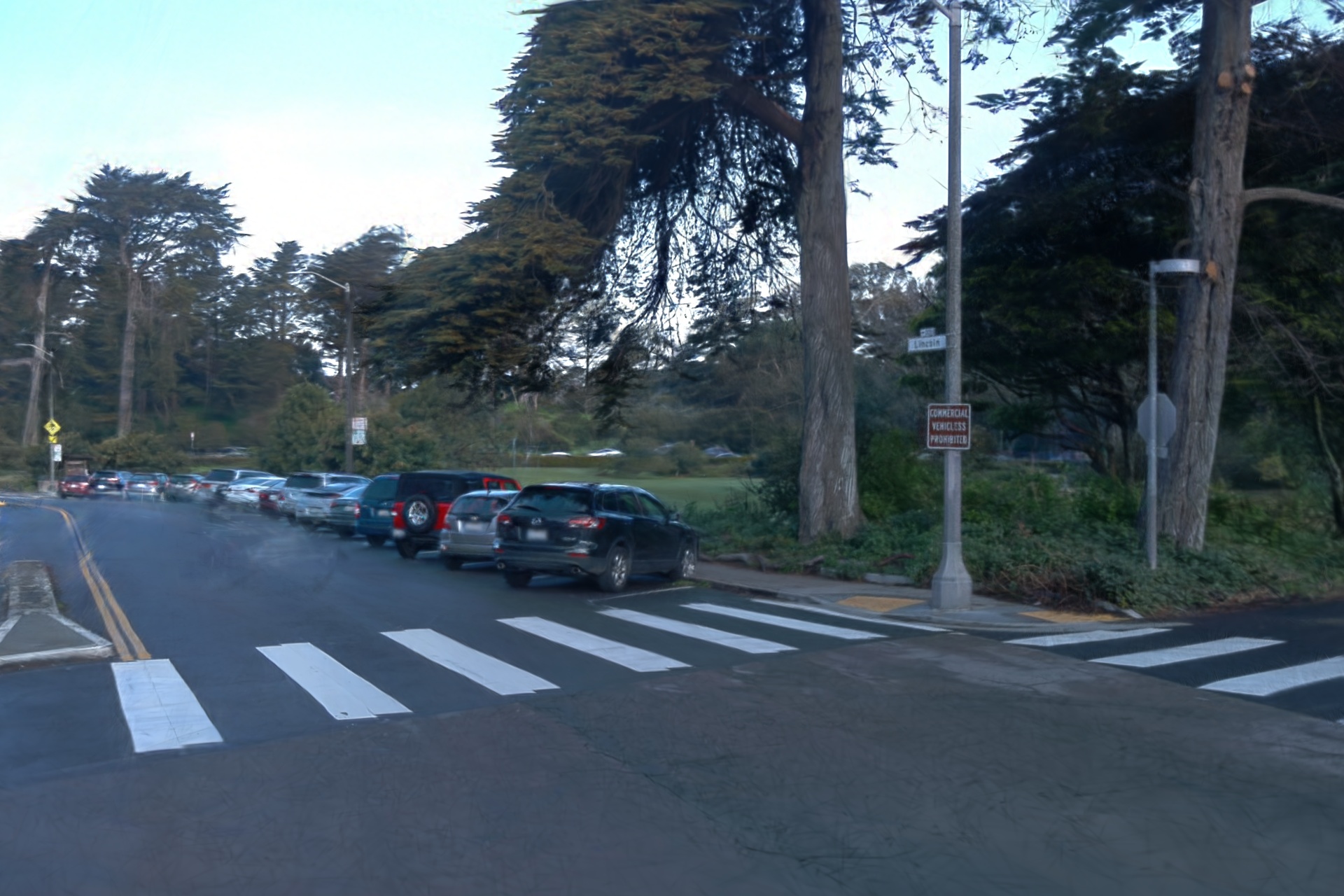}
\hspace{0.01\textwidth}
\includegraphics[width=0.3\textwidth]{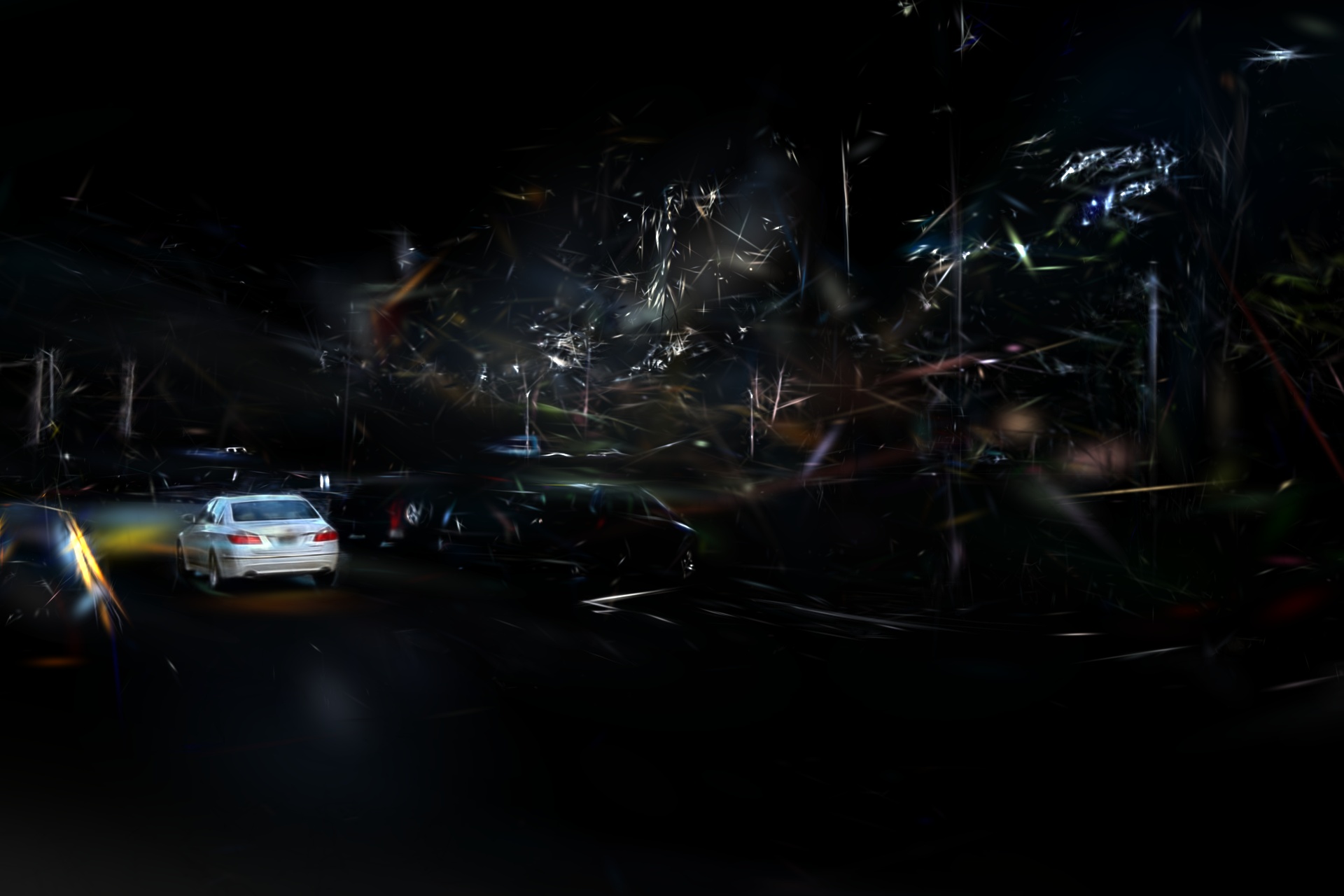}

\vspace{1em}

\includegraphics[width=0.3\textwidth]{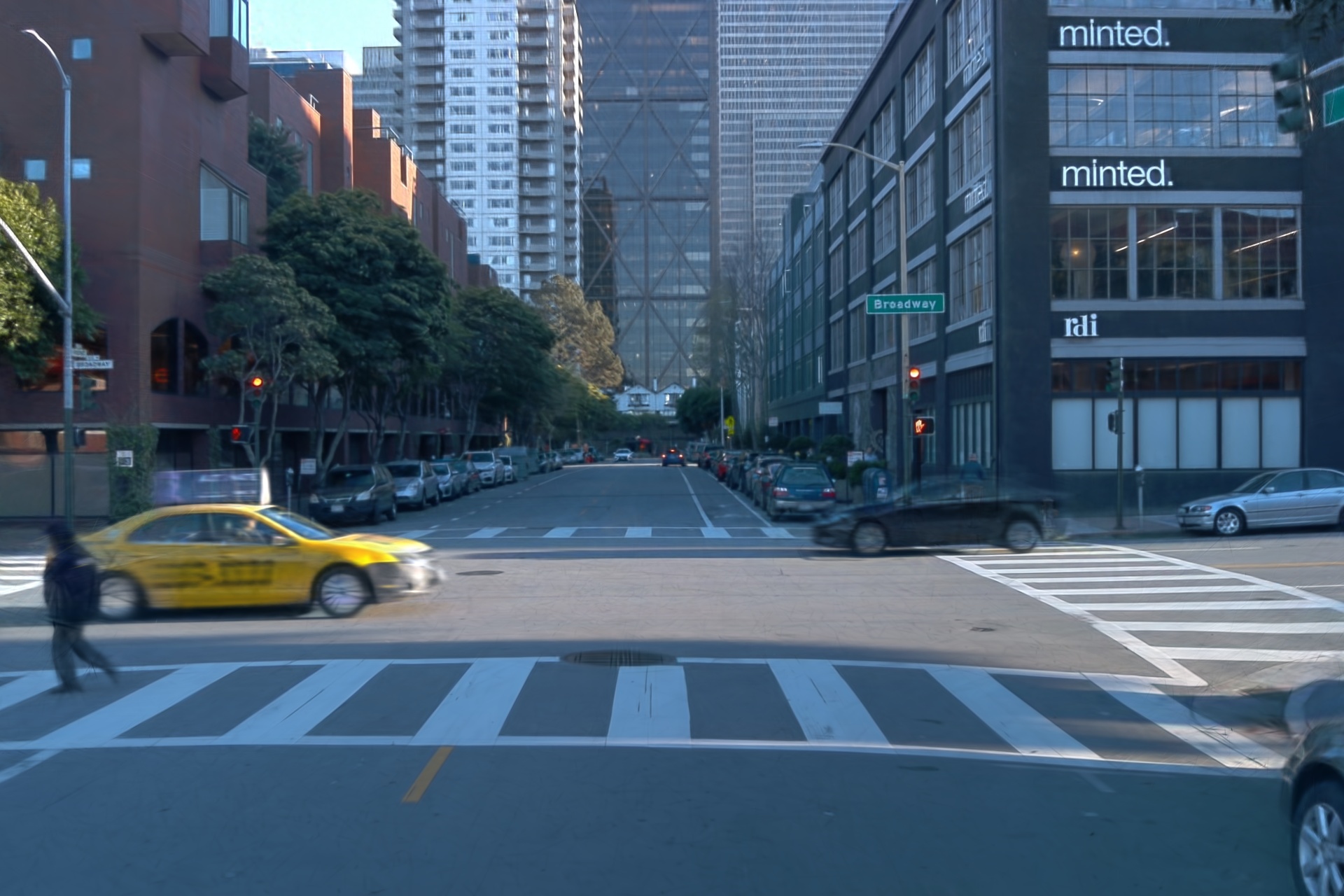}
\hspace{0.01\textwidth}
\includegraphics[width=0.3\textwidth]{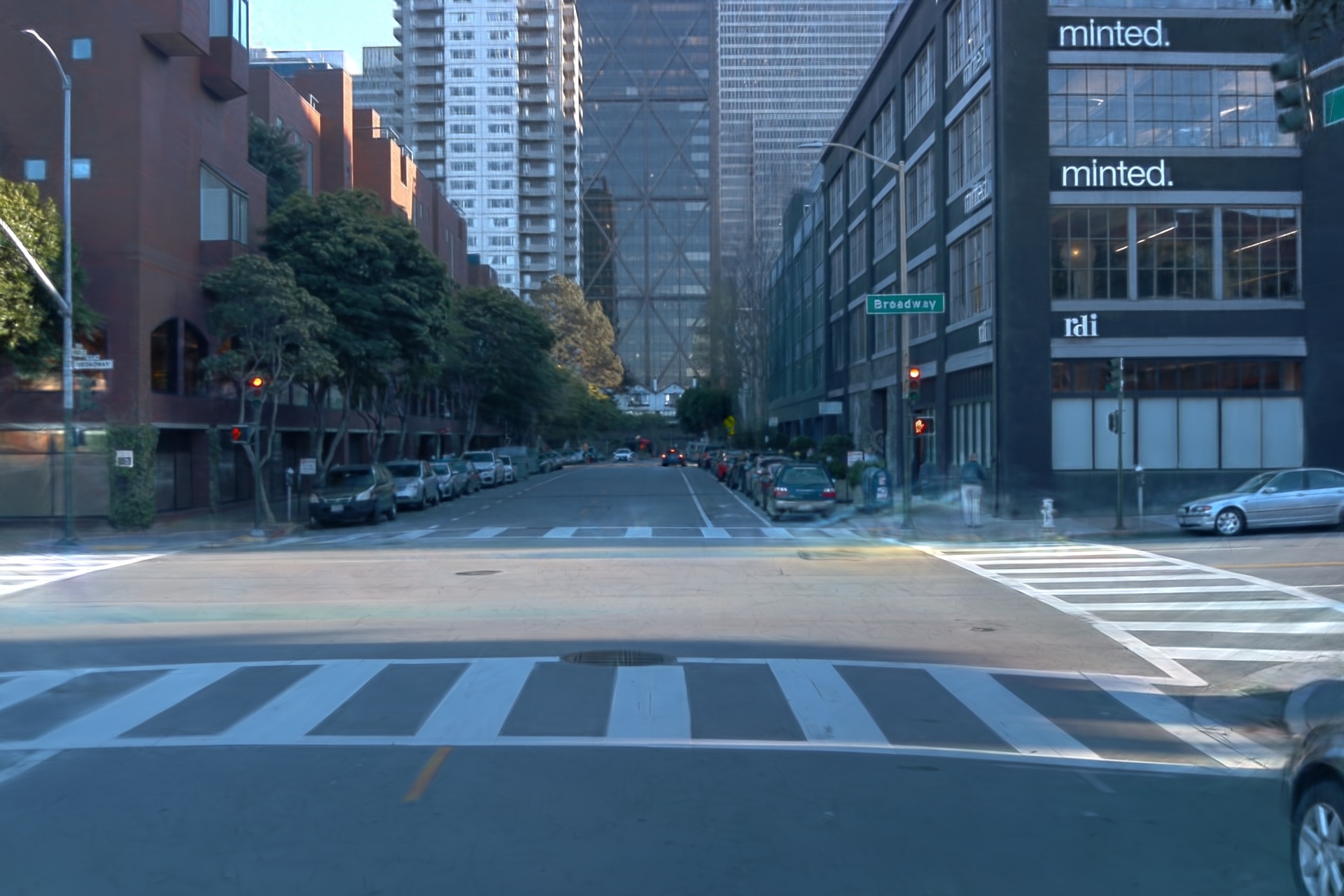}
\hspace{0.01\textwidth}
\includegraphics[width=0.3\textwidth]{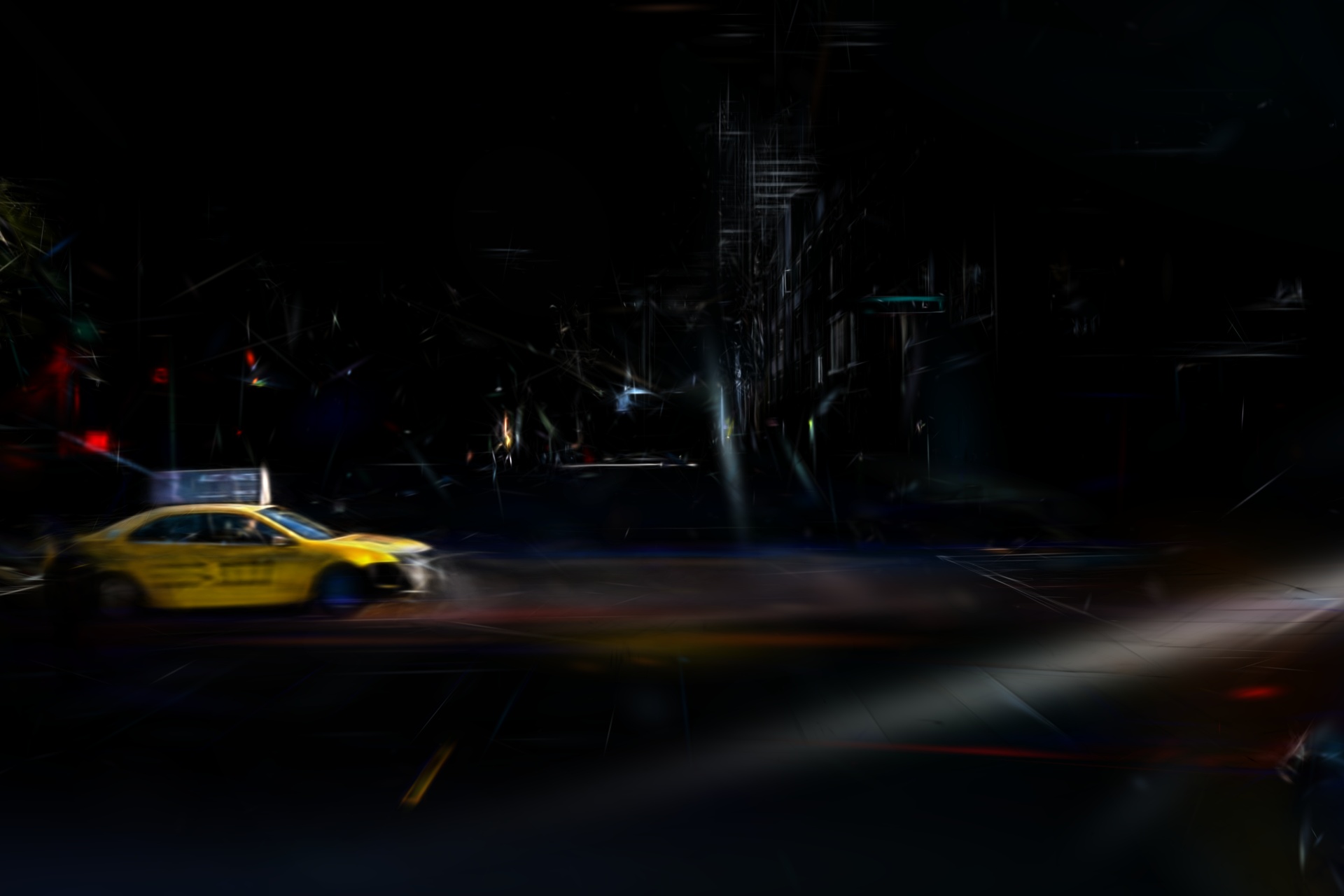}

\vspace{0.5em}

\makebox[0.3\textwidth][c]{\textbf{Full}}
\hspace{0.01\textwidth}
\makebox[0.3\textwidth][c]{\textbf{Static}}
\hspace{0.01\textwidth}
\makebox[0.3\textwidth][c]{\textbf{Dynamic}}

\caption{\rebuttal{Scene separation into static and dynamic elements by PVG.
}}
\label{fig:ablation_visual_grouped_simple}
\end{figure*}

\subsection{Dynamic element analysis}

We visualize the renderings of velocity map $\mathcal{\Bar{V}}$ and staticness $\rho$ map to analyze the behavior of PVG. For the visualization of $\mathcal{\Bar{V}}$, we first transform $\bm{\Bar{v}}$ of each pixel from the world coordinate system to the camera coordinate system and project the $\bm{v}_{cam}$ onto a plane parallel to the camera plane. Then we use the color coding of optical flow for visualization. 
For the visualization of staticness $\rho$ map, we need to clamp each point's $\rho$ to the range of $[0,2]$ and render, otherwise the visualization is not visually distinctive. \rebuttal{Note, pixels with small $\rho$ point to dynamic areas 
and large $\rho$ for static ones.}

\rebuttal{As shown in Fig.~\ref{fig:supp_mvs},  PVG captures not only dynamic objects like cars and people, but also moving light and shadow.
We see from Figure~\ref{fig:ablation_visual_grouped_simple} that, 
static regions remain visually stable and robust to temporal variations, while dynamic regions such as vehicles and trees are accurately captured. This demonstrates PVG's inherent ability to effectively separate and render static and dynamic scene elements.}

\subsection{Ablation study}
\label{sec:experiment ablation}

\begin{figure}[t]
    \centering
    \begin{minipage}[t]{0.37\textwidth}
    \begin{subfigure}[b]{0.49\textwidth}
        \includegraphics[width=\linewidth]{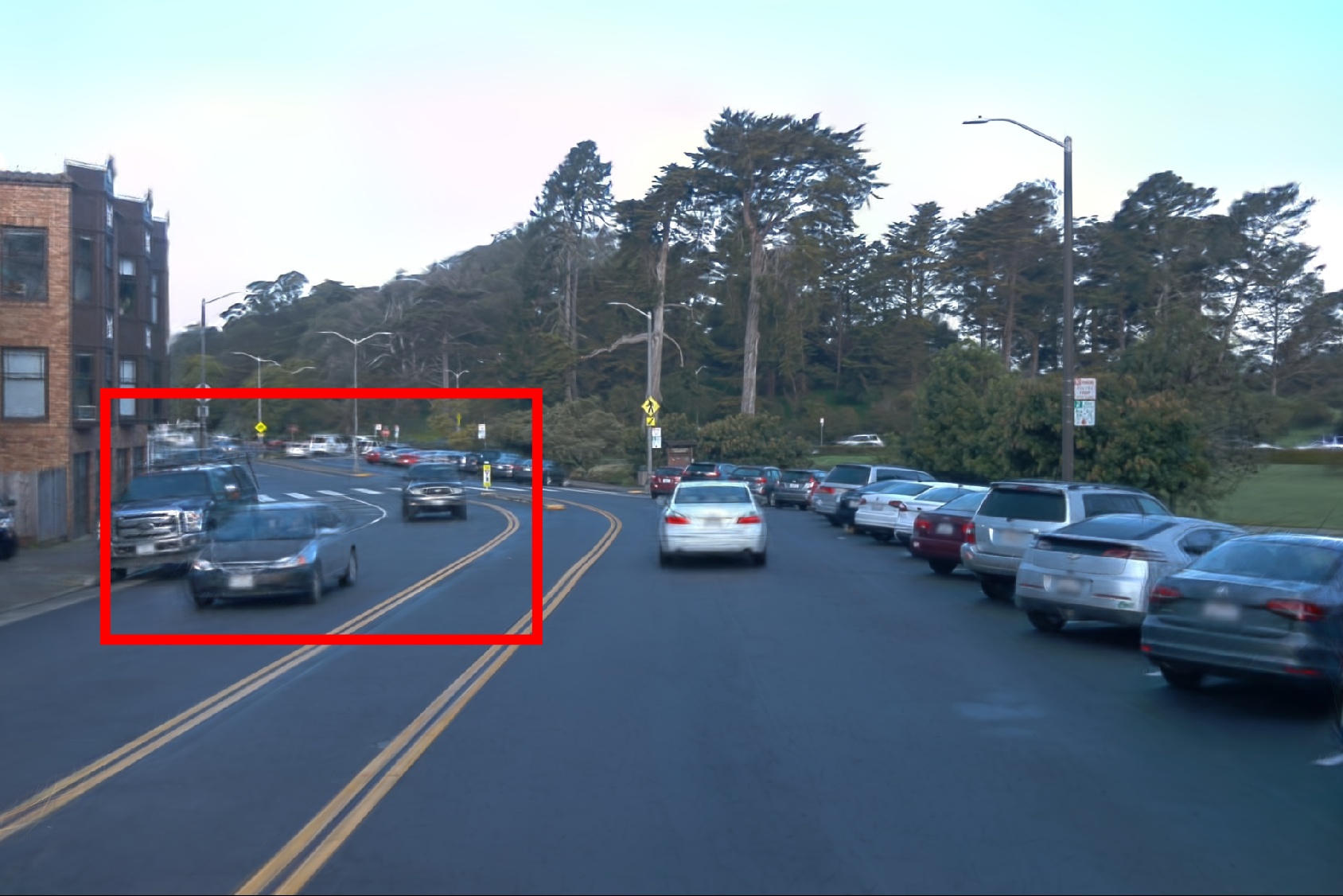}
        \caption{On}
    \end{subfigure}
    \begin{subfigure}[b]{0.49\textwidth}
        \includegraphics[width=\linewidth]{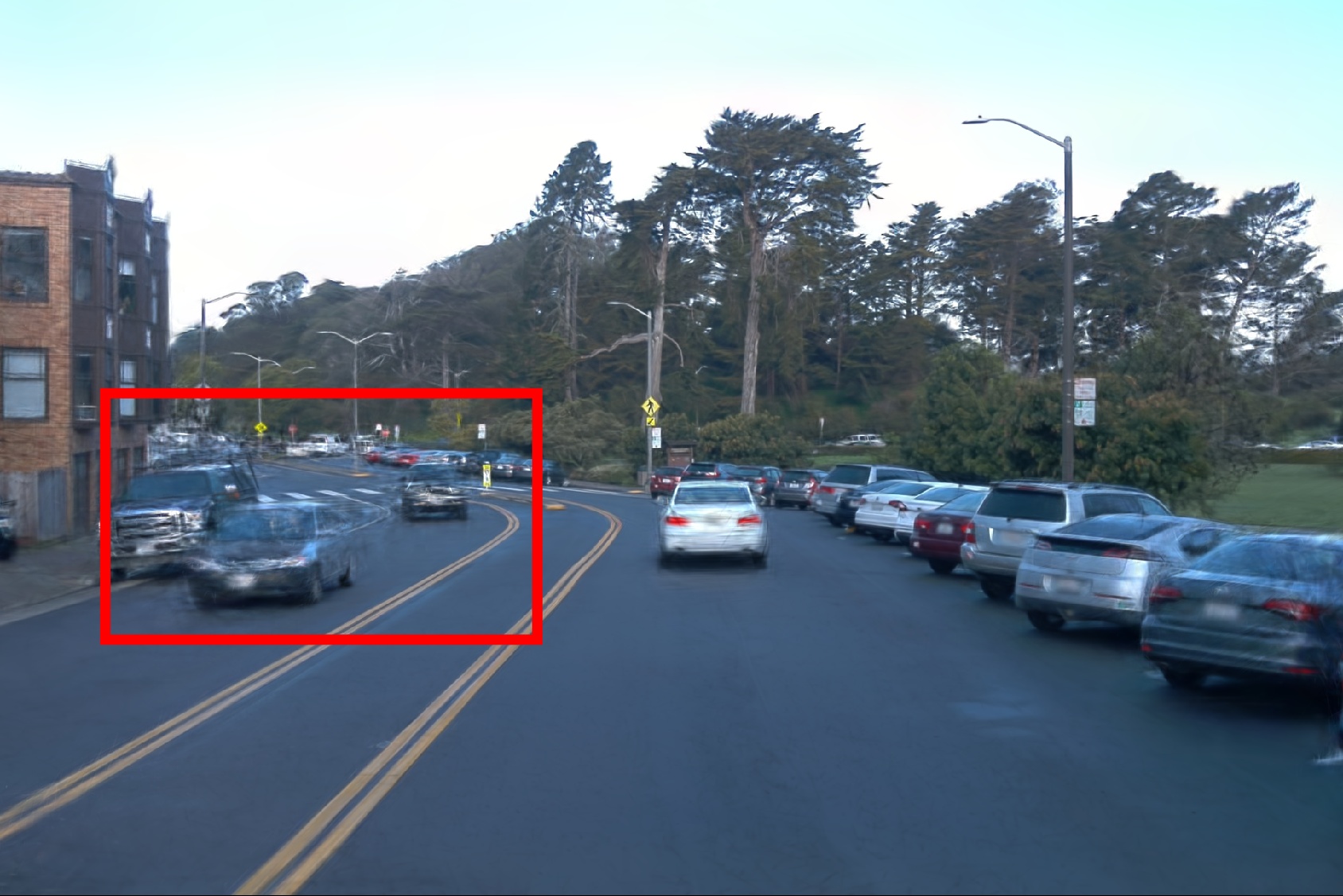}
        \caption{Off}
    \end{subfigure}

        \vspace{-2mm}
        \captionsetup{font=scriptsize}
        \caption{Temporal smoothing (a) on and (b) off.}

        \label{fig:ablation flow-based}
    \end{minipage}
    \begin{minipage}[t]{0.61\textwidth}
    \begin{subfigure}[b]{0.32\textwidth}
        \includegraphics[width=\linewidth]{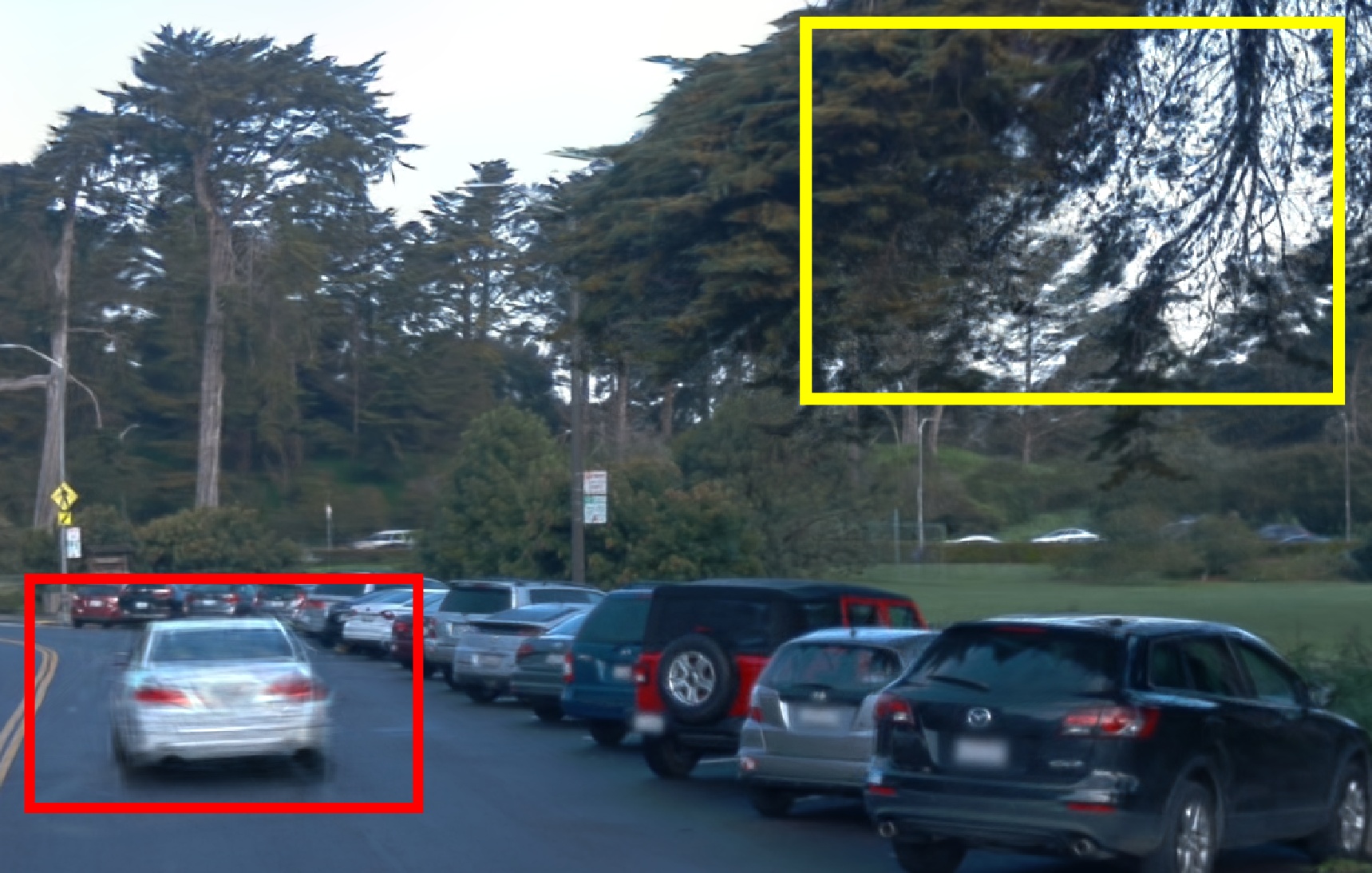}
                \caption{Constant}
    \end{subfigure}
    \begin{subfigure}[b]{0.32\textwidth}
        \includegraphics[width=\linewidth]
        {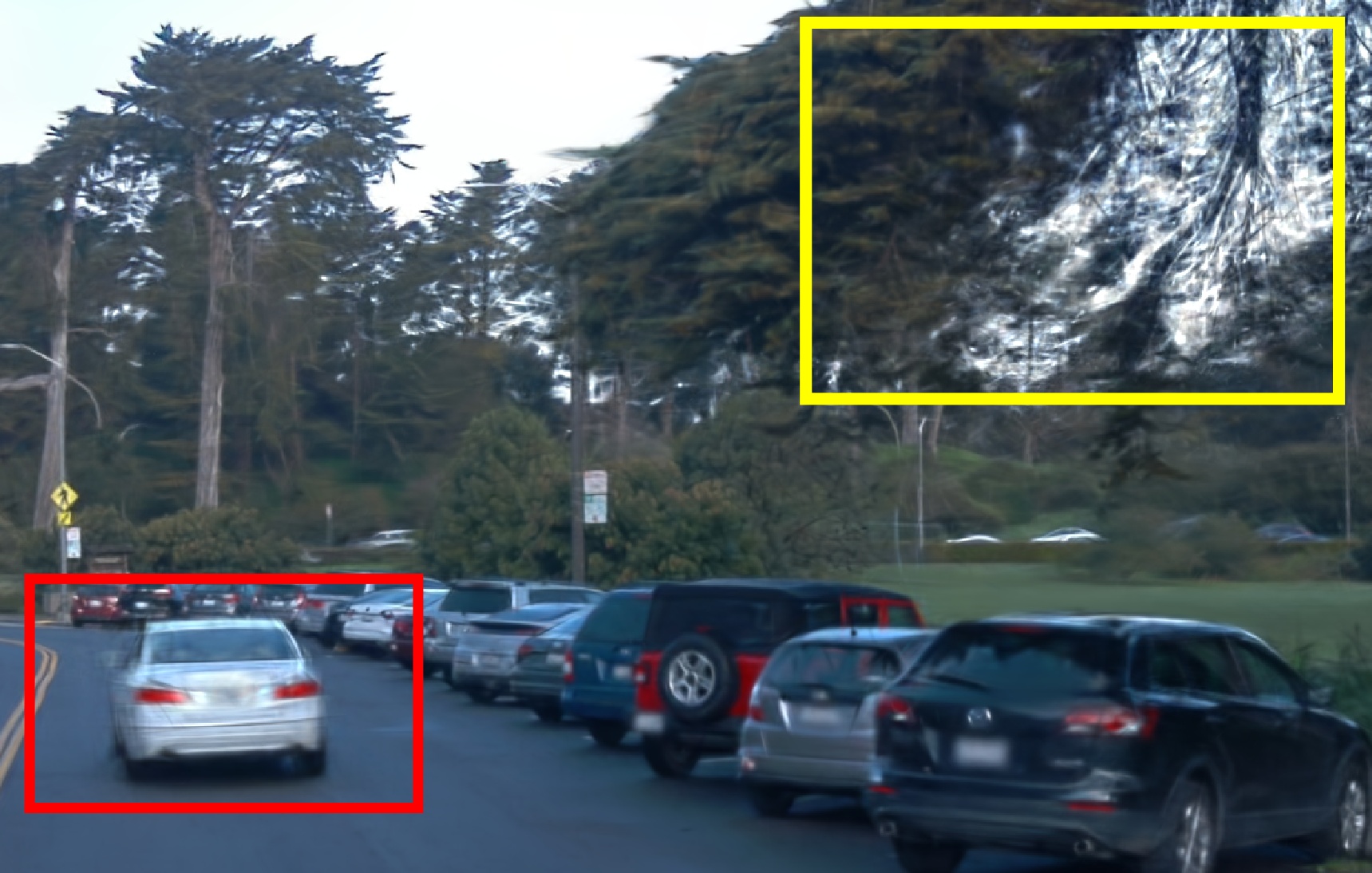}
                \caption{Linear}
    \end{subfigure}
    \begin{subfigure}[b]{0.32\textwidth}
        \includegraphics[width=\linewidth]
        {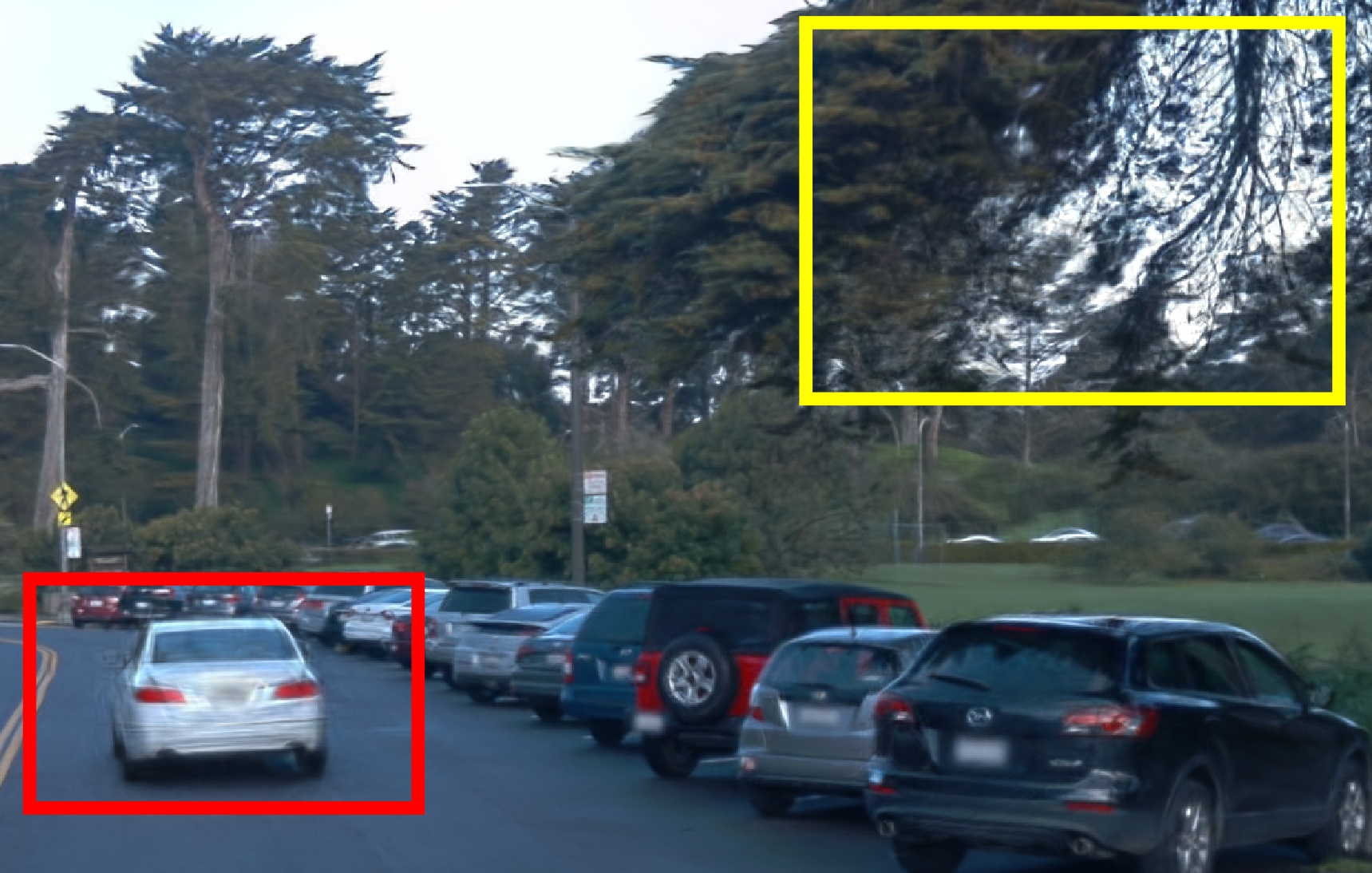}
                \caption{Ours}
    \end{subfigure}
        \captionsetup{font=scriptsize}
                \vspace{-2mm}
        \caption{Dynamics models: (a) Constant, (b) Linear, (c) Ours,
        with PSNR=27.77/27.09/28.11, respectively.}
        \label{fig:ablation cycle length}
    \end{minipage}
\end{figure}

\begin{figure}[t]
\centering
\begin{subfigure}{0.96\textwidth}
\includegraphics[width=\linewidth]{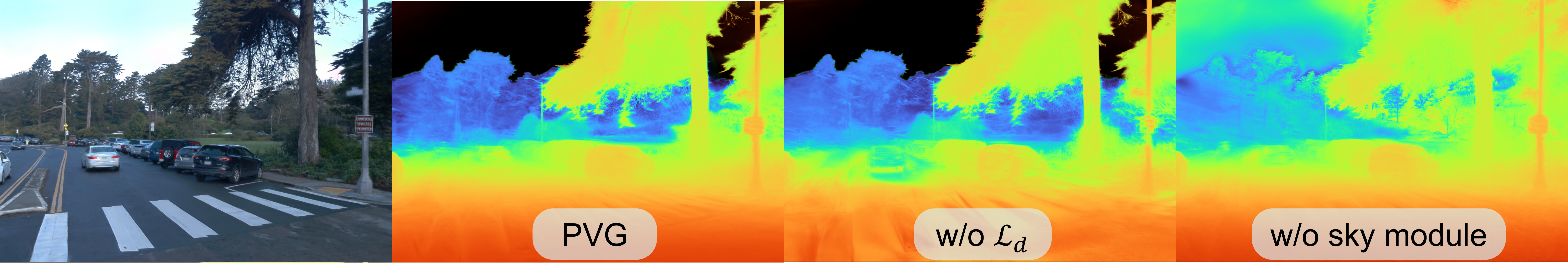}
  \caption{Effect of the depth $\mathcal{L}_d$ and sky module.
  }
  \label{fig: ab on depth map}
\end{subfigure}
\begin{subfigure}{0.96\textwidth}
      \includegraphics[width=\linewidth]{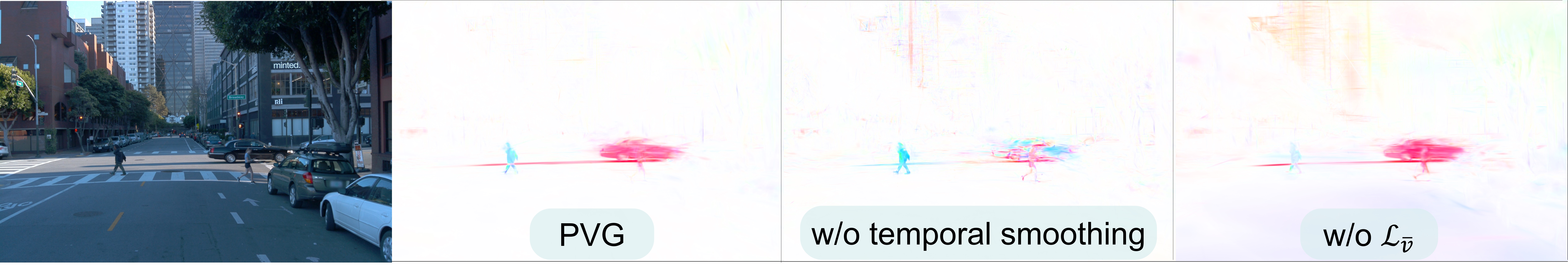}
  \caption{Effect of temporal smoothing for motion trend (car, pedestrian and shadow) and sparse velocity $\mathcal{L}_{\Bar{v}}$ for enabling static component modeling.
  We render the average velocity map $\mathcal{\Bar{V}}$, and color its uv component by optical flow color coding. 
  }
  \label{fig: ab on flow map}
\end{subfigure}
\begin{subfigure}{0.96\textwidth}
      \includegraphics[width=\linewidth]{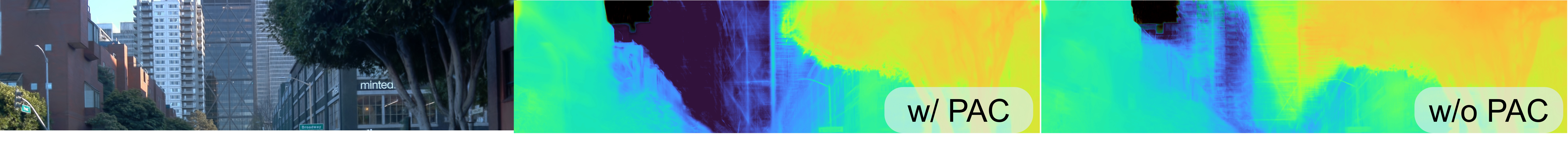}
  \caption{Effect of Positional-Aware Control (PAC)
  in helping the depth of distant components.
  }
  \label{fig: ab on pac}
\end{subfigure}
\caption{More visualization of the ablation study.}
\label{fig: more ab}
\vspace{-5mm}
\end{figure}

We conduct an ablation study to investigate the impact of the primary components of PVG on novel view synthesis of dynamic scenes on the Waymo Open Dataset. We set $\eta=1$ to deactivate the scene flow-based temporal smoothing mechanism, a crucial element in RGB rendering. 
The results are given in Table~\ref{table:ablation study}.
The temporal smoothing mechanism results in enhanced smoothness in novel view rendering and promotes temporal and spatial consistency of PVG points (Fig.~\ref{fig:ablation flow-based}). Meanwhile, temporal smoothing can help the model to learn the correct motion trend (Fig.~\ref{fig: ab on flow map}). Our findings indicate that the integration of LiDAR supervision, sky refinement module contributes to more plausible geometry (Fig.~\ref{fig: ab on depth map}). Depth loss and sky module make a minimal effect on novel view RGB rendering mainly because test novel views are interpolations of training views (both in the driving paths), making these metrics {\em not fully} reflect the quality of synthesis. Fig.~\ref{fig: ab on pac} demonstrates positional-aware control strategy significantly improves the reconstruction of distant views. Additionally, Fig.~\ref{fig: ab on flow map} shows the inclusion of velocity loss facilitates the convergence of velocity to a sparser rank, thereby simplifying the segmentation process between dynamic and static elements.

\begin{table*}
    \centering
    \begin{minipage}[t]{0.45\textwidth}
        \centering
        \caption{Novel view synthesis of static scene, same settings as SNeRF~\cite{xie2023s} }
        \scalebox{0.75}{
        \begin{tabular}{lccc}
            \toprule
            Methods & PSNR$\uparrow$ & SSIM$\uparrow$ & LPIPS$\downarrow$ \\
            \midrule
            S-NeRF~\cite{xie2023s} & 23.60 & 0.743 & 0.422 \\
            StreetSurf~\cite{guo2023streetsurf} & 23.08  & 0.806 & 0.396 \\
            3DGS~\cite{kerbl20233d} & 24.08 & 0.807 & 0.322\\
            \bf PVG (Ours) & \textbf{26.06} & \textbf{0.832} & \textbf{0.266}\\
            \bottomrule
        \end{tabular}
        }
        \label{tab:static waymo snerf}
    \end{minipage}
    \hfill
    \begin{minipage}[t]{0.5\textwidth}
        \centering
        \caption{Ablation on novel view synthesis.}
        \scalebox{0.70}{
        \begin{tabular}{lccc}
            \toprule
             & $\uparrow$PSNR & $\uparrow$SSIM & $\downarrow$LPIPS  \\ \midrule
            w/o temporal smoothing& 26.50 & 0.805 & 0.315 \\
            w/o depth loss $\mathcal{L}_{d}$ & 28.10 & 0.849 & 0.279 \\
            
            w/o velocity loss $\mathcal{L}_{\bm{\Bar{v}}}$& 28.04 & 0.847 & 0.281  \\
            
            w/o sky refinement & 28.01 & 0.846 & 0.278  \\
            w/o position-aware control & 27.90 & 0.846 & 0.282  \\
            
            \cmidrule(lr){1-4}
            PVG (Ours)& \textbf{28.11} & \textbf{0.849} & \textbf{0.279}  \\ \bottomrule
            \end{tabular}
            \label{table:ablation study}
            }
    \end{minipage}
\end{table*}

\noindent{\bf Dynamics formulation}
We investigate the influence of $\widetilde{\bm\mu}(t)$ formulation. 
We compare our design (Eq. \eqref{eq: mu}) with two alternatives: (a)
A constant model, %
i.e., $\widetilde{\bm{\mu}}(t)=\bm{\mu}$,
and 
(b) A linear model, i.e., %
$\widetilde{\bm{\mu}}(t)=\bm{\mu} + \bm{v}(t-\tau)$.
We observe from Fig.~\ref{fig:ablation cycle length} that
(1) The constant model adeptly represents static elements (see yellow box) but exhibits suboptimal performance on dynamic elements (red box);
(2) The linear model can more smoothly capture dynamic aspects. However, it consistently misrepresents static components resulting in ambiguities and ghosting effects. This is because linear model has a hard time optimizing static points to the right position and zero speed.
Our method overcomes both limitations 
by using a sine function based vibration design.

\begin{table*}[ht]
\centering
\begin{minipage}{0.48\textwidth}
\centering
\caption{\rebuttal{Ablation on cycle length \(\l\). 
The \textit{constant} setting corresponds to $l=0$, i.e., $\widetilde{\bm{\mu}}(t)=\bm{\mu}$. The \textit{linear} setting corresponds to $l=\infty$, i.e., $\widetilde{\bm{\mu}}(t)=\bm{\mu}+\bm{v}(t-\tau)$.}}
\label{tab:ablation_l}
\resizebox{0.8\textwidth}{!}{
\begin{tabular}{c|ccc}
\toprule
\(l\) & PSNR↑ & SSIM↑ & LPIPS↓ \\
\hline
constant & 27.77 & 0.848 & 0.281 \\
0.1 & 27.83 & 0.845 & 0.278 \\
0.2 & \textbf{28.11 } & \textbf{0.849 }& \textbf{0.279} \\
0.5 & 28.10 & 0.846 & 0.283 \\
1.0 & 28.01 & 0.843 & 0.289 \\
linear & 27.09 & 0.841 & 0.296 \\

\bottomrule
\end{tabular}
}
\end{minipage}%
\hfill
\begin{minipage}{0.48\textwidth}
\centering
\caption{\rebuttal{Ablation on the scale factor threshold \(r\). %
\textit{Off} means we do not use the positional-aware control strategy.}}
\label{tab:ablation_r}
\begin{tabular}{c|ccc}
\toprule
\(r\) & PSNR↑ & SSIM↑ & LPIPS↓ \\
\hline
0.5$r_0$ & 28.05 &0.847 & 0.281 \\
1.0$r_0$ &  \textbf{28.11} &\textbf{0.849 } &\textbf{0.279 } \\
2.0$r_0$ & 28.03 & 0.847 & 0.280 \\
$\infty$/Off & 27.90 & 0.846 & 0.282 \\
\bottomrule
\end{tabular}
\end{minipage}
\end{table*}

\begin{table}[ht]
\centering
\caption{\rebuttal{Ablation on temporal smoothing probability \(\eta\).}}
\label{tab:ablation_eta}
\begin{tabular}{c|ccc |ccc}
\toprule
 & \multicolumn{3}{c|}{Image reconstruction}& \multicolumn{3}{c}{Novel view synthesis}\\
  $\eta$& PSNR$\uparrow$ & SSIM$\uparrow$ & LPIPS$\downarrow$  & PSNR$\uparrow$ & SSIM$\uparrow$ & LPIPS$\downarrow$\\

\hline
0.25 & \textbf{33.08} & \textbf{0.914 }& \textbf{0.215} & 27.67 & 0.837 & 0.288 \\
0.50 & 32.46 & 0.910 & 0.229 & 28.11 & 0.849 & 0.279 \\
0.75 & 32.60 & 0.909 & 0.224 & 28.16 & 0.851 & 0.276\\
1.00& 
32.27 & 0.905 & 0.230 & \textbf{28.18} &\textbf{0.852}  & \textbf{0.276}\\
\bottomrule
\end{tabular}
\end{table}

\noindent\rebuttal{{\bf Effect of cycle length $l$} We investigated how different values of $l$ affect the reconstruction quality. The setting $l = 0$ corresponds to the \textit{constant} case, while $l \to \infty$ corresponds to the \textit{linear} case. 
As shown in Table~\ref{tab:ablation_l}, the best overall performance is achieved when $l = 0.2$.
We observed that smaller values of $l$ lead to finer reconstruction of static elements, but overly small values can negatively affect the reconstruction of dynamic elements. A moderately chosen $l$ strikes a good balance between static and dynamic reconstruction quality, leading to the best overall results—as illustrated in Fig.~\ref{fig:ablation cycle length}. Meanwhile, the reconstruction quality remains stable across a wide range of $l$ values, as long as extreme value are avoided.
}

\noindent\rebuttal{ {\bf Effect of scale factor $r$}
We tested the effect of varying the threshold $r$ on reconstruction quality. Specifically, we define the base radius $r_0$ as the ego vehicle’s travel range and scale it by different factors to assess its influence. As shown in Table~\ref{tab:ablation_r}, as long as this strategy is applied, the reconstruction performance remains largely consistent across different values of $r$. This is because the threshold mainly affects distant background regions, which are generally less sensitive to the choice of $r$. As a result, PVG demonstrates strong robustness with respect to this parameter.}

\noindent\rebuttal{{\bf Effect of temporal smoothing probability $\eta$} We evaluated the influence of $\eta$ on both reconstruction and novel view synthesis performance. As reported in Table~\ref{tab:ablation_eta}, setting $\eta$ to a smaller value notably degrades the quality of novel view synthesis. This is primarily because weaker temporal regularization leads to insufficient self-supervision, impairing the model's generalization capability. Conversely, increasing $\eta$ introduces stronger temporal constraints, which may slightly compromise reconstruction accuracy due to potential over-regularization. However, beyond a certain threshold, the performance on both tasks plateaus, indicating that the model is relatively robust to the choice of $\eta$ within a broad range. These results highlight the importance of temporal smoothing while demonstrating the stability of PVG.}

\subsection{View synthesis of different camera settings}
\begin{figure}[t]
\vspace{-3mm}
\centering
\begin{subfigure}{0.19\textwidth}
  \includegraphics[width=\linewidth]{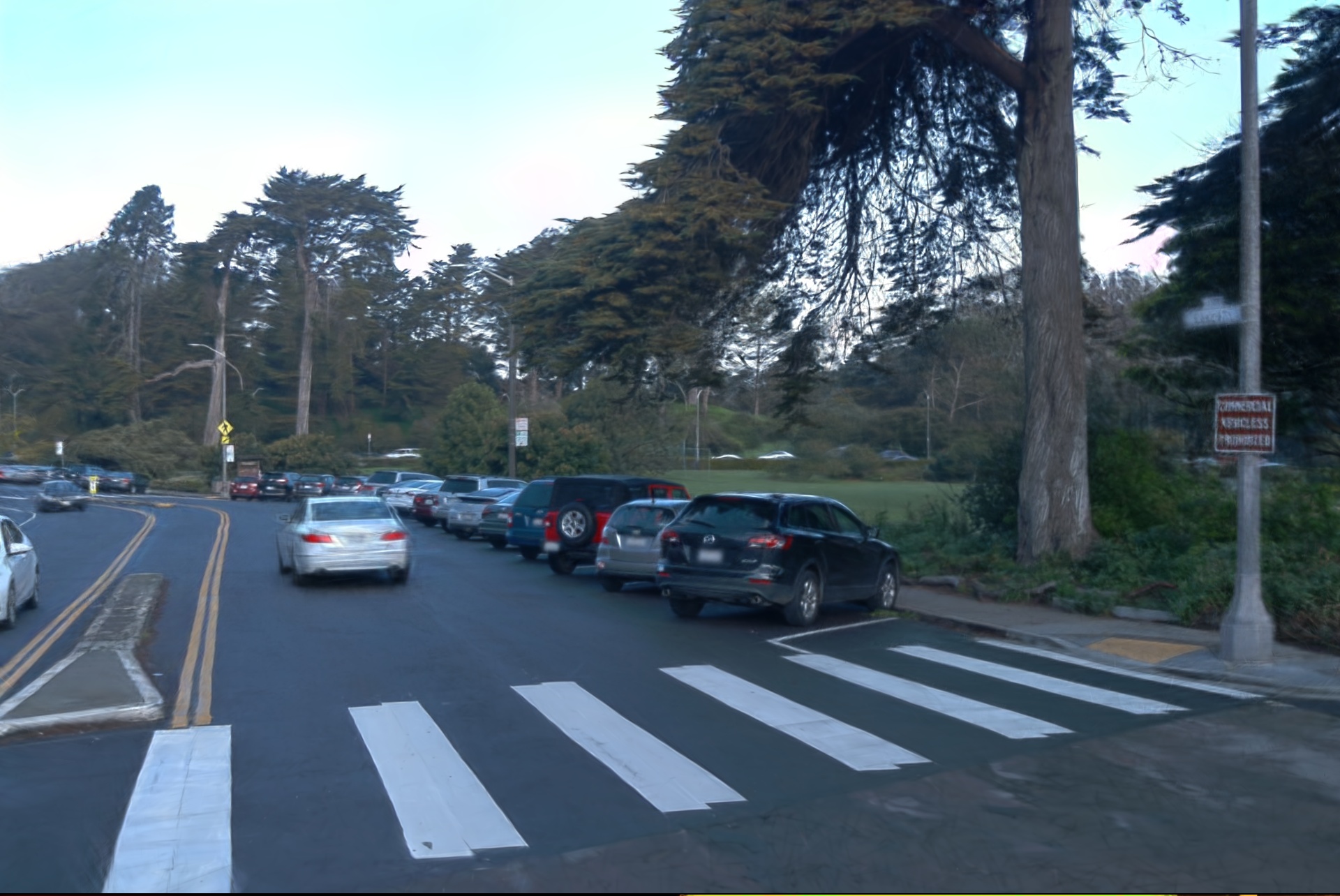}
  \caption{Origin}
  \label{fig:origin}
\end{subfigure}
\begin{subfigure}{0.19\textwidth}
  \includegraphics[width=\linewidth]{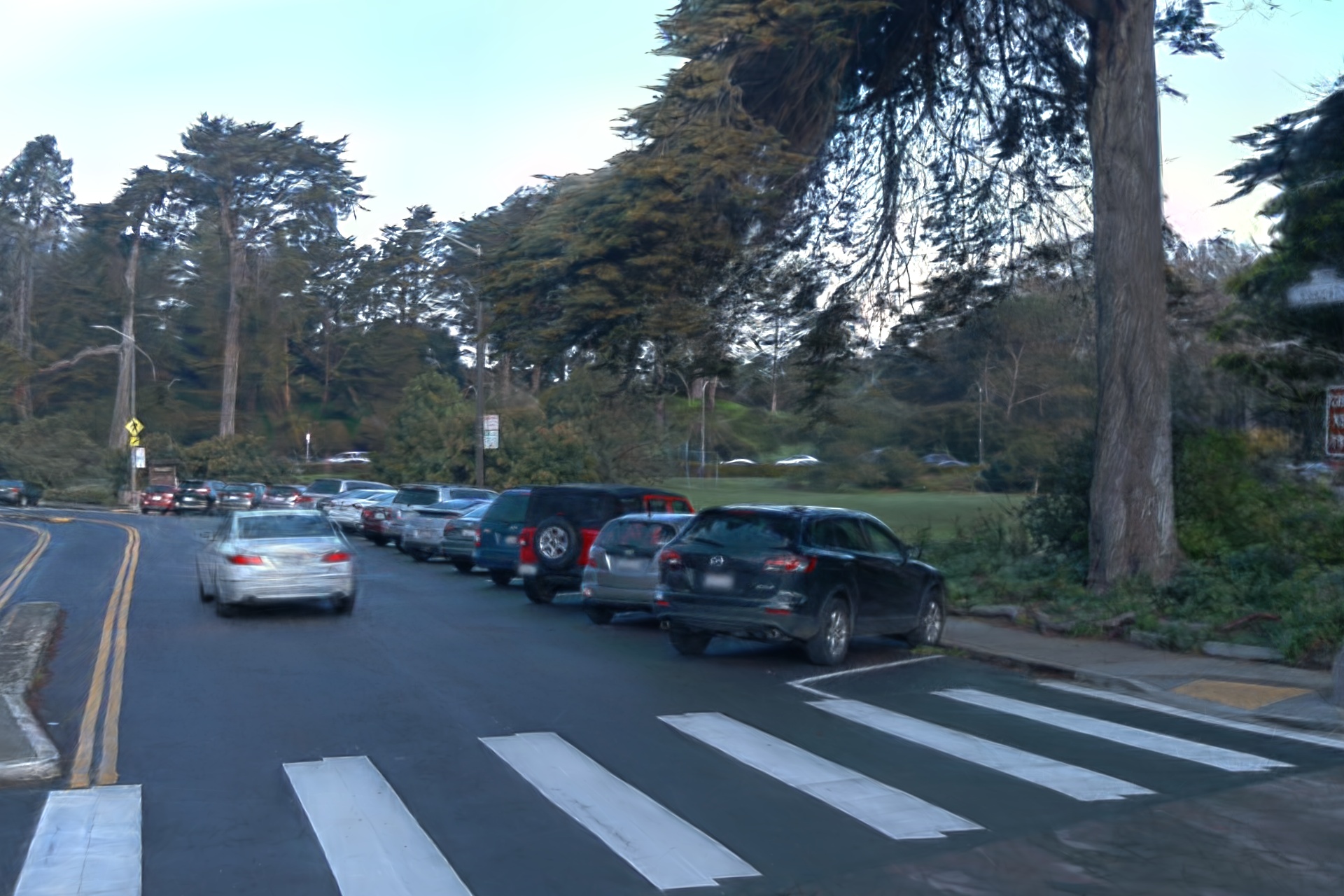}
  \caption{Zoom in}
  \label{fig:zoom in}
\end{subfigure}
\begin{subfigure}{0.19\textwidth}
  \includegraphics[width=\linewidth]{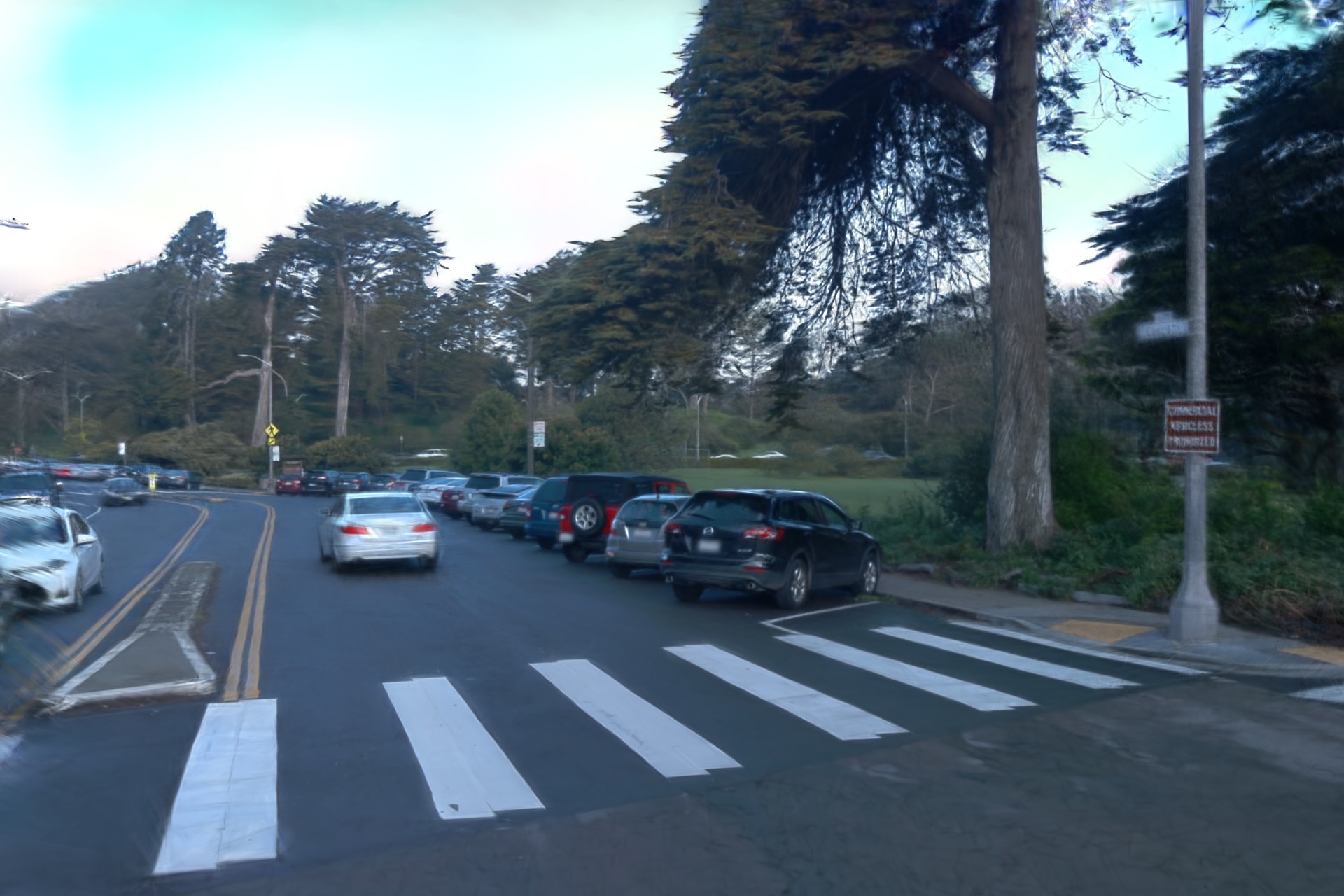}
  \caption{Zoom out}
  \label{fig:zoom out}
\end{subfigure}
\begin{subfigure}{0.19\textwidth}
  \includegraphics[width=\linewidth]{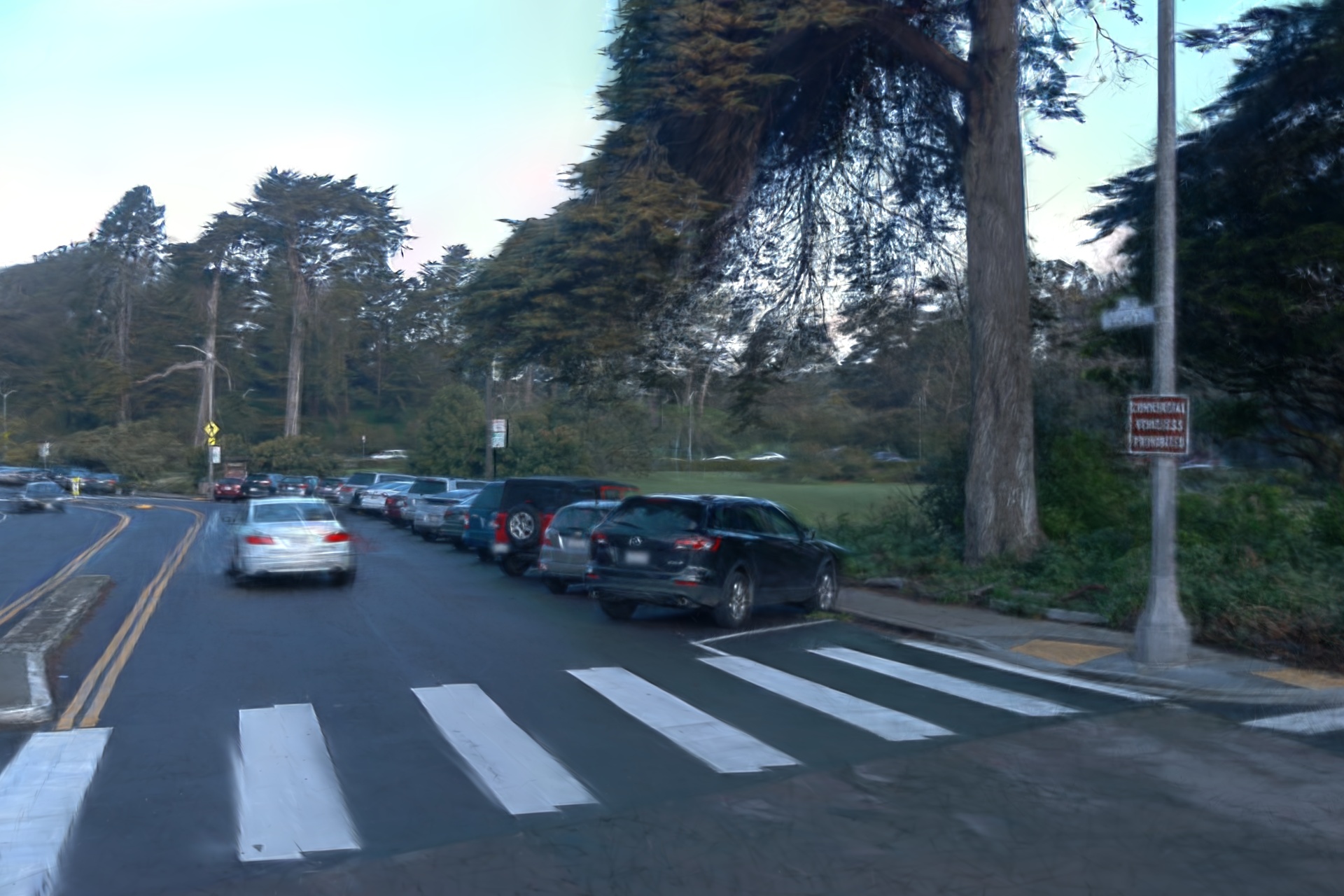}
  \caption{Shift right}
  \label{fig:shift right}
\end{subfigure}
\begin{subfigure}{0.19\textwidth}
  \includegraphics[width=\linewidth]{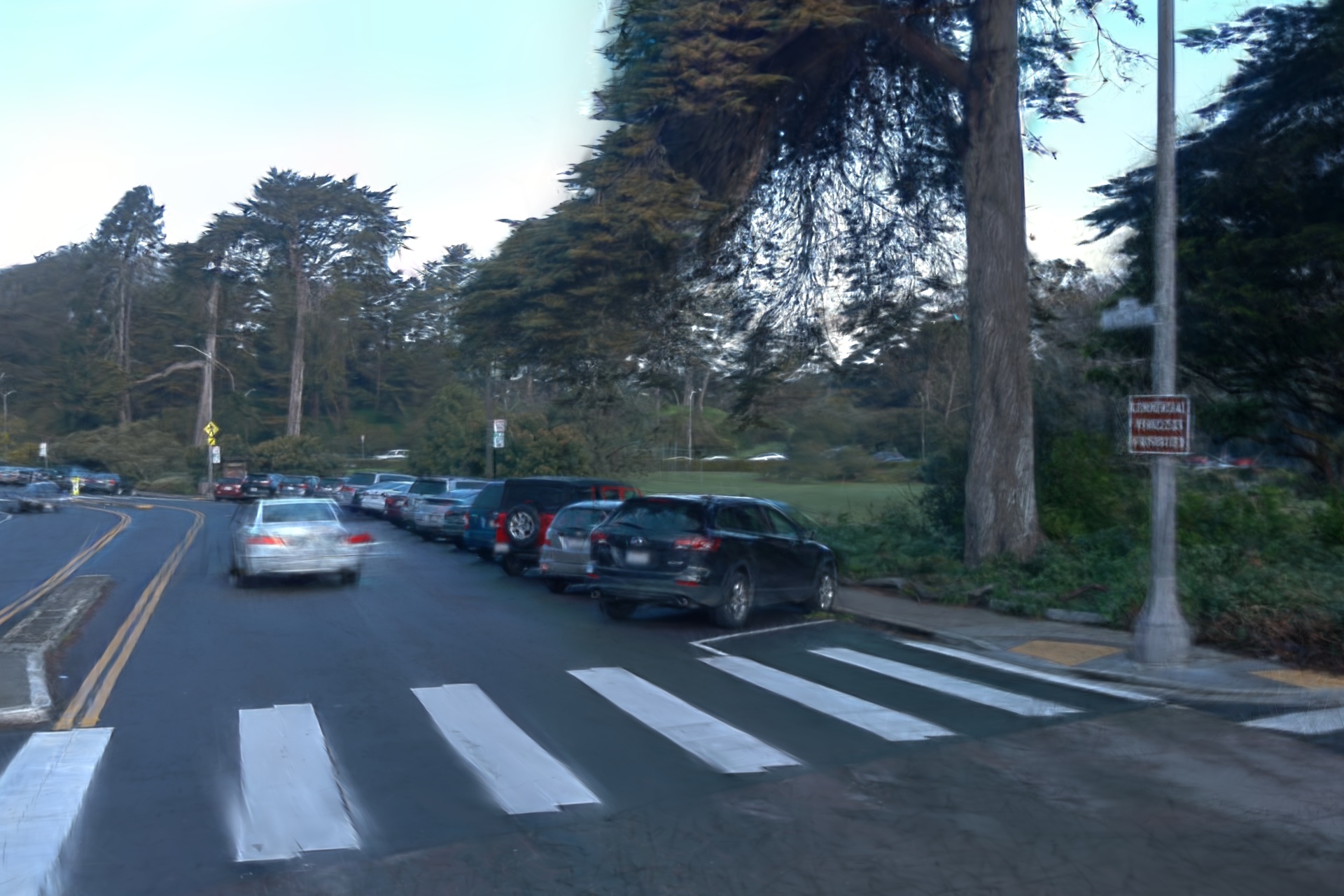}
  \caption{w/o $\mathcal{L}_d$}
  \label{fig:wo ld}
\end{subfigure}

\caption{Rendered images under different camera settings.}
\label{fig:focal}
\vspace{-5mm}
\end{figure}

We test the rendering quality of PVG under 
different camera settings, for evaluating the robustness of novel view synthesis. As in Fig.~\ref{fig:focal}, we zoom in/out, disturb the camera extrinsics and intrinsics for novel view synthesis, and PVG remains stable and compact rendering quality. $\mathcal{L}_d$ improves geometry and (e.g., road and dynamics) and increases generalization performance beyond the driving path (Fig.~\ref{fig:shift right} vs.~\ref{fig:wo ld}). 

\subsection{Evaluation on Plenoptic video dataset}

\begin{table*}[ht]
\caption{\rebuttal{Per-scene PSNR results on the Plenoptic video dataset.}}
\centering
\scalebox{0.7}{
\begin{tabular}{l|c|c|c|c|c|c}
\hline

\hline

\hline

\hline
    & Coffee Martini &  Cook Spinach &  Cut Beef &  Flame Salmon &  Flame Steak &  Sear Steak \\ 

\hline
4DGS(K-Planes)~\cite{wu20244d} & 27.34 & 32.46 & 32.90 & 29.20 & 32.51 & 32.49  \\
4DGS~\cite{yang2023real} & 28.33 & 32.93 & \textbf{33.85} & 29.38 & \textbf{34.03} & 33.51  \\
\bf PVG (Ours) & \textbf{28.64} & \textbf{33.01}& 33.20 & \textbf{29.65} & 33.45 & \textbf{33.60} \\
\hline
\end{tabular}
}
\label{tab:panoptic}
\end{table*}

\rebuttal{To further assess the effectiveness of our model, we conduct experiment on the Plenoptic video dataset~\cite{li2022neural} which was widely evaluated in those 4D Gaussian splatting methods~\cite{wu20244d, yang2024real}.
This dataset consists of six real-world scenes of approximately ten seconds each and involves diverse human motions. For each scene, one view is held out for testing, while the remaining views are used for training. 
To ensure consistency, the reported baselines~\cite{wu20244d, yang2024real} are directly taken from their original papers.
As shown in Table \ref{tab:panoptic} above, PVG achieves competitive performance, %
particularly in dynamic scenes with complex motion and background such as water flow.}

\section{Conclusions}

We present the Periodic Vibration Gaussian (PVG), a model adept at capturing the diverse characteristics of various objects and materials within dynamic urban scenes in a unified formulation. By integrating periodic vibration, time-dependent opacity decay, and a scene flow-based temporal smoothing mechanism into the 3D Gaussian Splatting technique, we have established that our model significantly outperforms the state-of-the-art methods on the Waymo Open Dataset and KITTI benchmark, with significant efficiency advantage in dynamics scene reconstruction and novel view synthesis. While PVG excels in managing dynamic scenes, it encounters limitations in precise geometric representation, attributable to its highly adaptable design. Future efforts will focus on improving geometric accuracy and further refining the model's proficiency in accurately depicting the complexities of urban scenes.

\noindent{\textbf{\textit{Limitations}}}
\rebuttal{Our PVG is built upon independent and discrete Gaussian points. While this design brings advantages such as flexibility, compositionality, and strong fitting capacity, it also introduces challenges in enforcing spatial and temporal coherence for scenarios involving fast-moving dynamic objects. Additionally, due to the independent nature of the points, efficiently editing dynamic objects remains difficult. We consider addressing these challenges through more fine-grained motion modeling and structured priors in future work.}

\section{Data availability statement}
The datasets generated and/or analysed during the current study are available in the Waymo Open Dataset~\cite{sun2020scalability} (\url{https://waymo.com/open}), KITTI~\cite{geiger2012we} (\url{https://www.cvlibs.net/datasets/kitti/index.php}) and Plenoptic video dataset~\cite{li2022neural} (\url{https://github.com/facebookresearch/Neural_3D_Video/releases/tag/v1.0}).

\begin{appendices}
\setcounter{page}{1}
\appendix

\section{More model interpretation}

\begin{figure}[h] 
  \centering
  \includegraphics[width=0.65\textwidth]{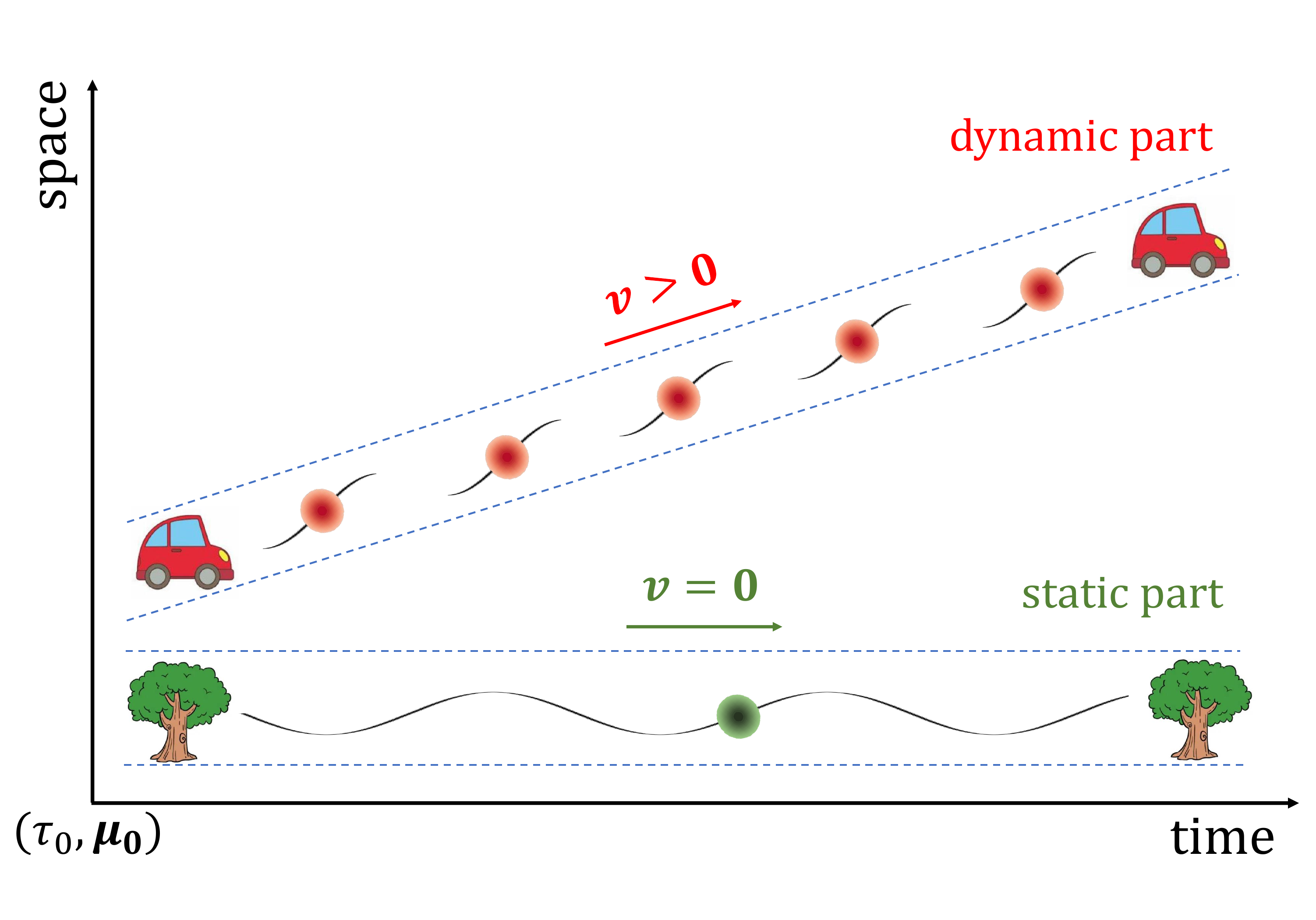}
  \caption{Consider a 4D space-time coordinate system with a non-zero slope for dynamic objects and zero slope for the static. Every PVG point's trajectory is characterized by a piecewise sine function with specific domain of definition and amplitude. 
  PVG points with small staticness coefficient $\rho$ (red points) and short lifespans learn to model dynamic scene parts, alongside those 
with large $\rho$ (green points) and long lifespans for explaining static scene parts. To represent unconstrained motion (e.g., moving car), a collection of PVG points will work out in a cohort.
  }  \label{fig:static and dynamic}  
\end{figure}

\paragraph{Representation}
To facilitate understanding how our proposed PVG, 
we consider a simplified scene with both static and dynamic components,
as illustrated in Fig.~\ref{fig:static and dynamic}.
Concretely, PVG points with long lifespans are used for quantifying static scene elements as the conventional 3D Gaussian counterpart,
whilst those with short lifespans distributed over space and time
for representing the unconstrained dynamic components of a scene.

\paragraph{Temporal smoothing by intrinsic motion}
Due to sparse training frames,  the renderings of novel timestamps are prone to underfitting.
To make the $\{\mathcal{H}_i\}$ temporally and spatially consistent with its intrinsic motion, we exploit the inherent temporal consistency law: 
The status $\mathcal{H}_i(t)$ at time $t$ is similar to the result $\widehat{\mathcal{H}}_i(t)$ of 
translating the status to $t-\Delta t$ by a distance $\bm{v}\Delta t$.
This introduces an additional optimization regularization defined as:
\begin{equation}
    \min \mathbb{E}_{t,\Delta t} \mathbf{Diff}(\{\mathcal{H}_i(t)\}, \{\widehat{\mathcal{H}}_i(t)\}),
\end{equation}
where $\mathbf{Diff}(\cdot)$ is a difference measurement about two set of 3D Gaussian points.
While it is hard for us to directly calculate $\mathbf{Diff}(\cdot)$, we use an 
indirect measurement by rendering function $\mathrm{Render}(\cdot)$. 

Our final objective function could be written as the expectation form of the differences between two ways of rendering:
\begin{equation}
\label{eq: selfsv}
        \min \mathbb{E}_{t,\Delta t,\mathbf{E},\mathbf{I}} \|\mathrm{Render}(\{\mathcal{H}_i(t)\})-\mathrm{Render}(\{\widehat{\mathcal{H}}_i(t)\})\|,
\end{equation}
for any camera extrinsic $\mathbf{E}$ and intrinsic $\mathbf{I}$, timestamp $t$ and small time shift $\Delta t$ (the camera pose and time parameter are omitted for simplicity). 

\begin{figure}[h] 
  \centering
  \includegraphics[width=0.95\textwidth]{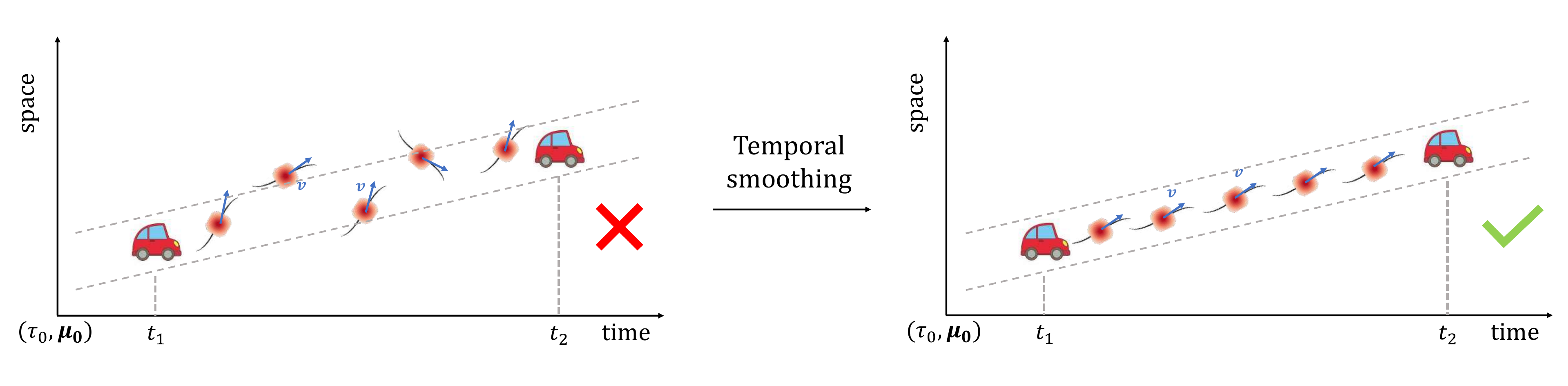}
  \caption{
 Consider two adjacent training frames with timestamp $t_1$ and $t_2$ (a small time window). In a small time period, we assume dynamic objects move linearly.
 At observed times  $t_1$ and $t_2$, RGB renderings could fit well.
 However for the moments in between $(t_1<t_b<t_2)$, we have no corresponding training data to constrain our model $\{\mathcal{H}_i(t_b)\}$. 
 To prevent our model from behaving improperly, we impose a smooth constraint subject to the slope of $\bm{v}$. The frames used to train are knots of a function, what we need to do is making the knots more smoothly connected.
  }  \label{fig:supp ts}
\end{figure}

To compute Eq.~\eqref{eq: selfsv}, we need to render twice, making sampling every camera pose and timestamp computationally expensive. For efficiency, in practice we sample $\{t,\mathbf{E},\mathbf{I}\}$ uniformly from the training set and sample $\Delta t$ from $U(-\delta, +\delta)$, and replace $\mathrm{Render}(\{\mathcal{H}_i(t)\})$ with the ground truth image to only render once for every step. Fig.~\ref{fig:supp ts} shows a more vivid illustration of the temporal smoothing mechanism.

\section{More implementation details}

For experiments on KITTI dataset and whole Waymo sequences compared with EmerNeRF~\cite{yang2023emernerf}, we adjusted some parameters for accommodating more training frames in these data. We train our model for 40000 iterations and densify and prune the gaussians every 200 iterations until 20000 iterations. To better scatter points in 4D space, when we split a PVG point, we shrink each new point' scales by a decay rate of 0.8 and randomly disturb the $\tau$ with $\Delta \tau$ sampled in $\mathcal{N}(0, \beta^2)$ and disturb its position with $\Delta \tau \cdot \Bar{\bm{v}}$. In the first 10000 iterations, we don't shrink $\beta$ and in the next 10000 iterations we shrink $\beta$ to $0.8\beta$ in split operation. In clone operation, we just copy every parameter in the PVG point to a new point.

For experiments with StreetSurf~\cite{guo2023streetsurf}, we deactivate its pose refinement module for a fairer comparison. Otherwise, the refined poses can not align with the GT poses which leads to a low level of PSNR in the novel view synthesis task.

\section{Visualization for Waymo}
\begin{figure}
\centering
\setlength\tabcolsep{1pt}
\begin{tabular}{cccccccc}
\raisebox{0.05\textwidth}[0pt][0pt]{\rotatebox[origin=c]{90}{\tiny{S-NeRF~\cite{xie2023s}}}} 
& \includegraphics[width=0.94\textwidth]{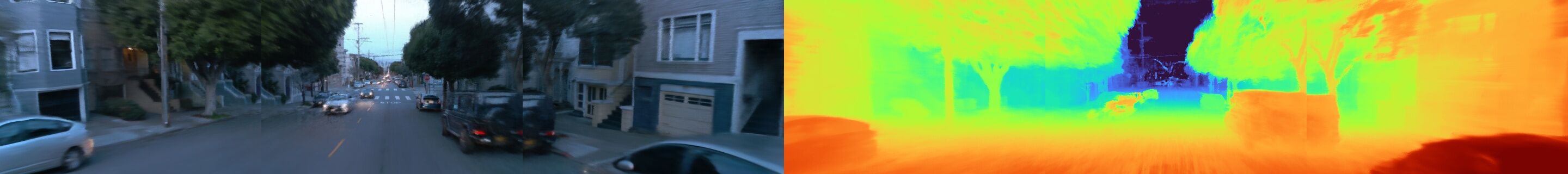} \\
\raisebox{0.05\textwidth}[0pt][0pt]{\rotatebox[origin=c]{90}{\tiny{StreetSurf~\cite{guo2023streetsurf}}}} 
& \includegraphics[width=0.94\textwidth]{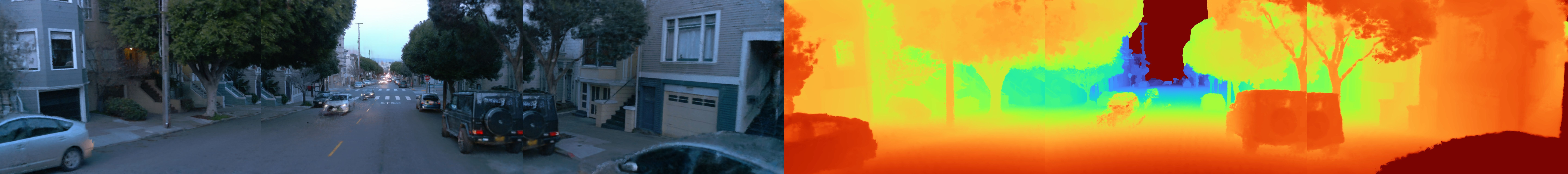} \\
\raisebox{0.05\textwidth}[0pt][0pt]{\rotatebox[origin=c]{90}{\tiny{3DGS~\cite{kerbl20233d}}}} 
& \includegraphics[width=0.94\textwidth]{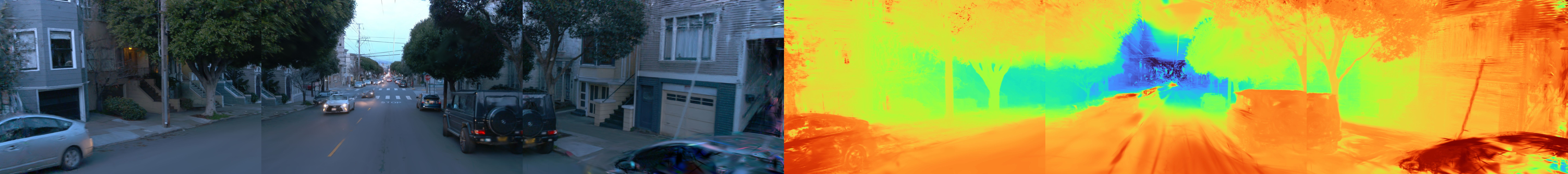} \\
\raisebox{0.05\textwidth}[0pt][0pt]{\rotatebox[origin=c]{90}{\tiny{NSG~\cite{ost2021neural}}}} 
& \includegraphics[width=0.94\textwidth]{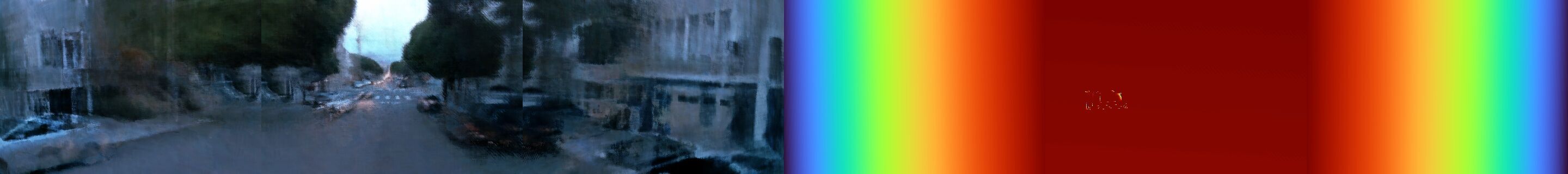} \\
\raisebox{0.05\textwidth}[0pt][0pt]{\rotatebox[origin=c]{90}{\tiny{SUDS~\cite{turki2023suds}}}} 
& \includegraphics[width=0.94\textwidth]{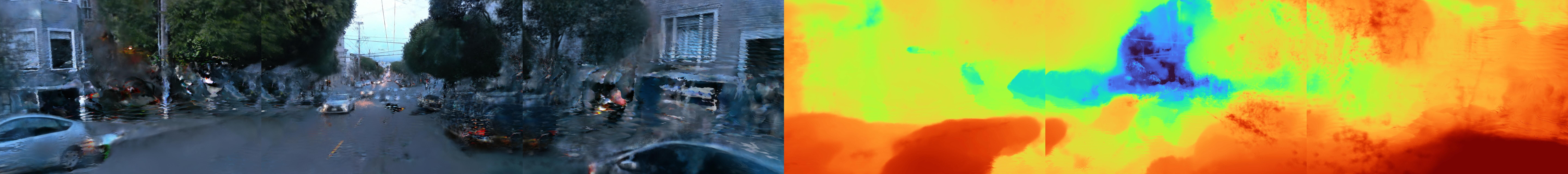} \\
\raisebox{0.05\textwidth}[0pt][0pt]{\rotatebox[origin=c]{90}{\tiny{Mars~\cite{wu2023mars}}}} 
& \includegraphics[width=0.94\textwidth]{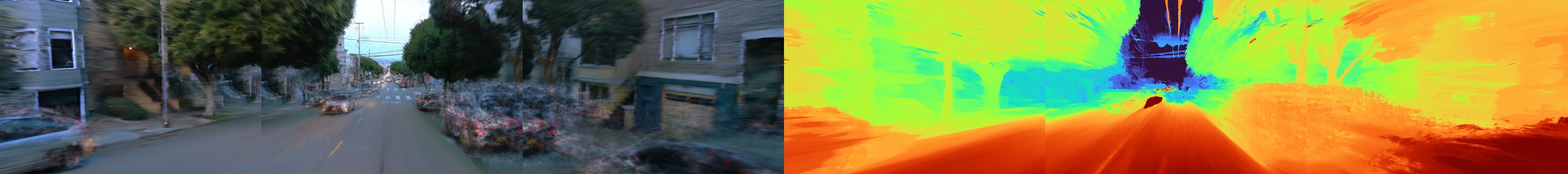} \\
\raisebox{0.06\textwidth}[0pt][0pt]{\rotatebox[origin=c]{90}{\tiny{EmerNeRF~\cite{yang2023emernerf}}}} 
& \includegraphics[width=0.94\textwidth]{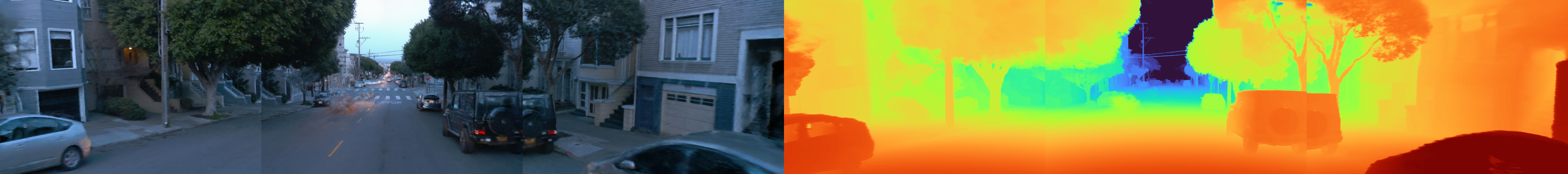} \\
\raisebox{0.05\textwidth}[0pt][0pt]{\rotatebox[origin=c]{90}{\tiny{PVG (Ours)}}} 
& \includegraphics[width=0.94\textwidth]{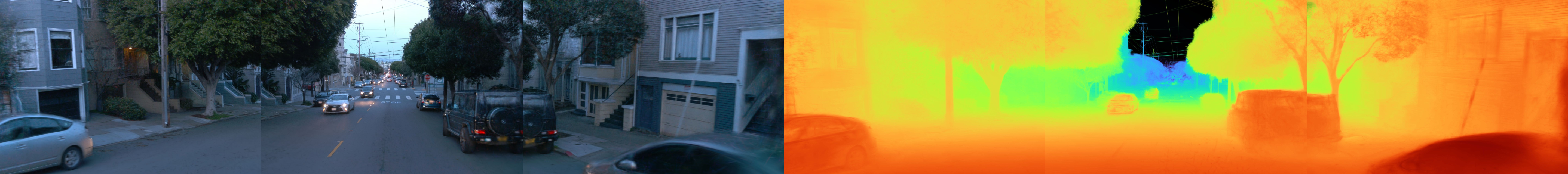} \\
\raisebox{0.05\textwidth}[0pt][0pt]{\rotatebox[origin=c]{90}{\tiny{GT}}} 
& \includegraphics[width=0.94\textwidth]{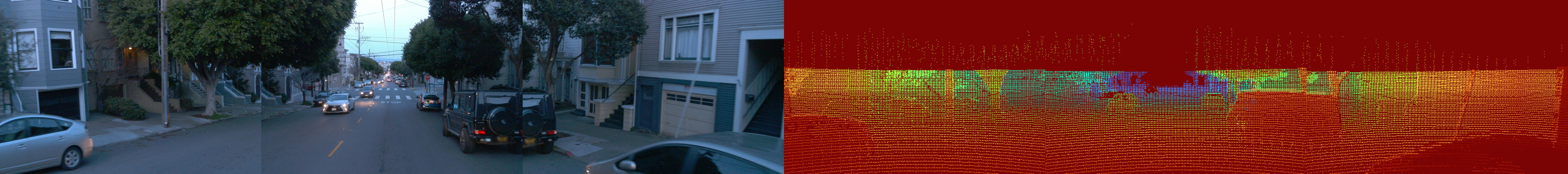} \\
\end{tabular}

\caption{Qualitative results of novel view synthesis on Waymo. GT: Ground-truth.}
\label{fig:supp_waymo_reconstruction}
\end{figure}

\begin{figure}
\centering
\setlength\tabcolsep{1pt}
\begin{tabular}{cccccccc}
\raisebox{0.05\textwidth}[0pt][0pt]{\rotatebox[origin=c]{90}{\tiny{S-NeRF~\cite{xie2023s}}}} 
& \includegraphics[width=0.94\textwidth]{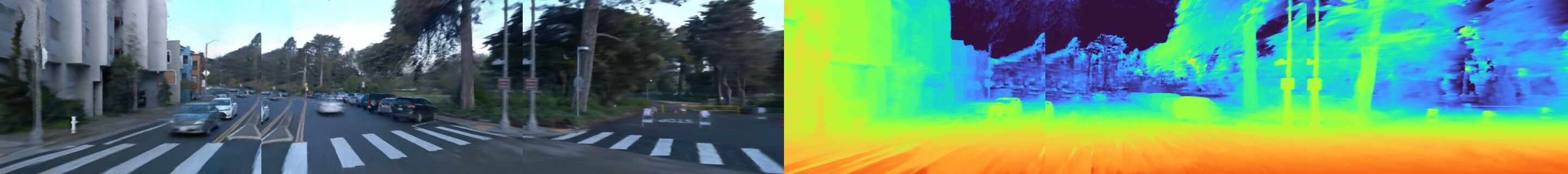} \\
\raisebox{0.05\textwidth}[0pt][0pt]{\rotatebox[origin=c]{90}{\tiny{StreetSurf~\cite{guo2023streetsurf}}}} 
& \includegraphics[width=0.94\textwidth]{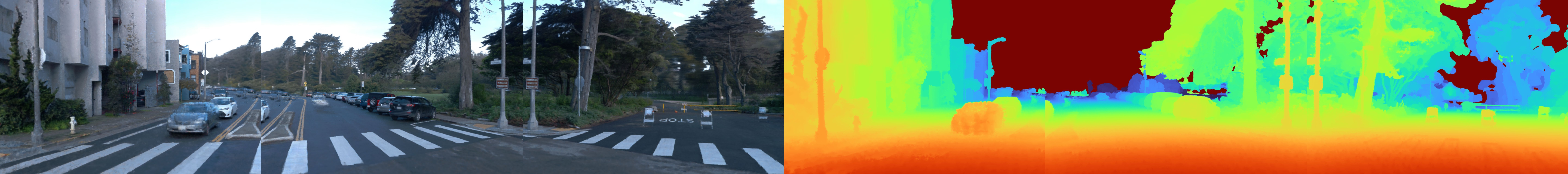} \\
\raisebox{0.05\textwidth}[0pt][0pt]{\rotatebox[origin=c]{90}{\tiny{3DGS~\cite{kerbl20233d}}}} 
& \includegraphics[width=0.94\textwidth]{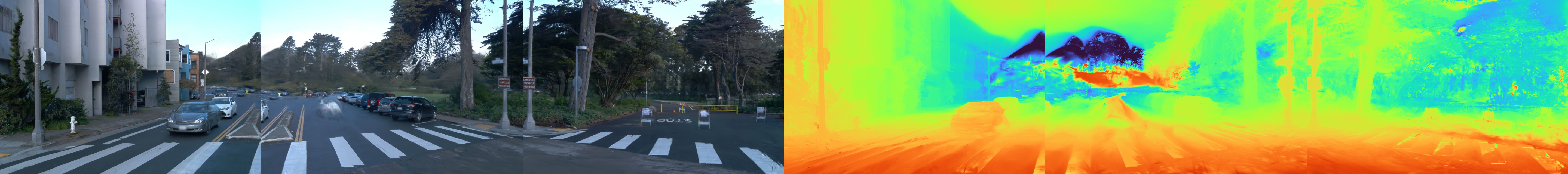} \\
\raisebox{0.05\textwidth}[0pt][0pt]{\rotatebox[origin=c]{90}{\tiny{NSG~\cite{ost2021neural}}}} 
& \includegraphics[width=0.94\textwidth]{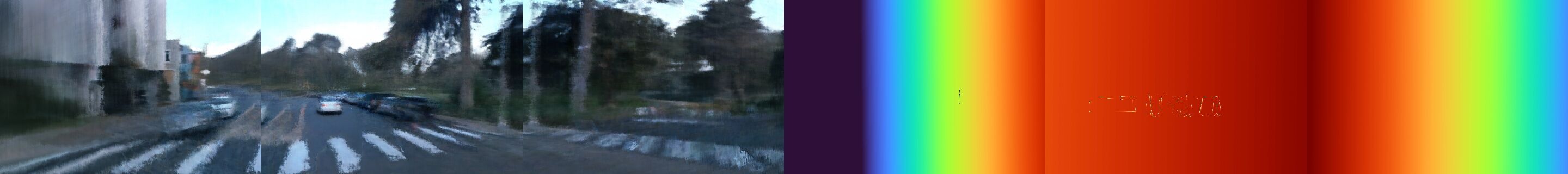} \\
\raisebox{0.05\textwidth}[0pt][0pt]{\rotatebox[origin=c]{90}{\tiny{SUDS~\cite{turki2023suds}}}} 
& \includegraphics[width=0.94\textwidth]{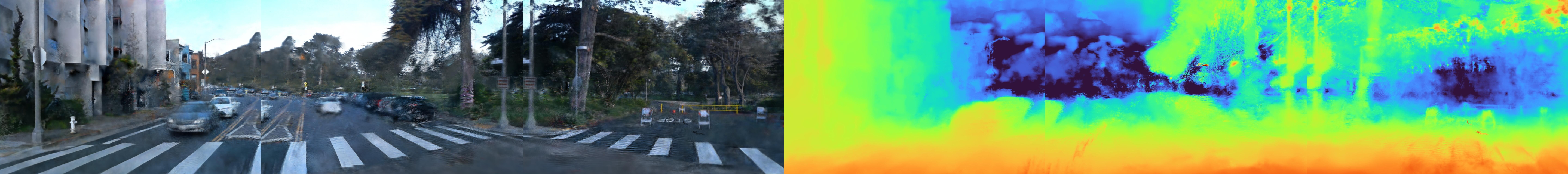} \\
\raisebox{0.05\textwidth}[0pt][0pt]{\rotatebox[origin=c]{90}{\tiny{Mars~\cite{wu2023mars}}}} 
& \includegraphics[width=0.94\textwidth]{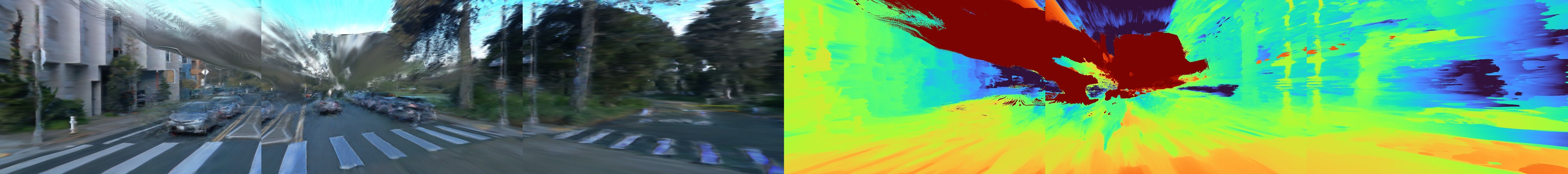} \\
\raisebox{0.06\textwidth}[0pt][0pt]{\rotatebox[origin=c]{90}{\tiny{EmerNeRF~\cite{yang2023emernerf}}}} 
& \includegraphics[width=0.94\textwidth]{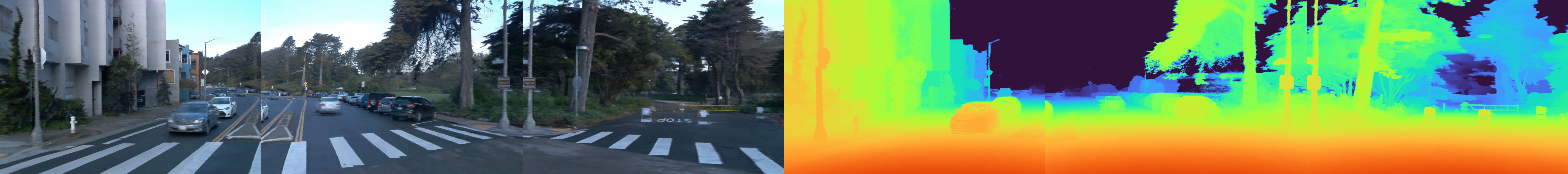} \\
\raisebox{0.05\textwidth}[0pt][0pt]{\rotatebox[origin=c]{90}{\tiny{PVG (Ours)}}} 
& \includegraphics[width=0.94\textwidth]{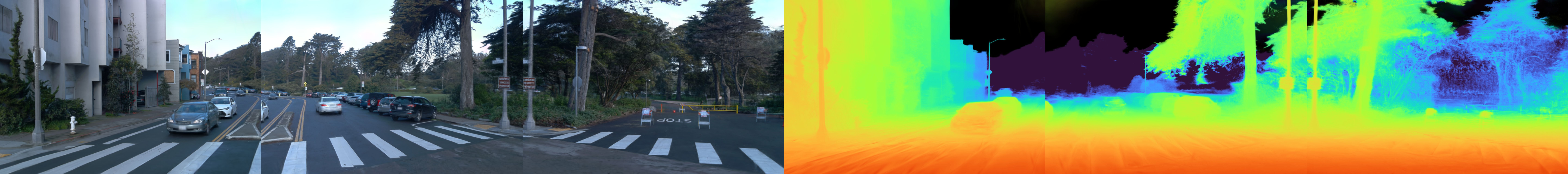} \\
\raisebox{0.05\textwidth}[0pt][0pt]{\rotatebox[origin=c]{90}{\tiny{GT}}} 
& \includegraphics[width=0.94\textwidth]{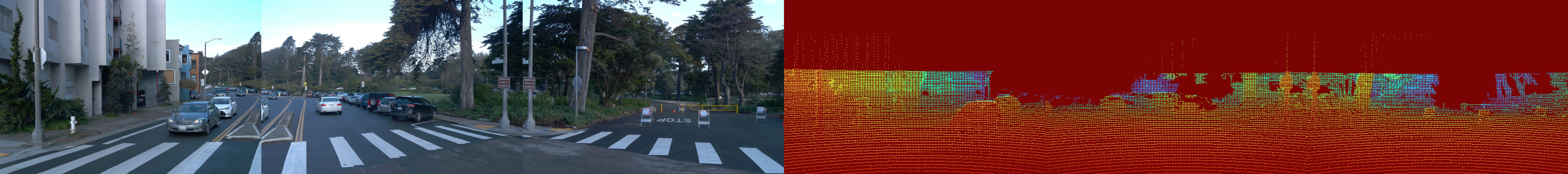} \\
\end{tabular}
\caption{Qualitative results of novel view synthesis on Waymo.}
\label{fig:supp_waymo_novel}
\end{figure}

 More novel view synthesis results are in  Fig.~\ref{fig:supp_waymo_reconstruction} and Fig.~\ref{fig:supp_waymo_novel}. We project the LiDAR points to the camera world as the ground truth depth map. SUDS~\cite{turki2023suds} is good at image reconstruction, but poor at novel view synthesis. 3DGS~\cite{kerbl20233d} is only good for static and close-range reconstruction but unable to handle distance view and dynamic objects. 
 In contrast, our method can not only reconstruct both the near and distance well, but also render images with the quality as the ground-truth.

\section{Visualization for KITTI}
\begin{figure}
\centering
\setlength\tabcolsep{1pt}
\begin{tabular}{cccccccc}
\raisebox{0.07\textwidth}[0pt][0pt]{\rotatebox[origin=c]{90}{\tiny{S-NeRF~\cite{xie2023s}}}} 
& \includegraphics[width=0.94\textwidth]{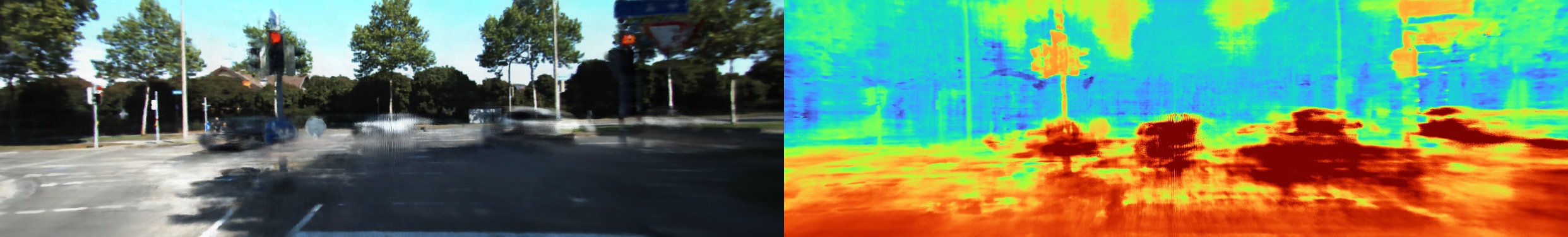} \\
\raisebox{0.07\textwidth}[0pt][0pt]{\rotatebox[origin=c]{90}{\tiny{StreetSurf~\cite{guo2023streetsurf}}}} 
& \includegraphics[width=0.94\textwidth]{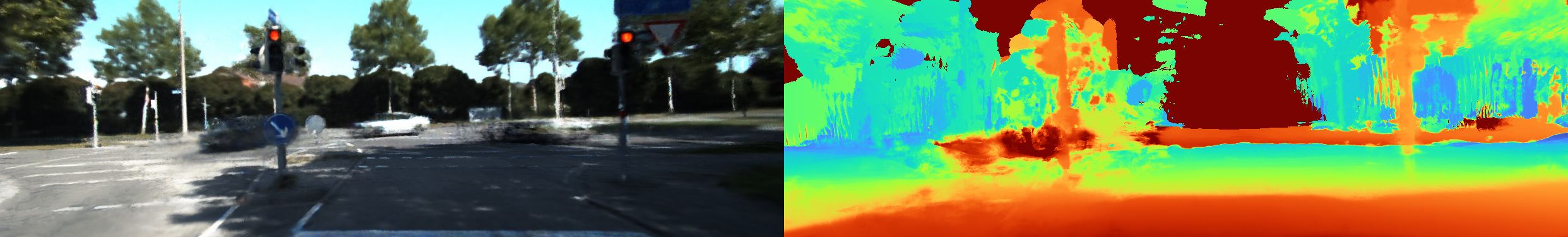} \\
\raisebox{0.07\textwidth}[0pt][0pt]{\rotatebox[origin=c]{90}{\tiny{3DGS~\cite{kerbl20233d}}}} 
& \includegraphics[width=0.94\textwidth]{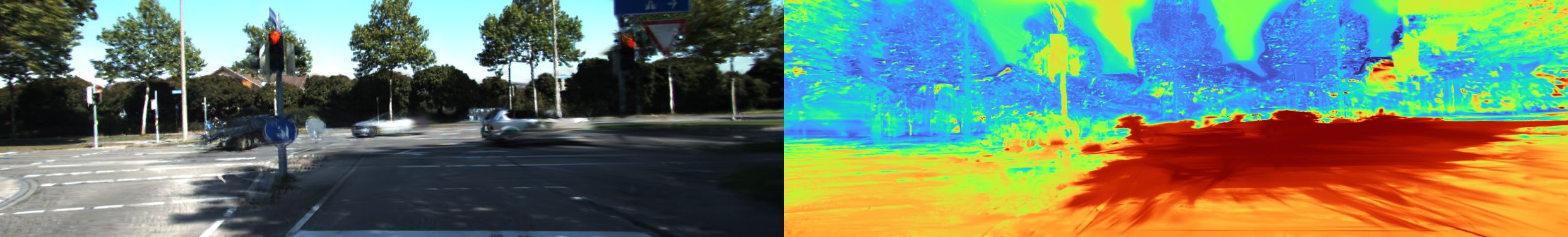} \\
\raisebox{0.07\textwidth}[0pt][0pt]{\rotatebox[origin=c]{90}{\tiny{NSG~\cite{ost2021neural}}}} 
& \includegraphics[width=0.94\textwidth]{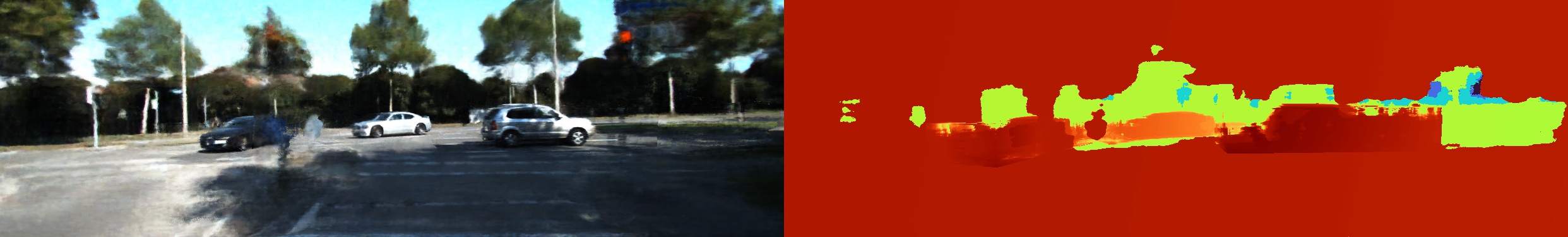} \\
\raisebox{0.07\textwidth}[0pt][0pt]{\rotatebox[origin=c]{90}{\tiny{SUDS~\cite{turki2023suds}}}} 
& \includegraphics[width=0.94\textwidth]{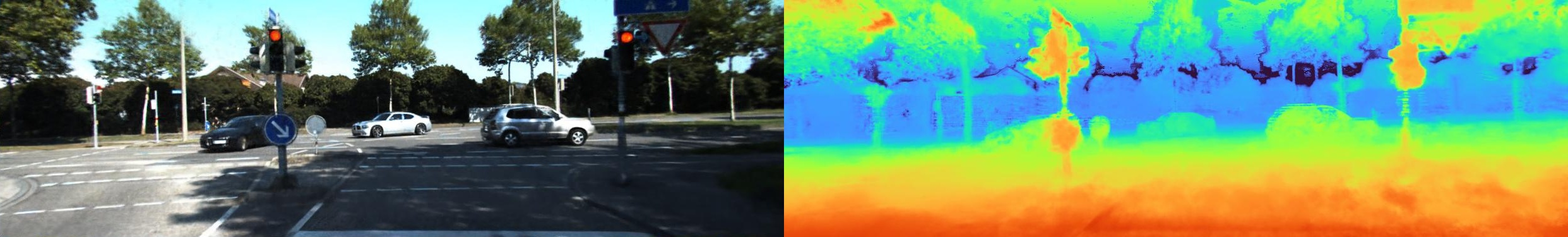} \\
\raisebox{0.07\textwidth}[0pt][0pt]{\rotatebox[origin=c]{90}{\tiny{Mars~\cite{wu2023mars}}}} 
& \includegraphics[width=0.94\textwidth]{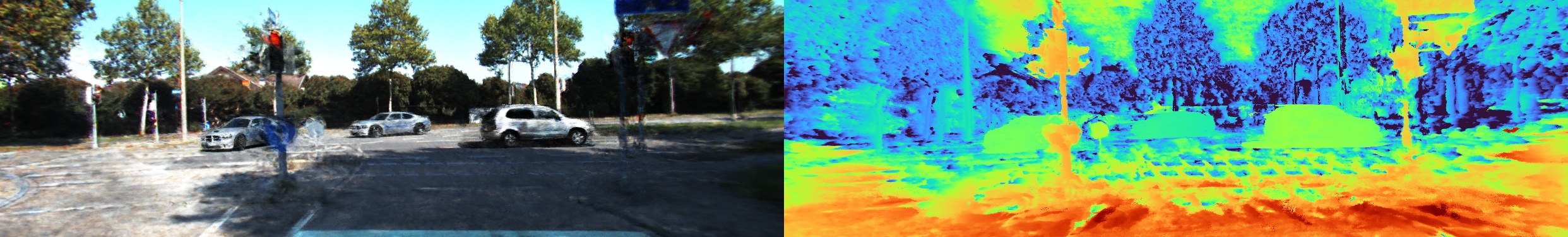} \\
\raisebox{0.07\textwidth}[0pt][0pt]{\rotatebox[origin=c]{90}{\tiny{EmerNeRF~\cite{yang2023emernerf}}}} 
& \includegraphics[width=0.94\textwidth]{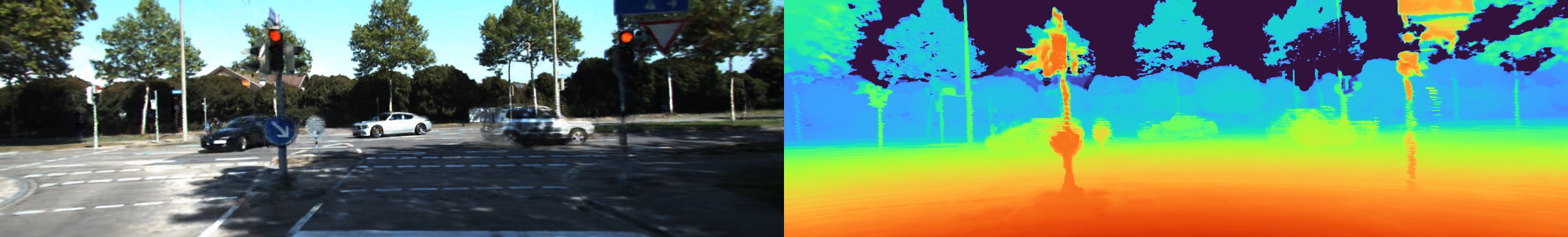} \\
\raisebox{0.07\textwidth}[0pt][0pt]{\rotatebox[origin=c]{90}{\tiny{PVG (Ours)}}} 
& \includegraphics[width=0.94\textwidth]{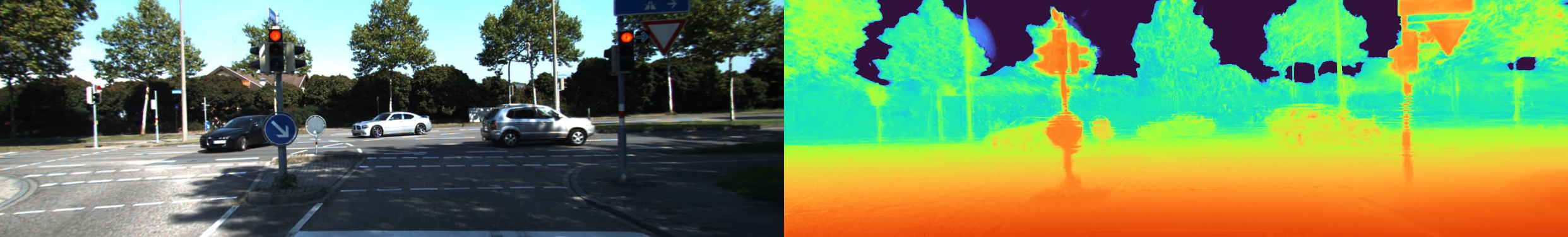} \\
\raisebox{0.07\textwidth}[0pt][0pt]{\rotatebox[origin=c]{90}{\tiny{GT}}} 
& \includegraphics[width=0.94\textwidth]{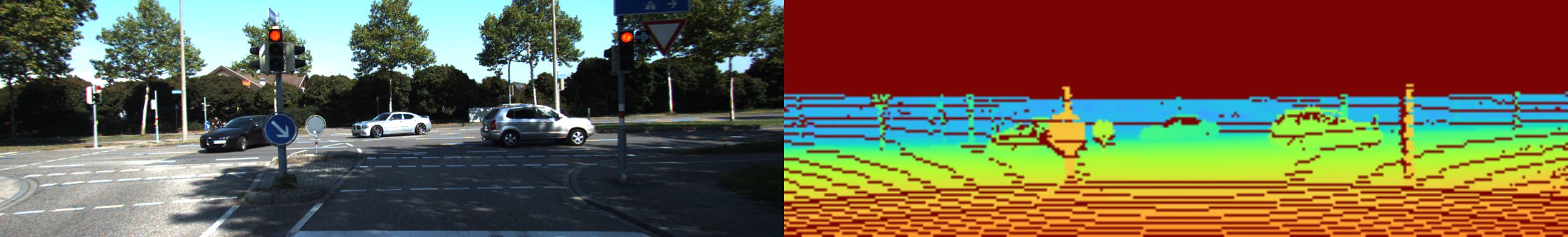} \\
\end{tabular}
\caption{Qualitative results of image reconstruction on KITTI.}
\label{fig:supp_kitti_reconstruction}
\end{figure}

\begin{figure}
\centering
\setlength\tabcolsep{1pt}
\begin{tabular}{cccccccc}
\raisebox{0.07\textwidth}[0pt][0pt]{\rotatebox[origin=c]{90}{\tiny{S-NeRF~\cite{xie2023s}}}} 
& \includegraphics[width=0.94\textwidth]{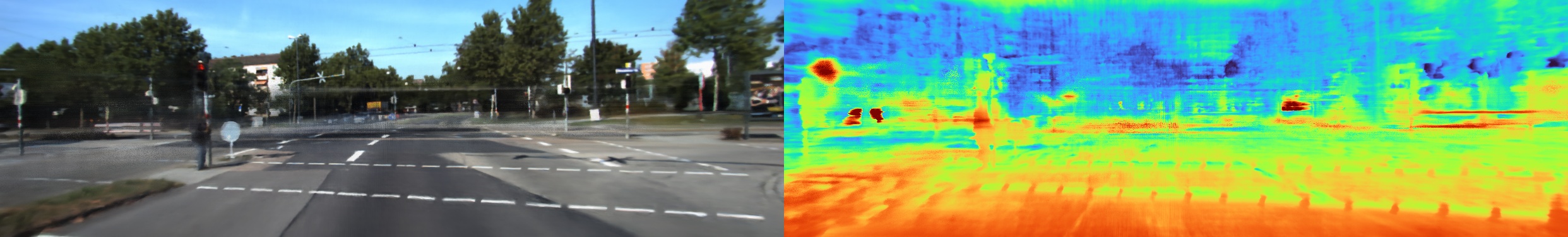} \\
\raisebox{0.07\textwidth}[0pt][0pt]{\rotatebox[origin=c]{90}{\tiny{StreetSurf~\cite{guo2023streetsurf}}}} 
& \includegraphics[width=0.94\textwidth]{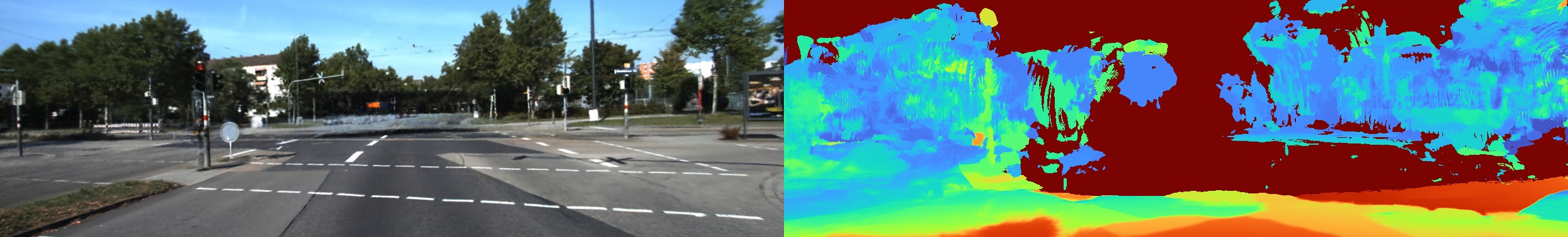} \\
\raisebox{0.07\textwidth}[0pt][0pt]{\rotatebox[origin=c]{90}{\tiny{3DGS~\cite{kerbl20233d}}}} 
& \includegraphics[width=0.94\textwidth]{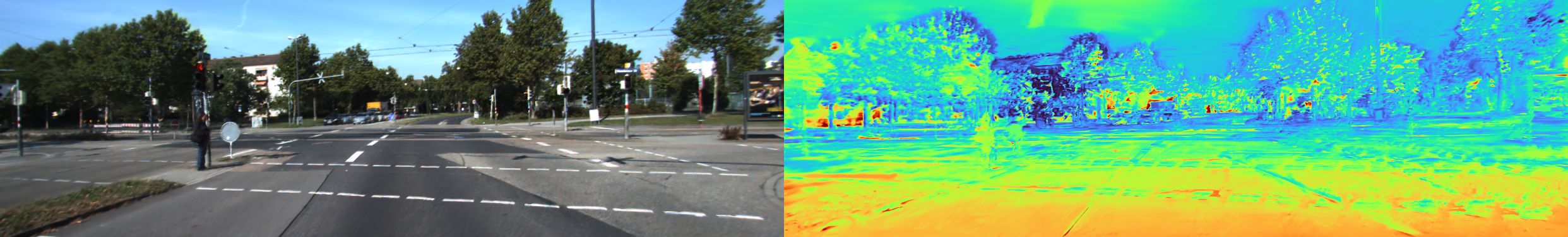} \\
\raisebox{0.07\textwidth}[0pt][0pt]{\rotatebox[origin=c]{90}{\tiny{NSG~\cite{ost2021neural}}}} 
& \includegraphics[width=0.94\textwidth]{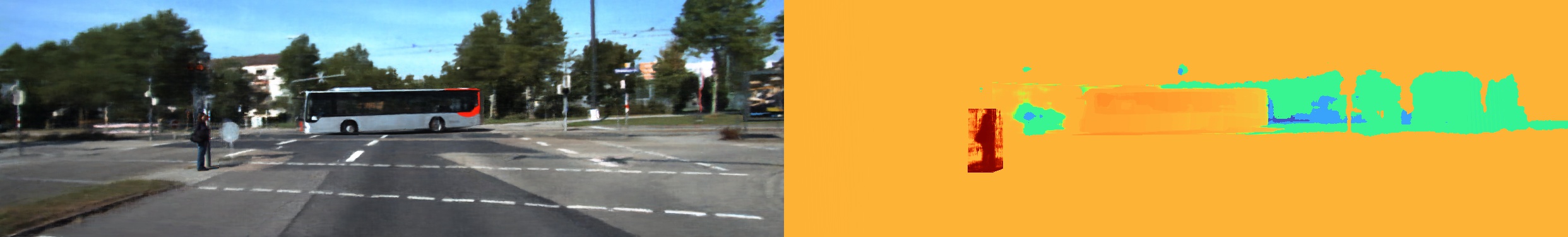} \\
\raisebox{0.07\textwidth}[0pt][0pt]{\rotatebox[origin=c]{90}{\tiny{SUDS~\cite{turki2023suds}}}} 
& \includegraphics[width=0.94\textwidth]{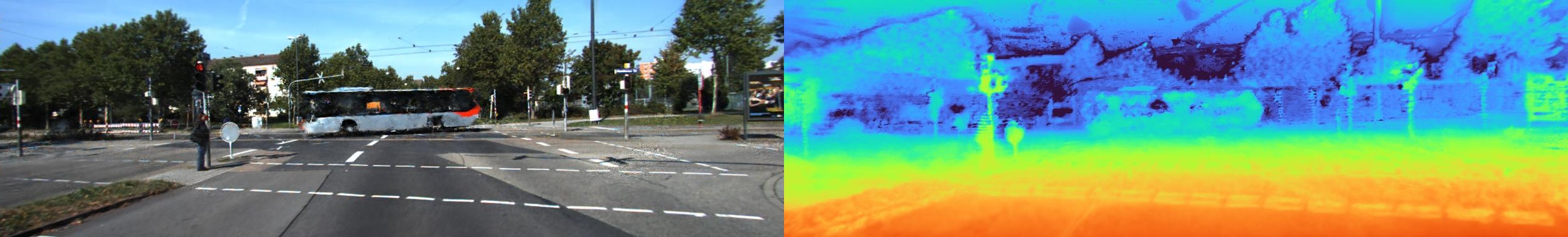} \\
\raisebox{0.07\textwidth}[0pt][0pt]{\rotatebox[origin=c]{90}{\tiny{Mars~\cite{wu2023mars}}}} 
& \includegraphics[width=0.94\textwidth]{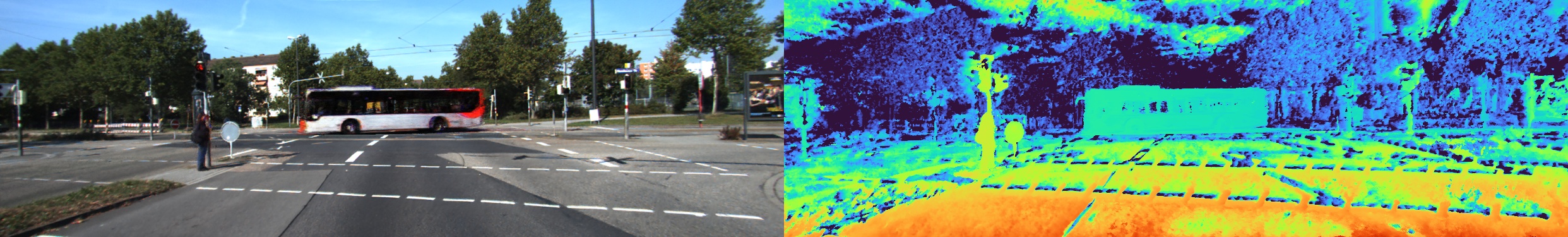} \\
\raisebox{0.07\textwidth}[0pt][0pt]{\rotatebox[origin=c]{90}{\tiny{EmerNeRF~\cite{yang2023emernerf}}}} 
& \includegraphics[width=0.94\textwidth]{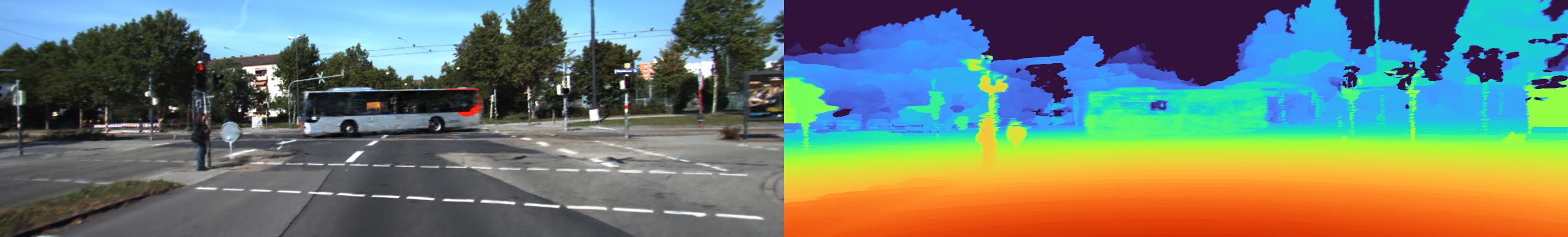} \\
\raisebox{0.07\textwidth}[0pt][0pt]{\rotatebox[origin=c]{90}{\tiny{PVG (Ours)}}} 
& \includegraphics[width=0.94\textwidth]{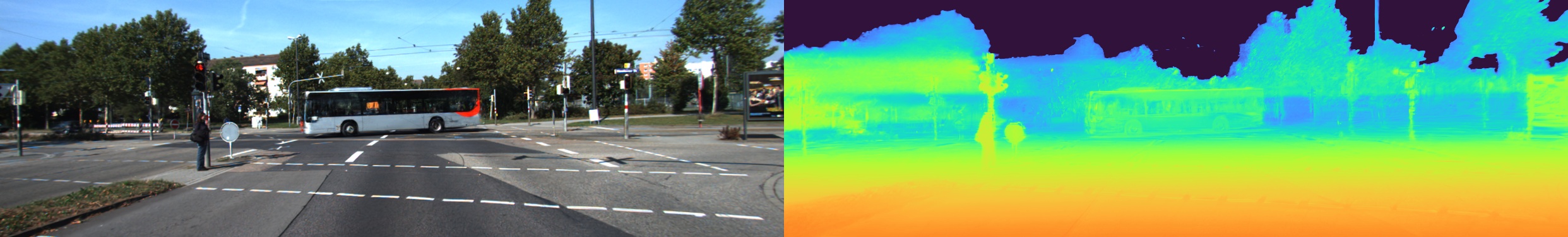} \\
\raisebox{0.07\textwidth}[0pt][0pt]{\rotatebox[origin=c]{90}{\tiny{GT}}} 
& \includegraphics[width=0.94\textwidth]{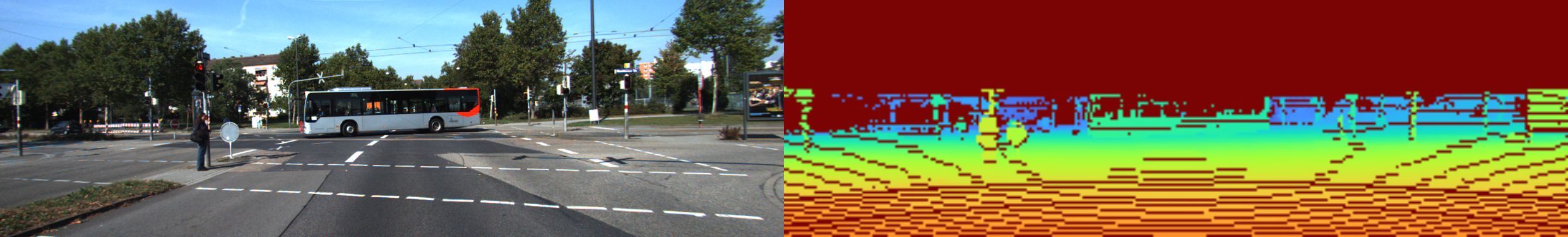} \\
\end{tabular}
\caption{Qualitative results of novel view synthesis on KITTI.}
\label{fig:supp_kitti_novel}
\end{figure}

The scenarios of KITTI used in~\cite{turki2023suds} are almost the scenarios where the ego vehicle is not moving. The quality of depth reconstruction depends on the point cloud. Image reconstruction results and novel view synthesis results are in  Fig.~\ref{fig:supp_kitti_reconstruction} and Fig.~\ref{fig:supp_kitti_novel}.

\clearpage

\clearpage

\clearpage

\end{appendices}
\bibliography{sn-bibliography}%

\end{document}